\documentclass[lettersize,journal]{IEEEtran}
\usepackage{amsmath,amsfonts}
\usepackage{array}
\usepackage[caption=false,font=normalsize,labelfont=sf,textfont=sf]{subfig}
\usepackage{textcomp}
\usepackage{stfloats}
\usepackage{url}
\usepackage{verbatim}
\usepackage{graphicx}
\usepackage{cite}
\usepackage{graphicx}
\usepackage{amsmath}
\usepackage{amssymb}
\usepackage{bm}
\usepackage{booktabs}
\usepackage{multirow}
\usepackage{longtable}

\usepackage{tcolorbox}
\tcbuselibrary{skins}
\usepackage{hyperref}
\usepackage{tikz}
\usepackage{xscommand}

\usepackage{forest}
\usepackage{enumitem}
\usetikzlibrary{shadows}
\usepackage{makecell}

\usepackage{booktabs}
\usepackage{multirow}
\usepackage{supertabular}
\usepackage{caption} 

\usepackage[capitalize]{cleveref}
\crefname{section}{Sec.}{Secs.}
\Crefname{section}{Section}{Sections}
\Crefname{table}{Table}{Tables}
\crefname{table}{Tab.}{Tabs.}
\begin{document}

\title{Devling into Adversarial Transferability on Image Classification: Review, Benchmark, and Evaluation}

\author{
Xiaosen Wang\textsuperscript{1}, Zhijin Ge\textsuperscript{2}, Bohan Liu\textsuperscript{2}, Zheng Fang\textsuperscript{3}, Fengfan Zhou\textsuperscript{1}, Ruixuan Zhang\textsuperscript{4}, \\
Shaokang Wang\textsuperscript{5*}, Yuyang Luo\textsuperscript{1*}\\
\textsuperscript{1}HUST, \textsuperscript{2}Xidian University, \textsuperscript{3}Wuhan University, \textsuperscript{4}HFUT, \textsuperscript{5}SJTU
\thanks{\textsuperscript{*}Corresponding author: \{lyuyang.andy, daniel095wang\}@gmail.com}
\thanks{TransferAttack:
\url{https://github.com/Trustworthy-AI-Group/TransferAttack}.}
}

\markboth{Journal of \LaTeX\ Class Files,~Vol.~14, No.~8, August~2021}%
{Shell \MakeLowercase{\textit{et al.}}: A Sample Article Using IEEEtran.cls for IEEE Journals}


\maketitle

\begin{abstract}
Adversarial transferability refers to the capacity of adversarial examples generated on the surrogate model to deceive alternate, unexposed victim models. This property eliminates the need for direct access to the victim model during an attack, thereby raising considerable security concerns in practical applications and attracting substantial research attention recently. In this work, we discern a lack of a standardized framework and criteria for evaluating transfer-based attacks, leading to potentially biased assessments of existing approaches. To rectify this gap, we have conducted an exhaustive review of hundreds of related works, organizing various transfer-based attacks into six distinct categories. Subsequently, we propose a comprehensive framework designed to serve as a benchmark for evaluating these attacks. In addition, we delineate common strategies that enhance adversarial transferability and highlight prevalent issues that could lead to unfair comparisons. Finally, we provide a brief review of transfer-based attacks beyond image classification.
\end{abstract}


\section{Introduction}
Recently, deep neural networks (DNNs) have achieved significant success in various domains, including image recognition~\cite{alex2012ImageNet,he2016deep,huang2017densely,dosovitskiy2020image} and natural language processing~\cite{hochreiter1997long,vaswani2017attention,devlin2018bert,brown2020language}.
However, their vulnerability against adversarial examples~\cite{szegedy2014intriguing}, in which imperceptible perturbation can mislead the DNNs, has posed significant threats to safety-critical applications that rely on DNNs, such as face recognition~\cite{tang2004video,wen2016discriminative,wang2018cosface} and autonomous driving~\cite{xu2017end,zhou2018voxelnet,vora2020pointpainting}. Investigating adversarial examples not only helps mitigate these threats but also provides insights into uncovering the black-box behavior of DNNs. Consequently, this area has drawn increasing attention for various tasks~\cite{sharif2016accessorize,eykholt2018robust,xie2017adversarial,wei2019transferable,alzantot2018generating,wang2021adversarial,yu2022texthacker,yang2024mma,goodfellow2015explaining,inkawhich2019feature,inkawhich2020transferable,wang2025attention}, particularly for image classification~\cite{moosavi2016deepfool,madry2017towards,croce2020reliable,brendel2017decision,ilyas2018black,luo2025disrupting,liu2025boosting,inkawhich2020perturbing}.

Based on victim model access, attacks can be categorized into white-box attack (full access)~\cite{goodfellow2015explaining,moosavi2016deepfool,madry2017towards} and black-box attack (limited access), which can be further divided into three categories, \ie, score-based attack~\cite{cheng2019improving,al2019sign,du2019query,dong2021query}, decision-based attack~\cite{cheng2018query,li2020qeba,wang2022triangle,reza2023cgba}, and transfer-based attack~\cite{liu2016delving,dong2018boosting,xie2019improving,wang2021enhancing,cao2025vit,gao2021feature,wang2023towardsb,wang2023lfaa}. Among the above attacks, transfer-based attacks present significant risks to real-world applications by crafting adversarial examples on surrogate models to deceive unknown target models directly~\cite{liu2016delving,dong2018boosting,xie2019improving}. Without access to victim models during the attack process, this has recently attracted enormous attention~\cite{dong2019evading,long2022frequency,wang2023rethinking}. However, despite its prevalence and several surveys~\cite{li2024towards,gu2023survey,zhao2023revisiting,zhang2024bag,jin2024short,wang2023beyond}, there is a lack of a unified framework and evaluation criteria. Consequently, many studies struggle to select appropriate benchmarks for assessments. To tackle this challenge, we summarize numerous transfer-based attacks and propose a standardized evaluation framework.

\begin{figure*}
    \centering
    \includegraphics[width=0.95\linewidth]{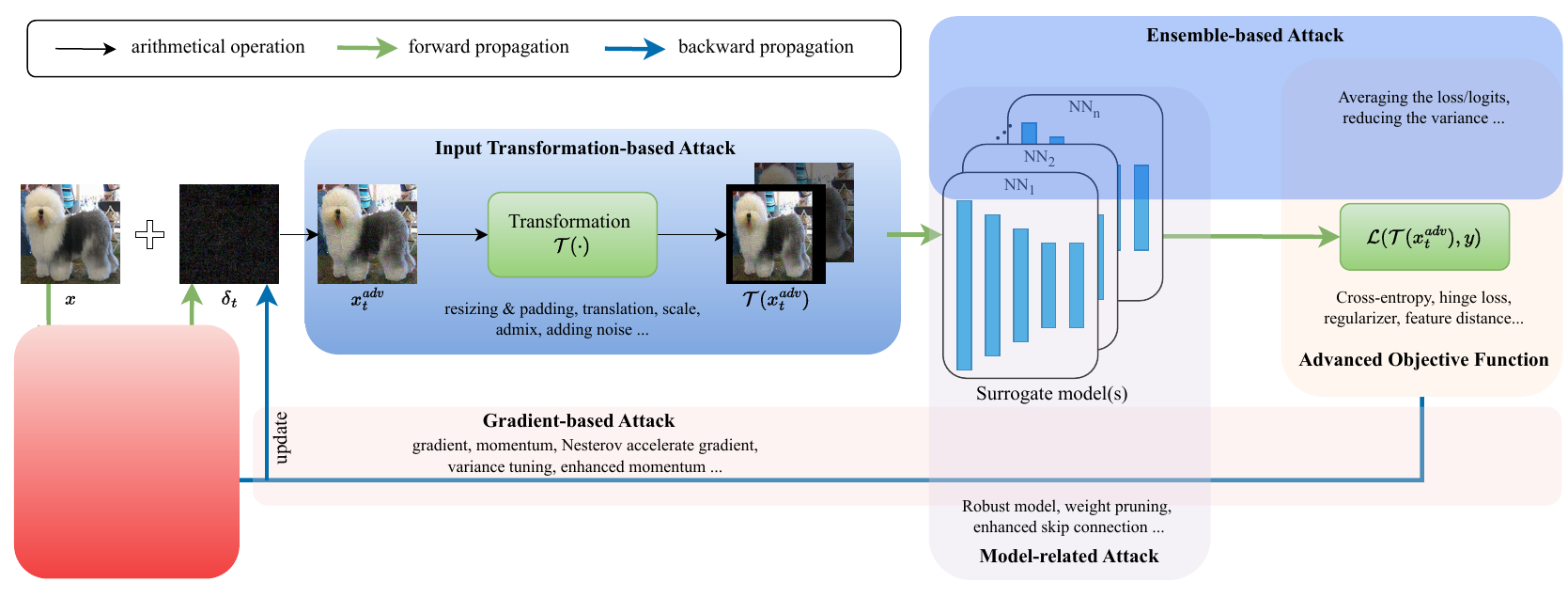}
    \caption{Categorization of existing transfer-based attacks. We systematically categorize existing transfer-based attacks into six distinct classes: 1) \textbf{Gradient-based Attack} optimizes the gradient calculation procedure to derive gradients that are more effective for the attack. 2) \textbf{Input Transformation-based attack} transforms the image prior to model to enhance the diversity of input image. 3) \textbf{Advanced objective function} replaces the conventional cross-entropy loss with a more complex objective function to enhance attack efficacy. 4) \textbf{Model-related attack} refines either the forward or backward propagation process, tailored to the architecture of the surrogate model employed. 5) \textbf{Ensemble-based Attack} adopts multiple surrogate models to generate adversarial examples. 6) \textbf{Generation-based Attack} trains a generator to directly crafts adversarial examples.
    }
    \label{fig:overview}
    \vspace{-1.em}
\end{figure*}

As shown in Fig.~\ref{fig:overview}, we categorize transfer-based attacks into six distinct classes according to their methodologies. Specifically, gradient-based attacks~\cite{goodfellow2015explaining,kurakin2017adversarial,dong2018boosting} boost transferability by utilizing advanced momentum or optimization strategies. Input transformation-based attacks~\cite{xie2019improving,dong2019evading,wang2021admix} transform the input image for gradient calculation. Advanced objective function~\cite{zhou2018transferable,huang2019enhancing,wu2020boosting} alters the traditional cross-entropy loss, focusing mainly on features. Generation-based attacks~\cite{naseer2019cross,kanth2021learning,kim2022diverse} train a generator to craft adversaries or perturbations. Model-related attacks~\cite{wu2020skip,guo2020backpropagating,zhu2021rethinking} modify the forward and backward propagation according to the model architecture. Ensemble-based attacks~\cite{liu2016delving,li2020learning} explore how to achieve better transferability by attacking multiple models.

Based on the objectives of attacks, existing transfer-based attacks can be categorized into untargeted and targeted attacks. Untargeted attacks deceive the victim models into making incorrect classifications, whereas targeted attacks intentionally direct the victim models to classify into predetermined categories. As summarized in Fig.~\ref{fig:summary}, we collect more than one hundred transfer-based attacks and classify them into the above types, forming the foundation of our framework. We briefly summarize each method and provide a unified evaluation under the same setting to evaluate the attacks among each category. Our contributions can be summarized as follows:
\begin{itemize}[leftmargin=*,noitemsep,topsep=1pt]
\item As depicted in Fig.~\ref{fig:overview}, we systematically categorize existing transfer-based attacks into six classes, namely gradient-based attacks, input transformation-based attacks, advanced objective function, generation-based attacks, model-related attacks, and ensemble-based attacks.
\item We provide a comprehensive summary of over one hundred untargeted and targeted transfer-based attacks, which is the most extensive overview of transfer-based attacks to date.
\item We introduce a unified framework for each attack category and pinpoint instances where certain methodologies fail to outperform established baselines. It indicates that some previous studies may not have conducted equitable comparisons, which informs the evaluation criteria for transfer-based attacks and promotes more comprehensive evaluation. 
\item Through an analysis of existing transfer-based attacks, we delineate common insights and potential factors contributing to enhanced transferability, thereby inspiring researchers to develop more robust and potent attack strategies.
\item We provide a concise overview of adversarial transferability across different domains, such as transfer-based attacks on other tasks (\eg, object detection, large language models, \etc), cross-task transfer-based attacks and so on.
\end{itemize}

The rest of paper is organized as follows. Sec.~\ref{sec:pre} shows the preliminaries on transfer-based attacks and evaluation settings. 
Sec.~\ref{sec:untargeted} and \ref{sec:targeted} give the summary, evaluations, and key insights regarding untargeted and targeted attacks, respectively. Sec.~\ref{sec:dis} discusses several critical issues that have not been sufficiently addressed. Sec.~\ref{sec:beyond} introduces transfer-based attacks beyond image classification. Finally, Sec.~\ref{sec:con} concludes.

\section{Preliminaries}
\label{sec:pre}
Here, we provide a brief overview of attacks, summarize terminology, and detail the evaluation settings.

\subsection{Adversarial Attacks}

Let $x$ represent a benign image and $y$ its corresponding true label, with $f(x)$ representing the classifier that yields the prediction result. The loss function of classifier $f$ is denoted as $\mathcal{L}(f(x),y)$, such as the cross-entropy loss. $\mathcal{B}_\epsilon(x)=\{x':\|x-x'\|_p\le\epsilon\}$ denotes the $\ell_p$-norm
ball centered at $x$ with radius $\epsilon$. The objective of an adversarial attack is to identify an adversarial example $x^{adv}\in \mathcal{B}_{\epsilon}(x)$ that misleads the target classifier $f$ such that $f(x)=y\neq f(x^{adv})$. In general, the adversarial attack can be formalized as:
\begin{equation}
    x^{adv}=\argmax_{x'\in \mathcal{B}_\epsilon(x)} \mathcal{L}(f(x'), y).
    \label{eq:goal}
\end{equation}

\definecolor{coral}{rgb}{1.0, 0.5, 0.31}
\definecolor{darkpink}{rgb}{0.91, 0.33, 0.5}
\definecolor{celadon}{rgb}{0.67, 0.88, 0.69}
\definecolor{tuftsblue}{rgb}{0.28, 0.57, 0.81}

\begin{figure*}
    \centering
    
    \begin{forest}
        for tree={%
            draw={
                gray, 
                thick, 
                font=\sffamily\scriptsize, 
                rounded corners=2, 
                fill=., 
            },
            where level=0{fill=yellow!30}{}, 
            where level=1{fill=coral!20}{}, 
            where level=2{fill=celadon!60}{}, 
            where level=3{fill=cyan!30}{}, 
            l sep+=5pt,
            grow'=east,
            edge={gray, thick},
            parent anchor=east,
            child anchor=west,
            if n children=0{tier=last}{},
            edge path={
                \noexpand\path [draw, \forestoption{edge}] (!u.parent anchor) -- +(10pt,0) |- (.child anchor)\forestoption{edge label};
            },
            if={isodd(n_children())}{
                for children={
                    if={equal(n,(n_children("!u")+1)/2)}{calign with current}{}
                }
            }{},
            align=left
        }
        [{\rotatebox{90}{\textbf{Transfer-based Attack}}}
            [{\rotatebox{90}{\textbf{Untargeted Attack}}}
                [Gradient-based Attack
                    [{\begin{tabular}{@{}l@{}}\cite{goodfellow2015explaining},\cite{kurakin2017adversarial},\cite{dong2018boosting},\cite{lin2020nesterov},\cite{gao2020patch},\cite{wang2021enhancing},\cite{wang2021boosting},\cite{gao2021staircase},\cite{zou2022making},\cite{zhang2022improvingb},\cite{qin2022boosting},\cite{han2023samplingbased},\cite{wan2023adversarial},\cite{peng2023boosting},\cite{zhu2023boosting},\cite{wu2023gnp},\cite{ma2023transferable},\cite{yang2023improving},\cite{ge2023boosting},\tabularnewline
                    \cite{fang2024strong},\cite{wang2022boosting},\cite{long2024convergence}\cite{wang2024fgsra},\cite{li2024foolmix},\cite{qiu2025mef},\cite{gan2025boosting},\cite{ren2025improving}
                    \end{tabular}}]
                ]
                [{\begin{tabular}{@{}c@{}}Input Transformation-\\based Attack\end{tabular}}
                    [{\begin{tabular}{@{}l@{}}\cite{xie2019improving},\cite{dong2019evading},\cite{lin2020nesterov},\cite{zou2020improving},\cite{wang2021admix},\cite{wu2021improving},\cite{long2022frequency},\cite{yuan2022adaptive},\cite{zhang2023improving},\cite{Wei2023BoostingAT},\cite{wang2023structure},\cite{ge2023improving},\cite{wang2023boost},\cite{lin2024boosting},\cite{zhu2024learning},\cite{wang2024boosting},\tabularnewline
                    \cite{qian2024enhancing},\cite{fan2025maskblock},\cite{guo2025OPS}\end{tabular}}]
                ]                
                [{\begin{tabular}{@{}c@{}}Advanced Objective Function \end{tabular}}
                    [{\begin{tabular}{@{}l@{}}\cite{zhou2018transferable},\cite{huang2019enhancing},\cite{wu2020boosting},\cite{li2020yet},\cite{wang2021feature},\cite{wang2021unified},\cite{wang2021exploring},\cite{huang2022integtransferable},\cite{he2022enhancing},\cite{zhang2022improvinga},\cite{zhang2022enhancing},\cite{yang2023fuzziness},\cite{jin2023danaa},\cite{li2023improving},\cite{wang2024improving},~\cite{liupixel2feature},~\cite{zheng2025enhancing}\end{tabular}}]
                ]                
                [Generation-based Attack 
                    [{\cite{naseer2019cross},\cite{kanth2021learning},\cite{kim2022diverse},\cite{zhu2024geadvgan},\cite{chen2024diffattack},\cite{wang2024boostingb},\cite{wu2025dsva}}]
                ]                
                [Model-related Attack 
                    [{\begin{tabular}{@{}l@{}}\cite{wu2020skip},\cite{guo2020backpropagating},\cite{fang2022learning},\cite{zhu2021rethinking},\cite{zhu2022towards},\cite{qin2023training},\cite{yang2023generating},\cite{yang2023boosting},\cite{wang2023diversifying},\cite{wang2023rethinking},\cite{wang2024ags},\cite{weng2024exploring},\cite{ma2024improving},\cite{yang2024quantization},                    \cite{wei2022towards},\tabularnewline
                     \cite{zhou2022improving},\cite{naseer2022on},\cite{wang2022generating},\cite{zhang2023transferable},\cite{zhang2024improving},\cite{ming2024att},\cite{ren2025fpr},\cite{chen2025enhancing},\cite{wang2025improving},\cite{li2025enhancing},\cite{gao2024attacking},\cite{liu2025ll2s}\end{tabular}}]
                ]                
                [Ensemble-based Attack 
                    [{\cite{liu2016delving},\cite{li2020learning},\cite{xiong2022stochastic},\cite{gubri2022lgv},\cite{li2023making},\cite{chen2023anadaptive},\cite{chen2024rethinking},\cite{tang2024ensemble}}]
                ]
            ]            
            [{\rotatebox{90}{\textbf{Targeted Attack}}}
                [{\begin{tabular}{@{}c@{}}Input Transformation-based Attack\end{tabular}} 
                    [{\cite{byun2022improving},\cite{wei2023enhancing},\cite{liu2024boosting},\cite{zeng2025everywhere}}]
                ]                
                [Advanced Objective Function 
                    [{\cite{li2020towards},\cite{zhao2021success},\cite{weng2023logit},\cite{byun2023introducing},\cite{zeng2024enhancing}}]
                ]                
                [Generation-based Attack
                    [{\cite{naseer2021generating},\cite{zhao2023minimizing},\cite{li2025aim}}]
                ]
                [Ensemble-based Attack 
                    [{\cite{wu2024improving},\cite{wang2025breaking}}]
                ]
            ]
        ]
    \end{forest}
    \caption{The overview of existing transfer-based adversarial attacks on Image Classification. Initially, these attacks are categorized into two principal types: untargeted and targeted attack. Subsequently, we further classify them into six distinct classes, as depicted in Fig.~\ref{fig:overview}. The arrangement of the referenced papers follows a chronological order, based on their dates of publication. }
    \label{fig:summary}
\end{figure*}
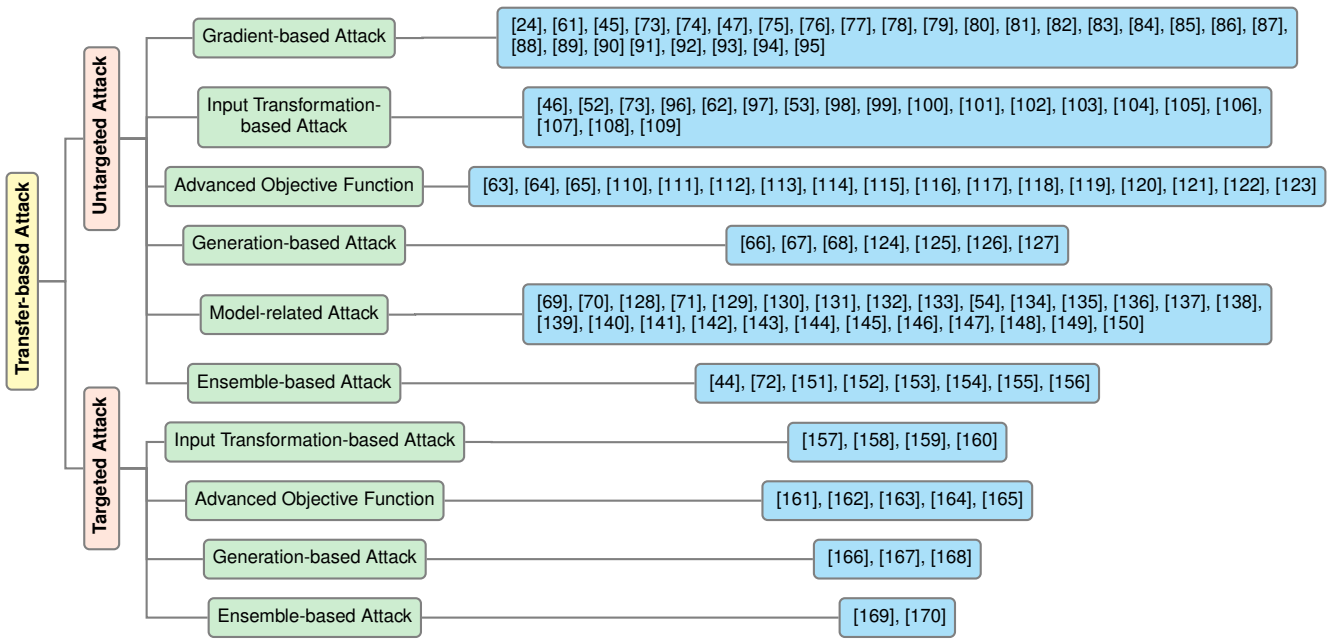

Based on the attacker’s knowledge of the victim model, attacks fall into two settings: 1) \textbf{White-box attacks} allow full access to target model, including its architecture, parameters, loss function, gradient, output, \etc. 2) \textbf{Black-box attacks} allow limited access (\eg, output) or no access to the victim model, which can be further split into three categories. a) \textbf{Score-based attacks} utilize the output logits to search for adversarial examples. b) \textbf{Decision-based attacks} adopt the output label to optimize the adversarial example. c) \textbf{Transfer-based attacks} employ the adversaries generated on surrogate model $f'$ to directly deceive victim model $f$. Among these, transfer-based attacks are particularly noteworthy, as they require no interaction with the victim model and have attracted substantial research interest.

Recently, numerous studies have been introduced to enhance adversarial transferability. Despite its prosperity, there remains no systematic review or standardized benchmark for evaluating transfer-based attacks. Consequently, some studies may rely on inadequate baselines for performance assessment, potentially leading to inaccurate conclusions. To address this issue, we provide a detailed summary of existing transfer-based attack methodologies and introduce a comprehensive, rigorous, and coherent benchmark to effectively assess their transferability. This benchmark aims to standardize evaluation metrics and conditions, thereby enabling more precise comparisons and advancing the field. By analyzing the quantitative results, we also delineate common insights for each category of transfer-based attacks, which can inspire more robust attack and defense strategies. The paper focuses mainly on the adversarial transferability of image classification, but we also provide a brief overview of transfer-based attacks across other domains.

\begin{table}[tb]
    \centering
    \caption{The terminologies used across the paper.}
    \label{tab:terminologies}
    \begin{tabular}{>{\centering\arraybackslash}m{.12\textwidth} m{.32\textwidth}}
        \toprule
        Terminologies & \multicolumn{1}{c}{Description}\\
        \midrule
        Benign sample & The clean input sample that can be correctly classified by DNNs.\\
        \rowcolor{Gray} Adversarial example & The sample that is visually realistic to humans but misleads the DNNs. It is often crafted by adding imperceptible perturbation on the benign samples.\\
        Perturbation & The noise that is added to the benign sample to generate an adversarial example.\\
        \rowcolor{Gray} Surrogate model & The white-box model that is adopted by transfer-based attack to generate adversarial examples.\\
        Victim model & The unknown target model that is attacked by transfer-based attack.\\
        \rowcolor{Gray} White-box/Black-box Attack & White-box attacks have full access to victim model, while black-box attacks not. One of the key criterions is whether to adopt the gradient for the attack.\\
         Untargeted/Targeted Attack & It generates adversarial examples that mislead the victim model into wrong/specific classes.\\
        \rowcolor{Gray} Transferability & The ability of adversarial examples generated on the surrogate model to fool victim model.\\
        Attack success rates & The misclassification rates of target models on the adversarial examples as shown in Eq.~\eqref{eq:asr}. \\
        \bottomrule
    \end{tabular}
\end{table}
\subsection{Terminologies}
For ease of understanding, we summarize the commonly used terminologies in Tab.~\ref{tab:terminologies}, including benign sample, adversarial example, (adversarial) perturbation, surrogate model, victim model, white-box/black-box attack, untargeted/targeted attack, transferability and attack success rates.

\subsection{Evaluation Settings}
We compare the attack performance of transfer-based attacks across each category in Fig.~\ref{fig:summary}. The combination of various attacks from different categories is beyond discussion. We summarize the evaluation settings as follows:

\textbf{Models}. We adopt four convolutional neural networks (ResNet-50 (RN-50)~\cite{he2016deep}, VGG-16~\cite{simonyan2015very}, Mobilenet-v2 (MN-v2)~\cite{sandler2018mobilenetv2}, and Inception-v3 (Inc-v3)~\cite{szegedy2016rethinking}), four vision transformers (ViT~\cite{dosovitskiy2020image}, PiT~\cite{heo2021rethinking}, Visformer~\cite{chen2021visformer}, and Swin~\cite{liu2021swin}), and five defense mechanisms (AT~\cite{wong2020fast}, HGD~\cite{liao2018defense}, RS~\cite{Cohen2019RS},  NRP~\cite{naseer2020a} and DiffPure~\cite{nie2022diffusion}).  Unless specified, we generate adversarial examples on RN-50 and evaluate the attack performance on the other models.

\textbf{Datasets}. We conduct our experiments on the ImageNet-compatible dataset~\cite{alex2012ImageNet}, which is widely used in previous works~\cite{long2022frequency,wang2021enhancing,dong2019evading}. It contains 1,000 images with the size of $299 \times 299$, ground-truth labels, and target labels for targeted attacks. All the images are resized to $224\times 224$.

\textbf{Parameters}. Following common practice, perturbations are constrained by the norm $\ell_\infty$, with an upper limit of $\epsilon = 16/255$ and a step size $\alpha=1.6/255$. The iterative process is configured with $T = 10$ iterations for untargeted attacks and extends to $T = 300$ iterations for targeted attacks. The default backbone attack is MI-FGSM~\cite{dong2018boosting} for input transformation-based attack, advanced objective, model-related attack and ensemble-based attack unless it has changed the optimization procedure. Detailed descriptions of the hyper-parameters for each attack are available in our framework.

\textbf{Evaluation metric}. We adopt the attack success rates (ASR) on the victim model $f$ to evaluate the transferability:
\begin{equation}
    \label{eq:asr}
    ASR(\mathcal{A}) = \frac{1}{|\mathcal{A}|} \sum_{x_i^{adv} \in \mathcal{A}} \mathbbm{1}(f(x_i^{adv}) \neq y_i),
\end{equation}
where $\mathcal{A}$ is the set of adversarial examples, $\mathbbm{1}(\cdot)$ is the indicator function and $y_i$ is the ground-truth label of $x_i^{adv}$.

\section{Untargeted Attacks}
\label{sec:untargeted}
In this section, we provide a brief overview and evaluation of untargeted transfer-based attacks across the six aforementioned categories and delineate common insights.

\subsection{Gradient-based Attack}
Gradient-based attacks modify the gradient calculation procedure to optimize the perturbation, incorporating techniques such as momentum, variance tuning, and guiding adversarial examples toward flat local minima. These attacks form the foundation for transfer-based attacks across various categories.

\Review{wxs}{\textbf{Fast Gradient Sign Method (FGSM)}~\cite{goodfellow2015explaining} generates adversarial examples by maximizing the classification loss value with one-step update:,
\begin{equation}
\label{eq:fgsm}
    x^{adv}=\Pi_{\mathcal{B}_\epsilon(x)}\left\{x+ \epsilon \cdot \operatorname{sign}\left(\nabla_{x}\mathcal{L}\left(f\left(x\right),y\right)\right)\right\},
\end{equation}
where $\mathcal{B}_\epsilon(x) = \{x': \Vert x-x'\Vert_p\le \epsilon \}$, $\Pi_{\mathcal{B}_\epsilon(x)}(x')$ clips $x'$ into $\mathcal{B}_\epsilon(x)$, and $\mathrm{sign}(\cdot)$ is the sign function.}

\Review{wxs}{\textbf{Iterative Fast Gradient Sign Method (I-FGSM)}~\cite{kurakin2017adversarial} iteratively apply FGSM to find an adversarial example with a small step size $\alpha=\epsilon/T$:
\begin{equation}
\label{eq:ifgsm}
    x^{adv}_{t+1}=\Pi_{\mathcal{B}_\epsilon(x)}\left\{x^{adv}_{t}+ \alpha \cdot \operatorname{sign}\left(\nabla_{x^{adv}_{t}}\mathcal{L}\left(f\left(x^{adv}_{t}\right),y\right)\right)\right\},
\end{equation}
where $x_0^{adv}=x$ and $T$ is the number of iteration.}

Though I-FGSM achieves superior white-box attack performances, it exhibits poor adversarial transferability. To boost the transferability, various gradient-based attacks are proposed.

\Review{wxs}{\textbf{Momentum Iterative Gradient Sign Method (MI-FGSM)}~\cite{dong2018boosting} integrates the momentum term into the iterative process to stabilize the update directions,
\begin{equation}
\begin{aligned}
    g_{t+1} &= \mu \cdot g_{t} + \frac{\nabla_{x}\mathcal{L}(f(x^{adv}_{t}),y)}{\Vert \nabla_{x}\mathcal{L}(f(x^{adv}_{t}),y) \Vert_{1}}, \\
    x^{adv}_{t+1} &= \Pi_{\mathcal{B}_\epsilon(x)}\left\{x^{adv}_{t} + \alpha \cdot \operatorname{sign}(g_{t+1})\right\},
    \label{eq:mifgsm}
\end{aligned}
\end{equation}
where $\mu$ is the decay factor.}


\Review{wxs}{\textbf{Nesterov accelerated Iterative Gradient Sign Method (NI-FGSM)}~\cite{lin2020nesterov} integrates the  Nesterov’s accelerated gradient by substituting $x_t^{adv}$ in Eq.~\eqref{eq:mifgsm} with $x_t^{adv}+\alpha\cdot\mu\cdot g_t$ to accumulate the momentum.}


\Review{wxs}{\textbf{Patch-wise Iterative  Fast Gradient Sign Method (PI-FGSM)}~\cite{gao2020patch} introduces a pre-defined project kernel $W$ to assign the pixel's gradient overflowing the $\epsilon$-constraint to its surrounding area:
\begin{gather}
    g_t = \nabla_{x}\mathcal{L}(f(x^{adv}_{t}),y), \quad c = \alpha_t + \beta \cdot \frac{\epsilon}{T}\cdot \operatorname{sign}(g_t), \nonumber\\
         C = \Pi_{[0,+\infty]}\{|c|-\epsilon\}\cdot \operatorname{sign}(c),\\
         \alpha_{t+1} = \alpha_t + \beta \cdot \frac{\epsilon}{T}\cdot\operatorname{sign}(g_t) + \gamma \cdot \operatorname{sign}(W*C),\nonumber\\
         x^{adv}_{t+1}=\Pi_{\mathcal{B}_\epsilon(x)}\left\{x^{adv}_{t} +  \beta \cdot \frac{\epsilon}{T}\cdot \operatorname{sign}\left(g_{t}\right)  + \gamma \cdot \operatorname{sign}(W*C) \right\}.\nonumber
\end{gather}
Here $\beta$ is a amplification factor and $\gamma$ is the project factor.}


\Review{wxs}{\textbf{Variance Tuning}~\cite{wang2021enhancing} defines the gradient variance as the difference between the mean gradient in its $\epsilon$-neighborhood and the gradient of raw input:
\begin{equation}
    V_\epsilon(x)=\mathbb{E}_{x'\in \mathcal{B}_\epsilon(x)}[\nabla_{x'}\mathcal{L}(f(x'),y)] - \nabla_{x}\mathcal{L}(f(x),y),
\end{equation}
which is used to tune the gradient at each iteration and general to any gradient-based attacks. For instance, variance tuning MI-FGSM (VMI-FGSM) tunes the gradient in Eq.~\eqref{eq:mifgsm} as:
\begin{equation}
    g_{t+1} = \mu \cdot g_t + \frac{\nabla_{x}\mathcal{L}(f(x^{adv}_{t}),y)+V_\epsilon(x_t^{adv})}{\Vert \nabla_{x}\mathcal{L}(f(x^{adv}_{t}),y) + V_\epsilon(x_t^{adv}) \Vert_{1}}.
\end{equation}}

\Review{wxs}{\textbf{Enhanced Momentum Iterative Gradient Sign Method (EMI-FGSM)}~\cite{wang2021boosting} additionally accumulates the average gradient of several data points linearly sampled in the gradient direction of previous iteration to enhance the momentum and stabilize the update directions:
\begin{equation}
    \begin{aligned}
        \bar{g}_t &= \frac{1}{N}\sum_{i=1}^N \nabla_x \mathcal{L}(f(x_t^{adv}+c_i\cdot \bar{g}_{t-1}), y), 
    \end{aligned}
\end{equation}
where $c_i$ is the linearly sampled coefficient in a predefined interval and $N$ is the sampling number.}

\Review{wxs}{\textbf{Iterative Fast Gradient Staircase Sign Method (I-FGS$^{2}$M)}~\cite{gao2021staircase} alleviates the issue of disregarding value differences in FGSM family by assigning staircase weights to each interval of the gradient:
\begin{equation}
    x_{t+1}^{adv} = \Pi_{\mathcal{B}_\epsilon (x)}\left\{x_t^{adv} + \alpha\cdot W_t \odot \operatorname{sign}(\nabla_x \mathcal{L}(f(x_t^{adv}),y))\right\},
\end{equation}
where $\odot$ is Hadamrd product and $W_t$ is a scale matrix based on $|\nabla_x \mathcal{L}(f(x_t^{adv}),y)|$.}

\Review{wxs}{\textbf{Adam Iterative Fast Gradient Tanh Method (AI-FGTM)}~\cite{zou2022making} adopts Adam~\cite{kingma2015adam} to adjust the step size and momentum using the tanh function:
\begin{gather}
    \beta^t = \frac{1-\beta_1^{t+1}}{\sqrt{1-\beta_2^{t+1}}}, \quad g_t = \nabla_x \mathcal{L}(f(x_t^{adv},y)), \nonumber\\
    m_{t+1} = m_t + \mu_1 \cdot g_t, \quad v_{t+1} = v_t + \mu_2 \cdot g_t^2,
    \\
    x_{t+1}^{adv}=\Pi_{\mathcal{B}(x)}\left\{x_t^{adv} + \frac{\epsilon \cdot \beta^{t+1}}{\sum_{i=0}^{T-1} \beta^i} \tanh(\lambda \frac{m_{t+1}}{\sqrt{v_{t+1}}+\delta})\right\}, \nonumber
\end{gather}
where $\mu_1$, $\mu_2$, $\lambda$, $\beta_1$ and $\beta_2$ are hyper-parameters.}

\Review{wxs}{\textbf{Iterative Fast Gradient Sign Method with Virtual step and Auxiliary gradients (VA-I-FGSM)}~\cite{zhang2022improvingb} allows the adversarial example to exceed the $\epsilon$-neighboorhood of the input image with a larger step size, and extra adopts the auxiliary gradients from other categories to update the perturbation.}

\Review{wxs}{\textbf{Reverse Adversarial Perturbation (RAP)}~\cite{qin2022boosting} injects the worst-case perturbation when calculating the gradient to generate adversarial example in a more flat local region:
\begin{gather}
    \max_{\bm{x}^{adv}\in \mathcal{B}(x)} \mathcal{L}(f(\bm{x}^{adv} + \bm{n}^{rap}), y), \\
    \mathrm{where}  \ \bm{n}^{rap} = \argmin_{\|\bm{n}^{rap}\|_\infty \le \epsilon_n} \mathcal{L}(f(\bm{x}^{adv} + \bm{n}^{rap}), y). \nonumber
\end{gather}}

\Review{lyy,wxs}{\textbf{Sampling-based Fast Gradient Rescaling Method (SMI-FGRM)}~\cite{han2023samplingbased} accumulates the rescaled gradient of several sampled data points:
\begin{gather}
    g_t = \frac{1}{N+1}\sum^{N}_{i=0}\operatorname{rescale}(\nabla J(x^{i}_{t},y;\theta)), \quad x^{i+1}_{t} = x^{i}_{t} + \xi_{i}, \nonumber \\
    \operatorname{rescale}(g) = c * \operatorname{sign}(g) \odot \sigma(\operatorname{nrom}(\operatorname{log}_2|g|)), \\
    \operatorname{norm}(x) = \frac{x-\operatorname{mean}(x)}{\operatorname{std}(x)}, \quad \sigma(x) = \frac{1}{1 + e^{-x}}.\nonumber
\end{gather}
}

\begin{table*}[tb]
    \centering
    \caption{Attack success rates of various gradient-based adversarial attacks where the adversarial examples are generated on RN50. We have highlighted the approaches that achieve the top-5 attack performance in \colorbox{gray!30}{gray}.}
    \label{tab:untargeted:gradient-based-attack}
    \vspace{-.5em}
    \resizebox{\textwidth}{!}{
    \begin{tabular}{l|cccccccc|ccccc|c}
        \toprule
        \multicolumn{1}{c}{\multirow{2}{*}{Attacks}} & \multicolumn{8}{c}{Standardly trained models} & \multicolumn{5}{c}{Defenses} & \multicolumn{1}{c}{\multirow{2}{*}{Average}} \\
    \cmidrule(lr){2-9} \cmidrule(lr){10-14}
        & RN-50 & VGG-16 & MN-v2 & Inc-v3 & ViT & PiT & Visformer & Swin & AT & HGD& RS & NRP & DiffPure\\
        \midrule\midrule
        FGSM~\cite{goodfellow2015explaining} & ~~49.2 & 54.6 & 48.2 & 32.8 & 11.9 & 14.1 & 18.7 & 19.4 & 40.0 & ~~8.8 & 26.8 & 53.1 & 15.8 & 30.3\\
        I-FGSM~\cite{kurakin2017adversarial} & ~~99.6 & 36.5 & 33.6 & 17.7 & ~~7.5 & 11.7 & 14.4 & 15.7 & 39.7 & ~~5.9 & 26.2 & 42.3 & ~~8.7 & 27.7\\
        MI-FGSM~\cite{dong2018boosting} & ~~99.9 & 57.9 & 53.4 & 37.4 & 14.5 & 22.5 & 26.2 & 28.1 & 40.6 & 17.9 & 27.4 & 58.5 & 13.3 & 38.3\\
        NI-FGSM~\cite{lin2020nesterov} & 100.0 & 66.5 & 59.3 & 38.9 & 15.4 & 23.4 & 30.1 & 29.7 & 40.8 & 18.2 & 27.6 & 57.7 & 12.7 & 40.0\\
        PI-FGSM~\cite{gao2020patch} & ~~98.8 & 74.9 & 59.2 & 57.9 & 16.3 & 16.9 & 25.3 & 20.6 & 42.1 & 35.3 & 35.0 & 60.1 & 38.3 & 44.7\\
          VMI-FGSM~\cite{wang2021enhancing}  & ~~99.6 & 70.8 & 66.9 & 57.3 & 31.9& 47.0& 54.5& 53.5& 41.9& 47.5& 30.5& 61.3 &22.9 & 52.7\\
        EMI-FGSM~\cite{wang2021boosting}  & 100.0 & 84.7 & 81.7 & 64.3 & 25.2 & 43.1 & 56.2 & 53.1 & 43.6 & 42.5 & 30.3 & 67.3 & 19.1 & 54.7\\
        I-FGS$^2$M~\cite{gao2021staircase}  & ~~99.8 & 45.6 & 41.5 & 24.6 & ~~9.5 & 14.3 & 19.1 & 20.4 & 39.8 & ~~9.3 & 26.7 & 47.1 & ~~9.9 & 31.4\\
        AI-FGTM~\cite{zou2022making} & ~~99.7 & 52.9 & 49.0 & 33.0 & 13.2 & 20.5 & 26.1 & 25.8 & 40.5 & 16.4 & 27.3 & 55.1 & 11.9 & 36.3 \\
        VA-I-FGSM~\cite{zhang2022improvingb} & ~~99.4 & 53.2 & 46.4 & 26.8 & ~~9.6 & 12.2 & 15.7 & 17.7 & 40.3 & ~~7.6 & 26.8 & 53.8 & ~~9.9 & 32.3\\
        RAP~\cite{qin2022boosting} & ~~99.9 & 87.8 & 85.9 & 63.0 & 23.8 & 40.8 & 54.1 & 53.7 & 42.9 & 26.5 & 32.0 & 65.6 & 21.0 & 53.6\\
        SMI-FGRM~\cite{han2023samplingbased} & ~~99.2 & 75.1 & 73.5 & 67.0 & 25.9 & 40.9 & 51.9 & 48.9 & 46.4 & 47.5 & 34.9 & 65.8 & 23.9 & 53.9\\
        PC-I-FGSM~\cite{wan2023adversarial} & ~~99.8 & 59.1 & 53.8 & 36.7 & 15.2 & 21.4 & 26.5 & 28.4 & 40.7 & 17.9 & 27.8 & 57.0 & 14.0 & 38.3\\
        IE-FGSM~\cite{peng2023boosting} & 100.0 & 67.3 & 63.0 & 43.8 & 17.4 & 29.0 & 37.2 & 36.7 & 40.6 & 24.6 & 27.7 & 59.1 & 13.3 & 43.1\\
        \rowcolor{Gray}GRA~\cite{zhu2023boosting} & ~~97.5 & 84.6 & 83.5 & 81.9 & 46.3 & 61.9 & 72.9 & 70.9 & 49.7 & 73.6 & 43.1 & 75.4 & 41.3 & 67.9\\
        GNP~\cite{wu2023gnp} & 100.0 & 68.9 & 63.8 & 45.8 & 16.9 & 28.9 & 38.3 & 36.4 & 41.2 & 25.5 & 27.8 & 59.3 & 14.1 & 43.6\\
        MIG~\cite{ma2023transferable} & 100.0 & 69.0 & 63.7 & 52.1 & 24.8 & 36.5 & 48.4 & 44.3 & 41.2 & 35.5 & 28.7 & 60.6 & 18.6 & 47.9\\
        DTA~\cite{yang2023improving} & 100.0 & 64.2 & 58.8 & 40.3 & 16.1 & 26.3 & 35.0 & 33.6 & 40.7 & 22.3 & 27.5 & 57.9 & 13.4 & 41.2\\
          \rowcolor{Gray}PGN~\cite{ge2023boosting} & ~~98.7 & 88.7 & 86.3 & 85.5 & 50.1 & 68.7 & 76.9 & 73.6 & 49.0 & 75.2 & 42.3 & 78.0 & 44.5 & 70.6\\
          ANDA~\cite{fang2024strong} & ~~99.9 & 84.5 & 81.6 & 72.4 & 41.4 & 60.7 & 71.3 & 66.8 & 43.1 & 65.5 & 30.9 & 65.1 & 28.3 & 62.4\\
        GI-FGSM~\cite{wang2022boosting} & 100.0 &72.6 & 65.4 & 49.1 & 18.9 & 29.5 & 37.5 & 36.3 & 40.8 & 25.1 & 28.1 & 61.7 & 16.6 & 44.7\\

        AdaMSI-FGM~\cite{long2024convergence} &100.0&64.6&58.5&39.7&15.1&24.9&30.0&31.0&40.6&15.7&27.5&58.0&13.4 & 39.9\\ 
        \rowcolor{Gray}FGSRA~\cite{wang2024fgsra} & ~~97.9 & 89.7 & 89.6 & 86.2 & 45.7 & 67.8 & 75.9 & 75.9 & 46.6 & 72.5 & 36.3 & 77.4 & 32.2 & 68.8 \\
        
        Foolmix~\cite{li2024foolmix} & ~~99.1 & 74.2 & 70.0 & 61.7 & 30.8 & 43.6 & 55.0 & 51.8 & 42.0 & 46.3 & 30.0 & 62.9 & 22.5 & 53.1 \\
          \rowcolor{Gray}MEF~\cite{qiu2025mef} & ~~99.3 & 95.3 & 94.1 & 91.4 & 68.4 & 80.8 & 88.7 & 88.3 & 47.6 & 86.7 & 41.2 & 81.6 & 44.3 & 77.5\\
        \rowcolor{Gray}GAA~\cite{gan2025boosting} & ~~95.5 & 84.6 & 82.5 & 81.8 & 43.3 & 59.9 & 70.7 & 66.3 & 49.3 & 71.1 & 43.5 & 75.9 & 40.9 & 66.6 \\

        MUMODIG~\cite{ren2025improving} & ~~98.6 & 87.4 & 85.0 & 79.9 & 45.4 & 65.6 & 75.8 & 71.7 & 43.4 & 69.7 & 31.7 & 67.0 & 27.0 & 65.2\\
        \noalign{\vskip 0.1ex}
        \bottomrule
    \end{tabular}
    }
    \vspace{-.5em}
\end{table*}

\Review{wxs}{\textbf{Prediction-Correction based I-FGSM (PC-I-FGSM)}~\cite{wan2023adversarial} first adopts an existing attack method to produce a predicted example, which will be combined with the current example to update adversarial perturbation.
}

\Review{lyy, wxs}{\textbf{Iterative enhanced Euler's FGSM (IE-FGSM)~}\cite{peng2023boosting} adopts the average gradient of the present adversarial example and the anticipatory data point in the direction of the current gradient to stabilize the update direction.
}

\Review{wxs}{\textbf{Gradient Relevance Attack (GRA)}~\cite{zhu2023boosting} corrects the gradient using the average gradient of several data points sampled in the neighborhood, and adopts a decay indicator to penalize the fluctuation of signs of gradient among two adjacent iterations.
} 

\Review{lyy,wxs}{\textbf{Gradient Norm
Penalty (GNP)} attack~\cite{wu2023gnp} introduces gradient norm to find local maxima with small gradient norm, which is approximated with finite difference method.
}

\Review{zzl, wxs}{\textbf{Momentum Integrated Gradients (MIG)}}~\cite{ma2023transferable} utilizes the integrated gradients~\cite{sundararajan2017axiomatic} instead of vanilla gradients to accumulate the momentum for crafting adversarial examples. 

\Review{zzl,wxs}{\textbf{Direction Tuning Attack (DTA)}~\cite{yang2023improving} calculates the gradient on several examples updated with small stepsize to diminish the discrepancy  between the actual update direction and the steepest update direction.
}

\Review{wxs}{\textbf{Penalizing Gradient Norm (PGN)} attack~\cite{ge2023boosting} approximates the Hessian matrix using the finite difference method to efficiently lead the adversaries towards flat local optima.}


\Review{wxs}{\textbf{Asymptotically Normal Distribution Attacks (ANDA)} \cite{fang2024strong} approximates the posterior distribution over the perturbations by utilizing the asymptotic normality property of stochastic gradient ascent with ensemble strategy.}

\Review{wxs}{\textbf{Global momentum Initialization Method (GI-FGSM)}~\cite{wang2022boosting} adopts a large stepsize for several iterations to initialize the momentum, which helps escape local optima.}

\Review{zff,wxs}{\textbf{Adaptive Momentum and Step-size Iterate Fast Gradient Method (AdaMSI-FGM)}\cite{long2024convergence} incorporates a non-monotonic adaptive momentum parameter and replaces the problematic sign operation with an adaptive step-size scheme to overcome the non-convergence issue.}

\Review{lbh,wxs}{\textbf{Frequency-Guided Sample Relevance Attack (FGSRA)} \cite{wang2024fgsra} adopts the hybrid gradient of the current sample and samples in the frequency neighborhood at each iteration to escape the model-sensitive region. }


\Review{wsk,wxs}{\textbf{FoolMix}\cite{li2024foolmix} blends the image with a set of random pixel-blocks to calculate the gradient and further blends the gradient \wrt various random labels.}

\Review{lbh,wxs}{\textbf{Maximin Excepted Flatness (MEF)}}~\cite{qiu2025mef} constructs a max-min bi-level optimization problem, adopts neighborhood conditional sampling to approximate worst-case neighborhoods and reuses outer gradients for inner updates. 


\Review{wsk,wxs}{\textbf{Gradient Aggregation Attack (GAA)}~\cite{gan2025boosting} aggregates the gradient of neighbors by simultaneously constraining empirical loss, worst-aware loss, and substitute gap loss.}

\Review{zrx,wxs}{\textbf{Multiple Monotonic Integrated Gradients (MuMoDIG)} \cite{ren2025improving} extends MIG~\cite{ma2023transferable} by generating integration paths from multiple baseline samples and enforcing the monotonicity of each path through a Lower-Bound Quantization mechanism.}

\textbf{Evaluations.} As shown in Tab.~\ref{tab:untargeted:gradient-based-attack}, I-FGSM exhibits overfitting on the surrogate model and results in the lowest transferability. MI-FGSM mitigates this overfitting by incorporating momentum into I-FGSM. VMI-FGSM further improves transferability by introducing variance tuning to stabilize updates. Notably, several subsequently proposed gradient-based attacks do not outperform VMI-FGSM, \eg AI-FGTM, AdaMSI-FGM, underscoring the issue of unfair comparisons in transfer-based attacks. This phenomenon is also observed in other categories of transfer-based attacks, which warrants increased attention. Currently, MEF achieves the best ASR by generating adversarial examples at flat local minima using neighborhood conditional sampling while PGN achieves the runner-up performance by approximating the Hessian matrix.

\begin{tcolorbox}[
    colframe=black,
    colback=white,
    left=0.1cm,
    right=0.1cm,
    top=0.1cm,
    bottom=0.1cm,
    boxrule=0.8pt,
    enhanced,
    drop fuzzy shadow,
    opacityback=0.95,
]
\textbf{Takeaways.}

\CirNum{1} Momentum stabilizes the update procedure and has been a widely adopted strategy for transfer-based attacks.

\CirNum{2} The convergence rate of the optimization method is positively correlated to the transferability~\cite{lin2020nesterov, wang2021boosting,zou2022making}.

\CirNum{3} It is beneficial to adopt the gradient of several data points in the neighborhood of current adversarial examples during the optimization procedure~\cite{wang2021enhancing,wang2021boosting,zhu2023boosting,wang2024fgsra}.


\CirNum{4} Generating the adversarial examples in flat local minima~\cite{wu2023gnp, ge2023boosting,qiu2025mef} tends to enhance transferability.
\end{tcolorbox}

\begin{table*}[tb]
    \centering
    \caption{Attack success rates of various input transformation-based adversarial attacks where the adversarial examples are generated on RN50. We have highlighted the approaches that achieve the top-5 attack performance in \colorbox{gray!30}{gray}.}
    \label{tab:untargeted:input-transformation}
    \vspace{-.5em}
    \resizebox{\textwidth}{!}{
    \begin{tabular}{l|cccccccc|ccccc|c}
        \toprule
        \multicolumn{1}{c}{\multirow{2}{*}{Attacks}} & \multicolumn{8}{c}{Standardly trained models} & \multicolumn{5}{c}{Defenses} & \multicolumn{1}{c}{\multirow{2}{*}{Average}} \\
        \cmidrule(lr){2-9} \cmidrule(lr){10-14}
        & RN-50 & VGG-16 & MN-v2 & Inc-v3 & ViT & PiT & Visformer & Swin & AT & HGD& RS & NRP & DiffPure \\
        \midrule\midrule
        DIM~\cite{xie2019improving} & ~~98.7 & 71.0 & 66.2 & 57.1 & 27.5 & 39.7 & 49.5 & 45.3 & 41.4 & 42.0 & 28.8 & 58.3 & 19.9 & 49.6\\
        TIM~\cite{dong2019evading} & ~~97.8 & 57.9 & 46.9 & 38.9 & 15.3 & 16.5 & 23.2 & 19.0 & 41.7 & 25.4 & 32.5 & 56.6 & 32.3 & 38.8\\
        SIM~\cite{lin2020nesterov} & 100.0 & 70.2 & 64.4 & 52.1 & 24.5 & 36.9 & 48.1 & 43.5 & 40.8 & 35.6 & 28.3 & 60.3 & 17.7 & 47.9\\
        DEM~\cite{zou2020improving} & ~~99.9 & 93.1 & 89.3 & 91.0 & 48.2 & 63.5 & 78.2 & 71.4 & 45.5 & 74.8 & 36.0 & 59.3 & 35.6& 68.1\\
        Admix~\cite{wang2021admix} & 100.0 & 79.9 & 77.7 & 67.7 & 32.5 & 49.3 & 62.5 & 58.2 & 41.8 & 50.7 & 30.1 & 65.1 & 20.7 & 56.6\\
        ATTA~\cite{wu2021improving} & ~~99.9 & 59.7 & 52.0 & 39.9 & 16.9 & 27.0 & 33.6 & 33.6 & 40.7 & 20.5 & 27.9 & 57.3 & 13.9& 40.2 \\

        SSM~\cite{long2022frequency} & ~~98.0 & 88.8 & 86.4 & 83.1 & 50.7 & 68.3 & 76.3 & 75.7 & 46.0 & 72.9 & 36.5 & 75.3 & 35.0 & 68.7\\
          AITL~\cite{yuan2022adaptive} & ~~94.5 & 87.2 & 84.9 & 82.8 & 52.7 & 68.0 & 75.3 & 72.7 & 45.8 & 77.2 & 36.9 & 67.6 & 40.4 & 68.1\\
        PAM~\cite{zhang2023improving} & 100.0 & 81.3 & 77.0 & 73.3 & 27.1 & 42.1 & 58.2 & 53.2 & 43.0 & 51.2 & 30.2 & 69.0 & 19.9 & 55.8\\
        LPM~\cite{Wei2023BoostingAT} & ~~98.6 & 61.5 & 53.2 & 39.2 & 15.2 & 24.3 & 31.5 & 31.8 & 40.3 & 19.6 & 27.7 & 58.5 & 14.3 & 39.7\\
          SIA~\cite{wang2023structure} & ~~99.4 & 94.3 & 92.7 & 83.1 & 50.0 & 76.7 & 86.5 & 83.6 & 44.0 & 79.5 & 32.1 & 74.9 & 27.8 & 71.1\\
        \rowcolor{Gray}STM~\cite{ge2023improving} & ~~97.5 & 90.2 & 88.9 & 88.3 & 56.7 & 73.0 & 80.3 & 77.5 & 49.0 & 82.7 & 41.8 & 79.3 & 43.6 & 73.0 \\
        US-MM~\cite{wang2023boost} & ~~99.7 & 90.1 & 88.4 & 78.8 & 37.5 & 57.5 & 72.4 & 70.1 & 44.0 & 62.8 & 32.2 & 74.8 & 22.3 & 63.9\\
          \rowcolor{Gray}DeCoWA~\cite{lin2024boosting} & ~~92.6 & 98.7 & 97.7 & 94.2 & 59.8 & 63.9 & 82.6 & 75.6 & 50.5 & 90.1 & 43.1 & 77.5 & 41.3 & 74.4\\
          \rowcolor{Gray}L2T~\cite{zhu2024learning} & ~~99.5 & 95.1 & 94.2 & 90.7 & 63.2 & 80.6 & 88.0 & 86.4 & 48.6 & 85.5 & 39.6 & 80.2 & 42.1 & 76.4\\
          \rowcolor{Gray}BSR~\cite{wang2024boosting} & ~~99.0 & 96.8 & 95.6 & 90.8 & 54.3 & 79.9 & 89.3 & 84.7 & 44.8 & 84.7 & 33.2 & 76.3 & 32.4 & 74.0\\
        MFI~\cite{qian2024enhancing} & ~~95.2 & 84.6 & 83.3 & 83.3 & 45.0 & 62.2 & 72.4 & 68.4 & 49.4 & 74.9 & 44.7 & 72.1 & 46.5 & 67.8\\
                MaskBlock~\cite{fan2025maskblock} & 100.0 & 64.3 & 59.6 & 41.9 & 17.0 & 26.8 & 35.0 & 34.5 & 41.0 & 21.5 & 28.0 & 58.9 & 14.4 & 41.8\\
        \rowcolor{Gray}OPS~\cite{guo2025OPS} & ~~99.8 & 98.8 & 98.6 & 98.7 & 90.6 & 95.3 & 97.1 & 96.3 & 60.9 & 98.1 & 69.4 & 92.5 & 88.9 & 91.2\\ 
        \noalign{\vskip 0.1ex}
        \bottomrule
    \end{tabular}
    }
    \vspace{-.5em}
    
\end{table*}

\subsection{Input Transformation-based Attack}

Input transformation-based attacks calculate the gradient over transformed images to boost the transferability.

\Review{wxs}{\textbf{Diverse Inputs Method (DIM)}~\cite{xie2019improving} adds padding to a randomly resized image to obtain an image with a fixed size, which are used for gradient calculation to generate more transferable adversarial examples.}

\Review{wxs}{\textbf{Translation Invariant Method (TIM)}~\cite{dong2019evading} optimizes the perturbation on several translated images. To accelerate the process, TIM uses a Gaussian kernel to smooth the gradient before update, which performs well on defense models.}

\Review{wxs}{\textbf{Scale Invariant Method (SIM)}~\cite{lin2020nesterov} calculates the average gradient of several scaled images with various scale factors.}

\Review{wxs}{\textbf{Diversity Ensemble Method (DEM)}~\cite{zou2020improving} calculates the average gradient on several diverse transformed images similar to DIM~\cite{xie2019improving} with different resizing factors.}

\Review{wxs}{\textbf{Admix}~\cite{wang2021admix} randomly combines the input image with an image from other categories to construct a new virtual sample for gradient calculation, in which the add-in image only occupies a small portion of the virtual sample.}

\Review{wxs}{\textbf{Adversarial Transformation-enhanced Transfer Attack (ATTA)}~\cite{wu2021improving} trains a CNN-based adversarial transformation network to model the most harmful transformation that eliminates the adversarial perturbation, which is further adopted in the gradient calculation for more transferable adversaries.}

\Review{wxs}{\textbf{Spectrum Simulation Method (SSM)}~\cite{long2022frequency} randomly scales the input image with Gaussian noise in the frequency domain using DCT and converts it back into the spatial domain using IDCT for gradient calculation.}

\Review{zzl,wxs}{\textbf{Adaptive Image Transformation Learner (AITL)}}~\cite{yuan2022adaptive} trains a neural network to predict the most effective combination of input transformations based on the input image.

\Review{lyy, wxs}{
\textbf{Path-Augmented Method (PAM)}~\cite{zhang2023improving} selects multiple augmentation paths based on their average transferability and trains a semantic predictor on a validation set. During attack generation, PAM estimates a semantic ratio via the predictor and linearly combines the adversarial example with selected base images for gradient accumulation.
}

\Review{yzy,wxs}{
\textbf{Learnable Patch-wise Mask (LPM)}~\cite{Wei2023BoostingAT} adopts differential evolution to craft a mask for the input image, which drops the model-specific discriminative regions to avoid overfitting. 
}

\Review{wxs}{\textbf{Structure Invariant Attack (SIA)}~\cite{wang2023structure} randomly splits the input image into several blocks and applies various input transformations onto each block to improve the diversity of transformed image while preserving the structure of image.}

\Review{gzj,wxs}{\textbf{Style Transfer Method (STM)}~\cite{ge2023improving} mixes the input with the stylized one using an arbitrary style transfer network to augment the diversity of transformed images.
}


\Review{zzl,wxs}{\textbf{Uniform Scale and Mix Mask Method (US-MM)}}~\cite{wang2023boost} uniformly scales the image with boundaries and adopts an element-wise product as the mix-up operation to introduce features of other categories.




\Review{zzl,wxs}{\textbf{Deformation-Constrained Warping Attack (DeCoWA)}}~\cite{lin2024boosting} employs constrained elastic deformation~\cite{xu2023comprehensive} to augment the inputs with enhanced local details, thus better simulating the various interested regions of different models.

\Review{zzl,wxs}{\textbf{Learn to Transform (L2T)}}~\cite{zhu2024learning} leverages the auto-augmentation strategy to learn the optimal combination of transformations at each iteration,  efficiently maximizing the input diversity to improve the adversarial transferability.  

\Review{wxs}{\textbf{Block Shuffle and Rotation (BSR)}~\cite{wang2024boosting} randomly shuffles and rotates the image blocks for gradient calculation to strengthen the consistency of attention heatmaps on various models, which effectively boosts the transferability.}

\Review{lyy,wxs}{\textbf{Mixed-Frequency Input (MFI)}~\cite{qian2024enhancing} injects high-frequency components from other images into the input during gradient computation to avoid overfitting surrogate model.}

\Review{zzl,wxs}{
\textbf{Mask Block Attack (MaskBlock)}~\cite{fan2025maskblock} splits the image into several blocks and iteratively masks each diagonal block to accumulate the gradient for perturbation update.
}

        \Review{gzj,wxs}{\textbf{Operator-Perturbation-based Stochastic optimization (OPS)}~\cite{guo2025OPS} applies transformation operators to the surrogate model with random perturbation for the adversarial examples.}

\textbf{Evaluations.} As shown in Tab.~\ref{tab:untargeted:input-transformation}, input transformation-based attacks substantially enhance the transferability of MI-FGSM. DIM, the first method in this category, yields a significant gain, while its ensemble variant DEM further enhances transferability. Despite using multiple transformed images, some input transformation-based attacks do not outperform DEM. A similar trend is observed for gradient-based attacks due to ignoring some related baselines. It is also noted that the attack performance on standardly trained models may differ from defenses like TIM, underscoring the importance of employing diverse models with various architectures and defense mechanisms. Recent advancements, such as more adaptive transformations (\eg, AITL) and transformations on local structures (\eg, DeCoWA, BSR), achieve top-tier attack transferability and may offer potential for further improvement. Moreover, input transformation-based attacks often outperform gradient-based attacks. However, several input transformation-based attacks are compared with gradient-based attacks for evaluations, leading to an unfair comparison. 
\begin{tcolorbox}[
    colframe=black,
    colback=white,
    left=0.1cm,
    right=0.1cm,
    top=0.1cm,
    bottom=0.1cm,
    boxrule=0.8pt,
    enhanced,
    drop fuzzy shadow,
    opacityback=0.95,
]
\textbf{Takeaways.}

\CirNum{1} Transforming the input image results in more diverse input image, which significantly boosts transferability using various transformations, \eg, resizing and padding~\cite{xie2019improving}, translation~\cite{dong2019evading}, scale~\cite{lin2020nesterov}, masking~\cite{fan2025maskblock}, \etc.

\CirNum{2} Mixing the input image with the image from other categories~\cite{wang2021admix} or with different styles~\cite{ge2023improving} results in more transferable adversarial examples.

\CirNum{3} Transforming the images introduces variance in the gradient, and accumulating the gradient of multiple transformed images~\cite{wang2023structure,wang2024boosting,guo2025OPS} mitigates this variance.

\CirNum{4} Instead of applying global transformations to the image, local transformations yield further improvement~\cite{lin2024boosting}.

\CirNum{5} Transforming the images using combinations of various augmentations boosts the transferability~\cite{yuan2022adaptive,zhu2024learning,guo2025OPS}. 

\end{tcolorbox}

\subsection{Advanced Objective Function}
In addition to the widely adopted cross-entropy loss for the attack, recent works have developed advanced objective functions to generate more transferable adversarial examples.

\Review{wxs}{\textbf{Transferable Adversarial Perturbations (TAP)}~\cite{zhou2018transferable} adopts two regularizers: 1) maximizing the difference of the feature maps for all layers between the benign sample and adversarial example; 2) maximizing the perturbation convolved by a pre-defined kernel to improve its smoothness.}


\Review{wxs}{\textbf{Intermediate Level Attack (ILA)}~\cite{huang2019enhancing} finetunes an adversarial example by enlarging the feature difference between the adversarial and benign example at a given layer.}


\Review{wxs}{\textbf{Attention-guided Transfer Attack (ATA)}~\cite{wu2020boosting} introduces a regularizer on the difference of the Grad-CAM~\cite{selvaraju2017grad} attention maps between the benign sample and adversarial example.}

\Review{yzy,wxs}{\textbf{Yet Another Intermediate-Level Attack (YAILA)}~\cite{li2020yet} models a linear relationship between intermediate-level feature discrepancies and the classification loss via a weight matrix, which is estimated from the feature difference between benign and adversarial samples at each iteration of a baseline attack.
}

\begin{table*}[tb]
    \centering
    \caption{Attack success rates of various advanced objective functions where the adversarial examples are generated on RN50. We have highlighted the approaches that achieve the top-5 attack performance in \colorbox{gray!30}{gray}.}
    \label{tab:untargeted:advanced_objectve_function}
    \vspace{-.5em}
    \resizebox{\textwidth}{!}{
    \begin{tabular}{l|cccccccc|ccccc|c}
        \toprule
        \multicolumn{1}{c}{\multirow{2}{*}{Attacks}} & \multicolumn{8}{c}{Standardly trained models} & \multicolumn{5}{c}{Defenses} & \multicolumn{1}{c}{\multirow{2}{*}{Average}} \\
        \cmidrule(lr){2-9} \cmidrule(lr){10-14}
        & RN-50 & VGG-16 & MN-v2 & Inc-v3 & ViT & PiT & Visformer & Swin & AT & HGD& RS & NRP & DiffPure\\
        \midrule\midrule
        TAP~\cite{zhou2018transferable} & ~~99.9 & 93.4 & 93.8 & 64.8 & 15.2 & 25.4 & 44.3 & 40.2 & 41.7 & 34.5 & 27.3 & 65.8 & 15.5 & 47.9\\
        ILA~\cite{huang2019enhancing} & ~~98.7 & 68.7 & 64.6 & 33.4 & 12.8 & 23.5 & 31.9 & 32.1 & 40.0 & 16.2 & 27.1 & 52.5 & 11.3 & 37.7\\
        ATA~\cite{wu2020boosting} & ~~99.8 & 35.8 & 35.1 & 19.2 & ~~7.6 & 11.9 & 14.9 & 15.0 & 39.6 & ~~6.2 & 26.4 & 43.0 & ~~9.4 & 22.7\\
        YAILA ~\cite{li2020yet} & ~~99.1 & 63.6 & 59.4 & 21.7 & ~~8.3 & 14.4 & 23.2 & 20.7 & 39.3 & 9.3 & 26.4 & 45.6 & ~~8.4 & 36.8\\
        FIA~\cite{wang2021feature} & ~~98.0 & 71.2 & 65.8 & 40.2 & 12.6 & 19.9 & 33.2 & 33.1 & 41.1 & 19.2 & 28.0 & 58.4 & 12.1 & 38.6\\
        IR~\cite{wang2021unified} & 100.0 & 59.4 & 53.4 & 36.2 & 14.8 & 22.9 & 27.7 & 28.1 & 40.6 & 18.6 & 27.3 & 57.5 & 13.2 & 36.9\\
        \rowcolor{Gray} TRAP~\cite{wang2021exploring} & ~~99.1 & 96.1 & 93.5 & 84.7 & 33.6 & 47.7 & 64.1 & 58.6 & 41.1 & 64.3 & 27.2 & 65.3 & 16.7 & 59.1\\
        TAIG~\cite{huang2022integtransferable} & ~~99.8 & 48.3 & 46.4 & 30.4 & 12.1 & 20.8 & 26.1 & 26.0 & 39.8 & 14.6 & 26.6 & 47.4 & ~~9.7 & 31.1\\
        FMAA~\cite{he2022enhancing} & ~~98.3 & 80.1 & 77.3 & 55.8 & 18.0 & 33.0 & 50.8 & 52.7 & 42.5 & 34.4 & 29.4 & 62.2 & 13.4 & 48.6\\
        NAA~\cite{zhang2022improvinga} & ~~84.7 & 61.6 & 58.6 & 38.2 & 19.5 & 29.6 & 36.6 & 39.0 & 40.3 & 25.8 & 28.0 & 52.7 & 12.6 & 40.5\\
        \rowcolor{Gray} RPA~\cite{zhang2022enhancing} & ~~95.1 & 85.8 & 86.0 & 76.3 & 35.9 & 47.5 & 63.4 & 62.1 & 45.0 & 58.3 & 33.2 & 70.8 & 23.6  & 59.5\\
        Fuzziness\_Tuned~\cite{yang2023fuzziness} & 100.0 & 57.3 & 51.8 & 33.7 & 14.1 & 21.7 & 26.7 & 26.6 & 40.3 & 16.4 & 27.1 & 57.1 & 12.5 & 36.5\\
        DANAA~\cite{jin2023danaa} & ~~96.8 & 86.3 & 82.5 & 69.9 & 23.4 & 36.0 & 54.3 & 51.6 & 43.4 & 51.4 & 31.0 & 71.8 & 17.6 & 56.6\\
        \rowcolor{Gray} ILPD~\cite{li2023improving} & ~~97.9 & 87.4 & 85.8 & 72.2 & 44.6 & 60.7 & 69.3 & 67.6 & 43.8 & 56.8 & 32.5 & 66.3 & 28.1 & 63.3\\
        \rowcolor{Gray} BFA~\cite{wang2024improving} & ~~98.4 & 94.1 & 92.3 & 84.8 & 52.7 & 73.2 & 85.8 & 82.3 & 44.6 & 78.0 & 33.0 & 79.7 & 26.0 & 71.6\\
        \rowcolor{Gray} P2FA~\cite{liupixel2feature} & 100.0 & 97.9 & 97.6 & 84.0 & 43.7 & 64.9 & 86.0 & 83.2 & 43.9 & 66.3 & 32.2 & 79.5 & 22.2 & 69.3\\
    MFAA~\cite{zheng2025enhancing} & ~~95.2 & 86.5 & 83.5 & 72.2 & 24.4 & 42.9 & 59.6 & 56.7 & 42.9 & 48.1 & 30.6 & 51.5 & 17.9 & 54.8\\
    \noalign{\vskip 0.1ex}
        \bottomrule
    \end{tabular}
    }
    \vspace{-.5em}
\end{table*}
\Review{wxs}{\textbf{Feature Importance-aware Attack (FIA)}~\cite{wang2021feature} minimizes a weighted feature map in the intermediate layer to disrupt the significant object-aware features, which is the average gradient \wrt the feature of several randomly masked input images.}

\Review{wky,wxs}{\textbf{Interaction-Reduced Attack (IR)}~\cite{wang2021unified} finds that negative interactions~\cite{michel1999interaction} inside adversarial perturbations result in better transferability and introduces the interaction regularizer into the objective function to minimize the interaction.
}


\Review{lyy,wxs}{\textbf{TRAP}~\cite{wang2021exploring} initializes intermediate-layer features via MI-FGSM\cite{dong2018boosting}, minimizes the cosine similarity and relative magnitude between benign and adversarial features with accumulated perturbations, and employs random geometric transformations during gradient computation.

}

\Review{wxs}{\textbf{Transferable
Attack based on Integrated Gradients (TAIG)}~\cite{huang2022integtransferable} updates the adversarial perturbation using the direction of the integrated gradients~\cite{sundararajan2017axiomatic}.}

\Review{wxs}{\textbf{Feature Momentum Adversarial Attack (FMAA)}~\cite{he2022enhancing} iteratively updates the weight in FIA~\cite{wang2021feature} using momentum with the same aggregation strategy, which describes more precise category-related feature for better transferability.}

\Review{wxs}{\textbf{Neuron Attribution-Based Attack (NAA)}~\cite{zhang2022improvinga} reformulates the output of the model with path integral and computes the feature importance of each neuron in the intermediate layer with decomposition on integral, which suppresses features by minimizing weighted feature importance.}

\Review{wxs}{\textbf{Random Patch Attack (RPA)}~\cite{zhang2022enhancing} computes the average gradient \wrt the feature of patch-wise masked images with various patch sizes, which acts as the weight in FIA~\cite{wang2021feature} to highlight the important intrinsic object-related features.}


\Review{wxs}{\textbf{Fuzziness-tuned}~\cite{yang2023fuzziness} scales non-ground-truth confidences and applies temperature scaling to the logits.}

\Review{lyy, wxs}{\textbf{Double Adversarial Neuron Attribution Attack (DANAA)}~\cite{jin2023danaa} utilizes a non-linear path to calculate the feature importance for each neuron as in NAA~\cite{zhang2022improvinga}}.

\Review{yzy, wxs}{\textbf{Intermediate-Level Perturbation Decay (ILPD)}~\cite{li2023improving} decays the intermediate-level perturbation of benign features by mixing the features of benign and adversarial examples.
}

\Review{lyy,wxs}{\textbf{Blackbox Feature-driven Attack (BFA)}~\cite{wang2024improving} calculates the weight matrix using the intermediate samples during the I-FGM attack process to suppress white-box features and strengthen black-box ones, and modifies the FIA's loss to distinguish positive and negative factors.}

\Review{zff,wxs}{\textbf{Pixel2Feature Attack (P2FA)}~\cite{liupixel2feature} perturbs the feature of adversarial pixels multiple times guided by the feature importance, then inverts the perturbed features to the pixels.}

\Review{lyy,wxs}{\textbf{Multi-Feature Attention Attack (MFAA)}~\cite{zheng2025enhancing} recursively aggregates gradients from deeper to shallower layers to form guidance maps for ensemble attention, which is iteratively disturbed to enhance transerability.}



\textbf{Evaluations.} 
As shown in Tab.~\ref{tab:untargeted:advanced_objectve_function}, incorporating advanced objective functions exhibits substantial variability in transferability, underscoring the pivotal influence of feature selection and optimization strategies (\eg, YAILA, NAA and DANAA). BFA attains state-of-the-art performance, markedly surpassing other approaches by differentiating between positive and negative feature factors, thereby suppressing white-box specific features while amplifying transferable ones. P2FA follows closely, primarily due to its effective perturbations and feature importance inversions. ILPD also performs well, as it decays intermediate-level perturbations by mixing features from benign and adversarial examples. Notably, refining feature weighting mechanisms consistently outperforms the baseline FIA. For example, RPA and FMAA substantially improve transferability using random patches and feature momentum, respectively, to better identify model agnostic regions.

\begin{tcolorbox}[
    colframe=black,
    colback=white,
    left=0.1cm,
    right=0.1cm,
    top=0.1cm,
    bottom=0.1cm,
    boxrule=0.8pt,
    enhanced,
    drop fuzzy shadow,
    opacityback=0.95,
]
\textbf{Takeaways.}

\CirNum{1} Interpretation-oriented metrics (\eg, attention maps) can be used as objective functions for better transferability.

\CirNum{2} Feature-level metrics between benign samples and adversarial examples can act as effective objective functions.

\CirNum{3} The weight matrix in FIA~\cite{wang2021feature} is crucial for enhancing adversarial transferability, leading to significant interest in developing more effective weight matrices.

\CirNum{4} Not all features contribute equally to the decision of the models. Identifying and attacking the most critical, model-agnostic features is the key to improving the transferability.
\end{tcolorbox}

\subsection{Generation-based Attack}
\Review{yzy,wxs}{\textbf{Cross Domain Attack (CDTP)}~\cite{naseer2019cross} trains a generator with a pre-trained discriminator using a relativistic loss on the benign sample and generated adversarial example. The generator is trained in the source domain (\eg, paintings, cartoon images) to craft unbounded perturbation followed by a projection for bounded perturbation, which can generate transferable perturbation in the target domain (\eg, ImageNet).
}

\Review{yzy,wxs}{
\textbf{Learning Transferable Adversarial Perturbations (LTP)} \cite{kanth2021learning} trains a generator to craft unbounded perturbation, which maximizes the distance between the normal and adversarial feature map of a given classifier.
}

\Review{wky,wxs}{\textbf{Attentive-Diversity Attack (ADA)}~\cite{kim2022diverse} maximizes the classification loss on the surrogate classifier, distracts the attention maps between adversarial example and benign sample, and diversifies the attention maps of adversaries crafted by various latent codes.
}

\Review{wxs}{\textbf{GE-ADVGAN}~\cite{zhu2024geadvgan} conducts gradient editing using the sign of gradient on several frequency domain transformed adversaries to update the parameters of generator.}

\Review{gzj,wxs}{\textbf{DiffAttack}~\cite{chen2024diffattack} manipulate the latent space diffusion models to generate adversarial examples that fool both the surrogate model and diffusion models, but are similar to the self-attention maps of benign samples.}

\Review{wsk,wxs}{\textbf{Frequency-Aware Perturbation (FAP)}~\cite{wang2024boostingb} optimizes the adversarial perturbation on the most critical and model-agnostic frequency components.}

\Review{lbh,wxs}{\textbf{Dual Self-supervised ViT features Attack (DSVA)}~\cite{wu2025dsva} employs a joint cosine similarity loss between the global structural features extracted from DINO~\cite{caron2021dino} and local textual features obtained from MAE~\cite{he2022mae} of benign inputs and crafted adversarial examples to train the generator.}

\textbf{Evaluations.} As shown in Table \ref{tab:untargeted:generation}, generation-based attacks exhibit transferability across diverse models, yet their effectiveness varies with different generator optimization objectives. CDTP and LTP achieve the highest success rates on standard CNNs, showing the effectiveness of optimization based on domain information and feature separation. In comparison, ADA and GE-ADVGAN are less effective against strong architectural inductive biases (\eg, vision transformers), showing that perturbation diversity and gradient editing offer limited gains under such architectures. DiffAttack prioritizes perturbation imperceptibility via diffusion priors, but this reduces its cross-model success rates, especially on CNNs. DSVA achieves balanced results across architectures and defenses by aligning global and local self-supervised features.

\begin{table*}[tb]
    \centering
    
    \caption{Attack success rates of various generation-based attacks. We have highlighted the approaches that achieve the top-1 attack performance in \colorbox{gray!30}{gray}.}
    \vspace{-.5em}
    \resizebox{\textwidth}{!}{
    \label{tab:untargeted:generation}
    \begin{tabular}{l|cccccccc|ccccc|c}
        \toprule
        \multicolumn{1}{c}{\multirow{2}{*}{Attacks}} & \multicolumn{8}{c}{Standardly trained models} & \multicolumn{5}{c}{Defenses} & \multicolumn{1}{c}{\multirow{2}{*}{Average}} \\
        \cmidrule(lr){2-9} \cmidrule(lr){10-14}
        & RN-50 & VGG-16 & MN-v2 & Inc-v3 & ViT & PiT & Visformer & Swin & AT & HGD& RS & NRP & DiffPure \\
        \midrule\midrule
        
        CDTP~\cite{naseer2019cross} &  97.1 & 99.2 & 98.4 & 95.9 & 38.3 & 44.6 & 91.6 & 68.7 & 42.1 & 94.8 & 31.3 & 70.8 & 32.5 &69.64\\
        \rowcolor{Gray}LTP~\cite{kanth2021learning} & 99.6 & 99.8 & 98.4 & 98.8 & 53.5 & 69.7 & 98.4 & 94.1 & 41.1 & 97.3 & 29.4 & 69.8 & 26.0 &75.07 \\
        ADA~\cite{kim2022diverse} & 71.7 & 89.4 & 80.6 & 75.9 & 18.1 & 20.4 & 50.9 & 39.5 & 38.4 & 1.4 & 27.9 & 32.2 & 26.3 &44.05 \\
        GE-ADVGAN~\cite{zhu2024geadvgan} & 68.1 & 90.1 & 90.3 & 79.8 & 18.1 & 12.1 & 37.8 & 27.0 & 41.2 & 73.5 & 36.0 & 59.8 & 45.3 &52.24\\
        DiffAttack~\cite{chen2024diffattack} & 92.3 & 51.3 & 51.3 & 47.1 & 30.7 & 44.3 & 44.5 & 47.1 & 45.7 & 39.0 & 39.7 & 59.2 & 34.3 &48.19\\

        FAP~\cite{wang2024boostingb} &87.0&	68.4&	63.5 &	46.4&	16.0&	26.6&	32.4&	32.6&	41.8&	25.4&	28.6&	66.9&	21.4 & 42.85\\

        DSVA~\cite{wu2025dsva} & 68.0& 90.3& 90.5& 89.3& 56.3& 36.5& 66.1& 53.1& 42.5& 72.7& 33.2& 15.3 & 46.9 &58.52 \\
        \noalign{\vskip 0.1ex}
        \bottomrule
    \end{tabular}
    }
    \vspace{-.5em}
\end{table*}
\begin{tcolorbox}[
    colframe=black,
    colback=white,
    left=0.1cm,
    right=0.1cm,
    top=0.1cm,
    bottom=0.1cm,
    boxrule=0.8pt,
    enhanced,
    drop fuzzy shadow,
    opacityback=0.95,
]
\textbf{Takeaways.}

\CirNum{1} Certain input transformations (\eg, SSA~\cite{long2022frequency}) can serve as powerful data augmentation to train the generator (\eg, GE-ADVGAN~\cite{zhu2024geadvgan}).

\CirNum{2} Feature-level optimization for the generator, such as maximizing representation discrepancies (\eg, LTP~\cite{kanth2021learning}), remains highly effective for improving transferability.

\CirNum{3} Attacks on transformers benefit from semantic-aware perturbations. For instance, DSVA~\cite{wu2025dsva} benefits from combining global and local semantic features from self-supervised transformers.

\CirNum{4} Attacks that incorporate diffusion models or other pretrained generative priors tend to produce more natural-looking perturbations, thereby enhancing imperceptibility while maintaining attack efficacy.
\end{tcolorbox}

\subsection{Model-related Attack}

\Review{wxs}{\textbf{Skip Gradient Method (SGM)}~\cite{wu2020skip} utilizes more gradient from the skip connections in the residual blocks~\cite{he2016deep} to effectively boost the transferability of existing attacks.}

\Review{wxs}{\textbf{Linear backpropagation (LinBP)}~\cite{guo2020backpropagating} directly backpropagates all the gradients for ReLU in VGG~\cite{simonyan2015very}. While for the residual block $z_{i+1}= z_i + W_{i+1}^\top \sigma(W_i^\top z_i)$~\cite{he2016deep}, $\frac{\partial z_{i+1}}{\partial z_i}$ will be $1+\alpha_i W_i W_{i+1}$, where $\alpha_i$ is a scaled factor corresponding to the gradients w/wo ReLU.}

\begin{table*}[tb]
    \centering
    \caption{Attack success rates of various model-related attacks where the adversarial examples are generated on RN50/ViT. MTA, DSM, AGS, QAA and SETR adopt their trained models as the surrogate model. We have highlighted the approaches that achieve the top-5 attack performance in \colorbox{gray!30}{gray}.}
    \label{tab:untargeted:model-related}
    \vspace{-.5em}
    \resizebox{.96\textwidth}{!}{
    \begin{tabular}{l|cccccccc|ccccc|c}
        \toprule
        \multicolumn{1}{c}{\multirow{2}{*}{Attacks}} & \multicolumn{8}{c}{Standardly trained models} & \multicolumn{5}{c}{Defenses} & \multicolumn{1}{c}{\multirow{2}{*}{Average}} \\
    \cmidrule(lr){2-9} \cmidrule(lr){10-14}
        & RN-50 & VGG-16 & MN-v2 & Inc-v3 & ViT & PiT & Visformer & Swin & AT & HGD& RS & NRP & DiffPure\\
        \midrule\midrule
        SGM~\cite{wu2020skip} & ~100.0 & 73.2 & 75.7 & 45.9 & 18.9 & 33.5 & 41.1 & 41.9 & 41.3 & 25.3 & 28.6 & 64.3 & 15.0 & 46.5\\
        LinBP~\cite{guo2020backpropagating} & ~100.0 & 87.7 & 81.2 & 47.5 & 18.9 & 22.7 & 44.0 & 37.2 & 41.3 & 19.9 & 27.9 & 59.9 & 17.1 & 46.6\\
        \rowcolor{Gray}LLTA~\cite{fang2022learning} & ~~94.9 & 96.6 & 95.1 & 83.8 & 45.7 & 57.1 & 74.1 & 73.8 & 47.7 & 74.3 & 39.5 & 82.5 & 28.2 & 68.7\\
        IAA~\cite{zhu2021rethinking} & ~100.0 & 81.7 & 74.7 & 55.2 & 19.5 & 30.1 & 41.9 & 40.7 & 42.1 & 26.8 & 29.6 & 66.7 & 15.9 & 48.1\\
        \rowcolor{Gray}DRA~\cite{zhu2022towards} & ~~94.3 & 96.8 & 97.8 &95.0 & 75.3 & 77.7 & 88.7 & 86.4 & 70.8 & 93.2 & 85.7 & 86.4 & 75.3 & 86.4\\
        MTA~\cite{qin2023training} & ~~68.7 & 89.8 & 82.5 & 68.8 & 20.4 & 24.4 & 37.3 & 33.6 & 40.4 & 48.9 & 26.4 & 55.9 & 14.5 & 47.0\\
        MUP~\cite{yang2023generating} & ~~98.8 & 76.5 & 71.0 & 53.3 & 18.5 & 32.3 & 41.3 & 40.0 & 41.3 & 27.8 & 28.4 & 61.6 & 15.3 & 46.6\\
        DSM~\cite{yang2023boosting} & ~~82.5 & 96.5 & 93.0 & 74.9 & 28.4 & 38.4 & 59.0 & 55.8 & 45.0 & 58.6 & 33.7 & 71.6 & 20.4 & 58.3\\
        DHF~\cite{wang2023diversifying} & ~~99.9 & 74.4 & 70.1 & 51.8 & 25.3 & 40.8 & 48.5 & 47.4 & 41.4 & 35.6 & 29.0 & 62.2 & 16.1 & 49.4\\
        \rowcolor{Gray}BPA~\cite{wang2023rethinking} & ~~96.2 & 96.9 & 94.5 & 84.9 & 41.4 & 50.2 & 75.1 & 65.3 & 43.9 & 76.3 & 33.2 & 76.9 & 29.4 & 66.5\\
        AGS~\cite{wang2024ags} & ~~74.1 & 86.5 & 85.2 & 82.6 & 28.2 & 24.9 & 45.2 & 35.7 & 45.3 & 67.1 & 34.0 & 56.7 & 38.0 & 54.1\\
        CMF~\cite{weng2024exploring} & ~~98.2 & 85.3 & 74.7 & 79.0 & 37.5 & 46.7 & 63.6 & 52.3 & 45.3 & 65.0 & 35.8 & 68.7 & 36.8 & 60.7\\
        MA~\cite{ma2024improving} & ~~96.0 & 95.8 & 92.4 & 83.5 & 38.4 & 53.2 & 74.3 & 70.9 & 44.1 & 80.9 & 34.3 & 72.7 & 23.0 & 66.1\\
        QAA~\cite{yang2024quantization} & ~~76.3 & 90.7 & 87.7 & 72.9 & 18.1 & 28.0 & 42.6 & 39.8 & 43.4 & 31.6 & 30.2 & 27.7 & 19.0 & 46.8\\

        \rowcolor{Gray}AWT~\cite{chen2025enhancing} & ~~98.6 & 91.5 & 90.4 & 86.5 & 60.0 & 75.7 & 81.0 & 81.3 & 50.5 & 77.4 & 44.9 & 80.5 & 49.2 & 74.4\\
        FAUG~\cite{wang2025improving} &~~95.1 & 69.5 & 66.4 & 56.5 & 26.1 & 38.9 & 47.6 & 45.6 & 42.1 & 38.2 & 30.2 & 62.8 & 21.0 & 49.2\\
        ANA~\cite{li2025enhancing} &~~99.6 & 60.9 & 55.0 & 37.5 & 14.2 & 21.2 & 27.4 & 25.5 & 40.2 & 16.5 & 27.5 & 57.8 & 12.8 & 38.2\\
        \noalign{\vskip 0.1ex}\cdashline{1-15}\noalign{\vskip 0.1ex}
        PNAPA~\cite{wei2022towards} & ~~47.3 & 69.3 & 62.7 & 53.9 & 95.3 & 55.8 & 58.7 & 64.2 & 42.9 & 37.2 & 32.7 & 51.3 & 25.4 & 53.6\\
        SAPR~\cite{zhou2022improving} & ~~49.2 & 74.5 & 69.1 & 59.1 & 99.9 & 57.2 & 63.2 & 68.1 & 43.4 & 37.8 & 32.7 & 54.8 & 24.0 & 56.4\\
        SETR~\cite{naseer2022on} & ~~48.9 & 82.7 & 81.2 & 63.8 & 66.3 & 41.5 & 51.1 & 66.6 & 45.9 & 38.6 & 34.5 & 62.0 & 26.1 & 54.6\\
        ATA~\cite{wang2022generating} & ~~24.6 & 87.0 & 36.7 & 56.2 & 99.9 & 13.3 & 15.7 & ~~8.4 & 50.4 & 52.9 & 36.8 & 87.8 & 44.7 & 47.3\\
        TGR~\cite{zhang2023transferable} & ~~57.9 & 80.3 & 76.1 & 61.4 & 99.8 & 61.3 & 69.2 & 75.1 & 45.4 & 47.1 & 37.0 & 58.0 & 34.0 & 61.7\\
        VDC~\cite{zhang2024improving} & ~~51.6 & 74.2 & 70.7 & 59.1 & 100.0 & 64.0 & 68.3 & 73.8 & 44.7 & 41.6 & 35.1 & 57.9 & 29.6 & 59.3\\
        ATT~\cite{ming2024att} & ~~61.1 & 79.7 & 76.4 & 63.0 & 100.0 & 68.3 & 75.0 & 81.6 & 45.2 & 49.7 & 36.1 & 60.7 & 33.4 & 63.9\\
        FDAP~\cite{gao2024attacking} & ~~40.4& 66.4 & 58.2 & 49.8 & 99.3 & 42.0 & 46.2 & 51.1 & 44.0 & 27.9 & 30.4 & 49.4 & 24.3 & 48.4\\
        FPR~\cite{ren2025fpr} & ~~56.6 & 83.6 & 81.4 & 70.9 & 99.5 & 54.6 & 66.9 & 71.0 & 46.4 & 47.0 & 35.4 & 63.0 & 30.9 & 62.1\\
        \rowcolor{Gray}LL2S~\cite{liu2025ll2s} &~~76.8 & 90.1 & 91.3 & 79.1 & 99.3 & 80.6 & 84.1 & 92.7 & 44.5 & 70.7 & 40.8 & 75.2 & 57.4 & 75.6\\

        \bottomrule
    \end{tabular}
    }
    \vspace{-1.5em}
\end{table*}

\Review{wxs}{\textbf{Learning to Learn Transferable Attack (LLTA)}~\cite{fang2022learning} augments the data by random resizing and padding as well as the model using various learnable decay factors on the skip connection during back-propagation process.}


\Review{wxs}{\textbf{Intrinsic Adversarial Attack (IAA)}~\cite{zhu2021rethinking} substitutes ReLU with $\mathrm{SoftPlus}_\beta$ and scales the residual function with a factor $\lambda$ to enhance the alignment of adversarial attack and gradient of joint data distribution, where $\beta$ and $\lambda$ are optimized by Bayesian optimization.}

\Review{wky, wxs}{
\textbf{Distribution Relevant Attack (DRA)}~\cite{zhu2022towards} finetunes surrogate model by minimizing the classification loss with the distance between the gradient of log conditional density and that of log ground truth class-conditional data distribution.
}

\Review{gzj, wxs}{\textbf{Meta-Transfer Attack (MTA)}~\cite{qin2023training} trains a meta-surrogate model (MSM), in which the adversarial examples generated on such model can maximize the loss on a single or a set of pre-trained surrogate models.}

\Review{yzy, wxs}{\textbf{Masking Unimportant Parameters (MUP)}~\cite{yang2023generating} utilizes a Taylor expansion-based metric to measure the importance of a surrogate model's parameters and masks unimportant parameters to avoid adversarial examples' over-fitting.
}


\Review{yzy, wxs}{\textbf{Dark Surrogate Model (DSM)}~\cite{yang2023boosting} adopts a pretrained model as the teacher to train a surrogate model and utilizes CutMix~\cite{yun2019cutmix} on the input to enhance dark knowledge.
}


\Review{zzl, wxs}{\textbf{Diversifying the High-level Features (DHF)}~\cite{wang2023diversifying} randomly adjusts high-level features and mixes them with the features of benign samples when calculating the gradient.
}

\Review{lyy, wxs}{\textbf{Backward Propagation Attack (BPA)}~\cite{wang2023rethinking} recovers the truncated gradient of non-linear layers by adopting the derivative of SiLU for ReLU and Softmax function to calculate the gradient for max-pooling to make the gradient \wrt input image more related to loss function.}

\Review{lyy,wxs}{\textbf{Affordable and Generalizable Substitute training framework~(AGS)}~\cite{wang2024ags} trains the surrogate model using adversarial-centric contrastive learning and adversarial invariant learning to generate more transferable adversaries.}


\Review{wxs}{\textbf{Cross-frequency Meta-optimization Framework (CMF)} \cite{weng2024exploring} utilizes low-frequency feature mixing in the meta-train step to compute gradients averaged by adversarial feature mixing in the meta-test step.
}

\Review{lyy,wxs}{\textbf{Model Alignment (MA)}~\cite{ma2024improving} finetunes the surrogate model to minimize the alignment loss, which quantifies prediction divergence with an independently trained witness model.}

\Review{lyy,wxs}{\textbf{Quantization Aware Attack (QAA)}~\cite{yang2024quantization} simulates quantization in the substitute models with virtual quantization layers and stochastic rounding, and adds a quantization-aware gradient-alignment loss during the attack process.}

\Review{fz,wxs}{\textbf{Adversarial Weight Tuning (AWT)}~\cite{chen2025enhancing} adaptively adjusts the parameters of the surrogate model using adversaries to optimize flat local maxima and model smoothness.}

\Review{fz,wxs}{\textbf{Feature AUGmentation Attack (FAUG)}~\cite{wang2025improving} injects random noise into intermediate feature maps of the surrogate model to enlarge the diversity of the attack gradient.}

\Review{zff,wxs}{\textbf{Alignment Network Attack (ANA)}~\cite{li2025enhancing} utilizes masking operations and a lightweight alignment network to make surrogate models focus on critical regions of images.}

With the great success of vision transformers (ViTs)~\cite{dosovitskiy2020image,liu2021swin}, researchers have paid increasing attention to their robustness and proposed numerous model-related approaches to boost the adversarial transferability across various ViTs.

\Review{wxs}{\textbf{Pay No Attention and PatchOut Attack (PNAPA)}~\cite{wei2022towards} ignores the backpropagation through the attention branch and randomly drops several patches among the adversarial perturbation to boost the adversarial transferability.}

\Review{yzy, wxs}{
\textbf{Self-Attention Patches Restructure (SAPR)}~\cite{zhou2022improving} randomly permutes the input tokens before each attention layer to boost patch connection diversity and transferability.
}

\Review{wxs}{\textbf{Self-Ensemble \& Token Refinement (SETR)}~\cite{naseer2022on} dissects a single ViT into an ensemble of networks by adding a shared local norm followed by a refinement module as well as a shared classifier after each transformer block.}



\Review{lyy,wxs}{\textbf{Architecture-oriented Transferable Attacking framework (ATA)}~\cite{wang2022generating} activates uncertain attention and perturbs sensitive embeddings by focusing on patch-wise attentional regions and pixel-wise positions to confuse perception and disrupting embedded tokens across various transformers.}

\Review{yzy, wxs}{
\textbf{Token Gradient Regularization (TGR)}~\cite{zhang2023transferable} scales the gradient and masks the maximum or minimum gradient magnitude after the MLP block or attention.
}

\Review{gzj,wxs}{\textbf{Virtual Dense Connection (VDC)}~\cite{zhang2024improving} recomposes the original ViT by adding virtual dense connections that preserve the forward pass, while backpropagating gradients from deeper attention maps and MLP blocks through virtual connections.}

\Review{lbh,wxs}{\textbf{Adaptive Token Tuning (ATT)}~\cite{ming2024att} adaptively scales gradient in each model layer to reduce variance, dynamically discards patches using semantic guided mask and truncates token gradient to weaken the effectiveness of attention.}

\Review{zff,wxs}{\textbf{Feature Diversity Adversarial Perturbation attack (FDAP)} \cite{gao2024attacking} minimize feature diversity in the middle layers to accelerate natural tendency towards feature collapse.}

\Review{gzj,wxs}{\textbf{Forward Propagation Refinement (FPR)}~\cite{ren2025fpr} proposes Attention Map Diversification to diversify attention maps and attenuate gradients during the backward pass, and Momentum Token Embedding to utilize historical embeddings for stabilizing forward updates in Attention and MLP blocks. }

\Review{gzj,wxs}{\textbf{LL2S}~\cite{liu2025ll2s} exploits ViT's redundancy with attention sparsity, head permutation, clean-token regularization, ghost MoE, and robust-token learning, coordinated by online learning.}

\textbf{Evaluations.} As shown in Tab.~\ref{tab:untargeted:model-related}, by reshaping the surrogate’s gradient pathway, architecture, or training strategy, early modifications to the backpropagation process in CNNs (\eg, SGM and LinBP) already provide substantial improvements, while subsequent approaches further strengthen transferability through model augmentation or gradient and distribution alignment (\eg, LLTA, IAA, DRA, BPA). 
Also, surrogate training-oriented frameworks (\eg, MTA, DSM, AGS, MA)  show that enhancing the surrogate’s representational capacity and gradient generalization is crucial for transferability. 
For transformer-based models, ViT-specific designs that manipulate tokens, attention, or gradient flow (\eg, SAPR, SETR, ATA, TGR, VDC, ATT) achieve strong cross-ViT transferability and broaden model-related attacks beyond CNN paradigms. It is also noted that performance on standardly trained models does not always correlate with performance under defended settings, underscoring the importance for evaluation across diverse architectures and robust training regimes.

\begin{tcolorbox}[
colframe=black,
colback=white,
left=0.1cm,
right=0.1cm,
top=0.1cm,
bottom=0.1cm,
boxrule=0.8pt,
enhanced,
drop fuzzy shadow,
opacityback=0.95,
]
\textbf{Takeaways.}

\CirNum{1} Adjusting the surrogate’s backpropagation pathway~\cite{wu2020skip,guo2020backpropagating,wang2023rethinking,zhang2024improving} can produce more effective gradients and consistently boost cross-model transferability.

\CirNum{2} Enhancing surrogate training and representation~\cite{qin2023training,yang2023boosting,chen2025enhancing} improves generalization and reduces overfitting to single architecture.

\CirNum{3} Manipulating token or patch interactions and attention or MLP gradients~\cite{wei2022towards,zhou2022improving,zhang2023transferable,ren2025fpr} effectively increases gradient diversity, which is beneficial for ViTs.

\CirNum{4} Fair and reliable evaluation requires testing across a diverse set of CNNs, ViTs, and representative defenses.

\end{tcolorbox}


\subsection{Ensemble-based Attack}

\Review{zzl,wxs}{Liu~\etal~\cite{liu2016delving} first proposed Ensemble-based attack, which averages the predicted probability of models. Dong~\etal~\cite{dong2018boosting} further proposed another two  strategies for the ensemble-based adversarial attack, namely fusing the logits and loss values of models, respectively. Recently, researchers pay more attention to ensemble-based attack to boost the transferability and decrease the training cost for multiple models.
}


\Review{zzl,fz,wxs}{
\textbf{Ghost Networks}~\cite{li2020learning} simultaneously attack numerous ghost models generated by densely applying dropout and perturbing skip connections in the surrogate model.
}

\Review{wxs}{\textbf{Stochastic Variance Reduced Ensemble (SVRE)}~\cite{xiong2022stochastic} exploits the gradient variance among various models to tune the average gradient on these models, which can stabilize the gradient direction and enhance the transferability.}


\Review{wxs}{\textbf{Large geometric vicinity (LGV)}~\cite{gubri2022lgv} additionally trains the model with high learning rate to produce several models and attacks the models sequentially in a random order to boost the existing attack's transferability.}


\begin{table*}[tb]
    \centering
    \caption{Attack success rates of various ensemble attacks where the adversarial examples are generated on RN-50, VGG-16, MN-v2 and Inc-v3, simultaneously. We have highlighted the approaches that achieve the top-1 attack performance in \colorbox{gray!30}{gray}.}
    \label{tab:untargeted:ensemble-attack}
    \vspace{-.5em}
    \begin{tabular}{l|cccccccc|ccccc|c}
        \toprule
        \multicolumn{1}{c}{\multirow{2}{*}{Attacks}} & \multicolumn{8}{c}{Standardly trained models} & \multicolumn{5}{c}{Defenses} & \multicolumn{1}{c}{\multirow{2}{*}{Average}} \\
        \cmidrule(lr){2-9} \cmidrule(lr){10-14}
        & RN-50 & VGG-16 & MN-v2 & Inc-v3 & ViT & PiT & Visformer & Swin & AT & HGD& RS & NRP & DiffPure & \\
        \midrule\midrule
        Base~\cite{liu2016delving} & 99.4 & 100.0 & 99.9 & 99.5 & 32.2 & 52.6 & 67.5 & 65.3 & 43.4 & 63.0 & 31.4 & 93.8 & 20.6 & 66.8\\
        Ghost~\cite{li2020learning} & 99.3 & 77.4 & 70.6 & 51.6 & 18.2 & 29.3 & 41.5 & ~9.9 & 41.6 & 26.3 & 28.6 & 61.4 & 14.5 & 43.9 \\
        SVRE~\cite{xiong2022stochastic} & 98.8 & 100.0 & 100.0 & 99.6 & 31.5 & 49.7 & 68.1 & 67.4 & 43.8 & 58.6 & 31.1 & 95.2 & 17.4 & 66.2\\
        LGV~\cite{gubri2022lgv} & 91.4 & 97.2 & 95.8 & 80.6 & 28.0 & 35.8 & 58.8 & 55.9 & 44.6 & 58.2 & 32.7 & 67.2 & 18.0 & 58.8\\
        \rowcolor{Gray}MBA~\cite{li2023making} & 99.8 & 99.9 & 99.5 & 96.5 & 53.7 & 60.7 & 87.3 & 82.1 & 48.6 & 90.8 & 36.7 & 82.3 & 24.8 & 74.1\\
        AdaEA~\cite{chen2023anadaptive} & 99.2 & 100.0 & 99.9 & 99.3 & 32.1 & 51.4 & 67.6 & 65.1 & 43.5 & 66.3 & 31.3 & 95.5 & 20.7 & 67.1\\
        CWA~\cite{chen2024rethinking} & 87.1 & 99.9 & 100.0 & 100.0 & 23.9 & 35.7 & 49.3 & 49.9 & 43.9 & 42.7 & 30.7 & 96.0 & 16.9 & 59.7\\
        SMER~\cite{tang2024ensemble} & 96.4 & 95.0 & 96.6 & 96.1 & 33.0 & 51.8 & 67.7 & 68.7 & 43.0 & 62.0 & 30.6 & 90.7 & 17.4 & 65.3\\
        \noalign{\vskip 0.1ex}
        \bottomrule
    \end{tabular}
    \vspace{-.5em}
\end{table*}

\Review{zzl,fz,wxs}{
\textbf{More Bayesian Attack (MBA)}~\cite{li2023making} maximizes the average prediction loss over several models sampled from a Bayesian posterior of the surrogate model, and the posterior is obtained via a single run of finetuning, thus reducing the training cost of obtaining multiple models}


\Review{gzj,wxs}{\textbf{AdaEA}~\cite{chen2023anadaptive} adaptively adjusts the weights of each surrogate model and reducing conflicts between surrogate models by reducing the disparity of gradients.}

\Review{gzj,wxs}{\textbf{Common Weakness Attack (CWA)}~\cite{chen2024rethinking} comprises of two sub-methods named Sharpness Aware Minimization (SAM) and Cosine Similarity Encourager (CSE) to optimize the flatness of loss landscape and the closeness between local optima of different models.}

\Review{gzj,wxs}{\textbf{Stochastic Mini-batch black-box attack with Ensemble Reweighing (SMER)}~\cite{tang2024ensemble} adopts stochastic mini-batch perturbation to craft adversaries from each surrogate model, and adopts reinforcement learning to tailor the ensemble weights.}

\textbf{Evaluations.} As shown in Tab.~\ref{tab:untargeted:ensemble-attack}, ensemble-based attacks consistently exhibit superior performance on CNNs, whereas their transferability to transformer-based models shows greater variability. Overall, self-ensemble methods (\eg, Ghost) tend to underperform other ensemble approaches, largely due to limited diversity among ensemble members. MBA achieves the best overall performance by sampling surrogate models from a Bayesian posterior rather than relying on a small set of deterministic networks. Explicitly enlarging or exploiting a wide weight optimum (\eg, LGV) enhances transferability while avoiding the high cost of training many independent models. Adaptive ensemble strategies (\eg, AdaEA, SVRE, CWA) adjust model weights according to gradient statistics, leading to high attack success rates. SMER achieves competitive based on heterogeneous targets, showing that carefully controlled ensemble diversity can further enhance transferability.


\begin{tcolorbox}[
    colframe=black,
    colback=white,
    left=0.1cm,
    right=0.1cm,
    top=0.1cm,
    bottom=0.1cm,
    boxrule=0.8pt,
    enhanced,
    drop fuzzy shadow,
    opacityback=0.95,
]
\textbf{Takeaways.}


\CirNum{1} Efficiently generating diverse surrogate variants from a limited number of base networks provides a practical trade-off between ensemble diversity and training cost.

\CirNum{2} Simple equal-weight averaging is suboptimal. Methods that adaptively adjust weights or explicitly regularize gradients~\cite{xiong2022stochastic,chen2023anadaptive,chen2024rethinking,tang2024ensemble} outperform vanilla ensembles.

\CirNum{3} Jointly shaping the loss landscape and ensemble diversity tends to yield more transferable adversarial examples.
\end{tcolorbox}

\section{Targeted Attacks}
\label{sec:targeted}
Though the above untargeted attack methods can improve targeted transferability, several approaches have been explicitly designed for targeted attacks. Here we summarize and evaluate the existing targeted transfer-based attacks and delineate common insights for each category. For reference, we also adopt several representative untargeted attacks for evaluation.


\subsection{Input Transformation-based Attack}
\Review{lyy,wxs}{
\textbf{Object-Based Diverse Input Method (ODI)}~\cite{byun2022improving} renders the adversarial example pasted onto a randomly sampled 3D object under a random viewpoint and lighting, while incorporating a noisy background for gradient calculation.

}

\Review{lyy,wxs}{
\textbf{Self-University attack (SU)}~\cite{wei2023enhancing} optimizes the adversarial perturbation on the original image and a randomly cropped patch by minimizing the cross-entropy loss with the target label while maximizing the cosine similarity of their features.
}

\Review{zzl,wxs}{\textbf{Input-Diversity-based Adaptive Attack (IDAA)}~\cite{liu2024boosting} employs local mixup over adversarial images, projects perturbations into \texttt{tanh} space, and adaptively adjusts region-wise step sizes using second-order momentum information.}

\Review{zrx,wxs}{\textbf{Everywhere Attack}~\cite{zeng2025everywhere} divides an image into multiple small local regions, which are concatenated together to be fed into a surrogate model to cover all attention regions.}

\begin{table*}[tb]
    \centering
    \caption{Attack success rates of various input transformation-based attacks where the adversarial examples are generated on RN50. We have highlighted the approach that achieves the top-1 attack performance in \colorbox{gray!30}{gray}.}
    \vspace{-.5em}
    \resizebox{\textwidth}{!}{
    \label{tab:targetedcover:input_transformation}
    \begin{tabular}{l|cccccccc|ccccc|c}
        \toprule
        \multicolumn{1}{c}{\multirow{2}{*}{Attacks}} & \multicolumn{8}{c}{Standardly trained models} & \multicolumn{5}{c}{Defenses} & \multicolumn{1}{c}{\multirow{2}{*}{Average}} \\
        \cmidrule(lr){2-9} \cmidrule(lr){10-14}
        & RN-50 & VGG-16 & MN-v2 & Inc-v3 & ViT & PiT & Visformer & Swin & AT & HGD& RS & NRP & DiffPure \\
        \midrule\midrule
        ODI~\cite{byun2022improving} &~~98.8 &21.3 &23.9 &28.9 &~~5.4 &~16.7 &~40.2 &~16.9 & ~0.0 & 31.6 & ~0.1 & ~0.0 & ~0.3 &21.9\\
        SU~\cite{wei2023enhancing} & ~100.0 &~6.2 &~7.0 &~2.8 &~~0.9 &~~2.7 &~13.6 &~~4.3 & ~0.0 &~3.4 & ~0.0 & ~0.0 & ~0.0 & 10.8\\
        IDAA~\cite{liu2024boosting} &~~34.3&19.4&18.8 &12.6&~~2.1&~~4.5 &~14.0 &~~7.8 & ~0.1 &~0.0 & ~0.1 & ~0.0 & ~0.0 & ~8.7\\
        Everywhere~\cite{zeng2025everywhere} &~~99.8 & 64.2 & 72.7 & 44.5 & 23.0 &~46.0 &~72.6 &~54.7 & ~0.2 & 56.3 & ~0.0 & ~2.1 & ~0.2 & 41.3\\
        \noalign{\vskip 0.1ex}\cdashline{1-15}\noalign{\vskip 0.1ex}
        STM~\cite{ge2023improving} &~~99.9 & 14.9 & 16.4 & 19.6 &~~3.4 &~~8.7 &~28.2 &~14.5 & ~0.0 & 22.8 & ~0.0 & ~0.0 & ~0.3 & 17.6\\
        DeCoWA~\cite{lin2024boosting} & ~100.0 & 20.7 & 21.7 & 15.3 &~~7.2&~26.7 &~58.3 &~31.5 & ~0.1 & 27.5 & ~0.0 & ~0.1 & ~0.1 & 23.8\\
        \rowcolor{Gray}L2T~\cite{zhu2024learning} & ~100.0 & 80.5 & 88.6 & 79.4 & ~41.5 &~75.3 &~93.6 & ~84.3 & ~0.2 & 86.8 & ~0.0 & ~2.7 & ~2.9 & 56.6\\
        BSR~\cite{wang2024boosting} & ~100.0 & 27.2 & 25.9 & 15.1 &~~4.4 &~31.1 &~55.8 &~29.2 & ~0.0 & 17.7 & ~0.0 & ~0.1 &  ~0.1 & 23.6\\
        OPS~\cite{guo2025OPS} &~~84.4 & 19.7 & 26.0 & 27.3 & 13.0 &~20.8 &~34.3 &~24.8 & 45.2 & 70.0 & 36.3 & 48.8 & 39.1 & 37.7\\
        \bottomrule
    \end{tabular}
    }
    \vspace{-.5em}
\end{table*}

\textbf{Evaluations.} As shown in Tab.~\ref{tab:targetedcover:input_transformation}, targeted input transformation-based attacks generally outperform untargeted attacks, underscoring the importance of task-specific designs for targeted settings. Overall, targeted input-transformation based attacks boost transferability by leveraging rich local semantic information. Although SU attains a 100\% white-box success rate, its poor cross-architecture transfer indicates overfitting to source-specific features. In contrast, Everywhere Attack achieves state-of-the-art performance across nearly all evaluated models and defense mechanisms, thanks to its substantially broader attention coverage. However, all the attacks exhibit limited effectiveness against defense methods, underscoring the inefficiency of existing targeted attacks.

\begin{tcolorbox}[
    colframe=black,
    colback=white,
    left=0.1cm,
    right=0.1cm,
    top=0.1cm,
    bottom=0.1cm,
    boxrule=0.8pt,
    enhanced,
    drop fuzzy shadow,
    opacityback=0.95,
]
\textbf{Takeaways.}

\CirNum{1} Enhancing the self-universality of perturbations through local patch transformations (\eg, random cropping and resizing~\cite{wei2023enhancing}, mixing up~\cite{liu2024boosting}) yields perturbations with superior adversarial transferability.

\CirNum{2} Transforming images to cover more attention regions yields more transferable targeted adversarial examples. 

\CirNum{3} Incorporating feature-level alignment into the loss function, such as maximizing cosine similarity~\cite{wei2023enhancing}, serves as an auxiliary optimization strategy.

\end{tcolorbox}

\subsection{Advanced Objective Function}



\Review{lyy,wxs}{
\textbf{Poincar\'e Distance with Triplet Loss (Po+Trip)}~\cite{li2020towards} 
employs an objective that minimizes the Poincaré distance between the accumulated logits of multiple models and the target label, augmented with a triplet loss on angular distances among the logits, target label, and ground-truth label.
}



\Review{zzl, wxs}{
\textbf{Logit Attack}~\cite{zhao2021success} directly maximizes the logit output of target class using a large number of iterations (\ie, 300).
}

\Review{lyy,wxs}{\textbf{Logit Margin Attack}~\cite{weng2023logit} proposes three strategies: a temperature-based approach that adopts a scaling coefficient in the cross-entropy loss, a margin-based approach that rescales logits using the gap between the top two logits, and an angle-based approach that minimizes the cosine similarity between the final-layer weights and feature representations.
}

\Review{lyy,wxs}{\textbf{Clean Feature Mixup (CFM)}~\cite{byun2023introducing} stochastically mixes the feature maps of adversarial examples with arbitrary clean feature maps of benign images after specific convolutional layers and all fully connected layers.} 


\Review{wxs}{\textbf{Feature space FineTuning (FFT)}~\cite{zeng2024enhancing} finetunes adversarial examples in the feature space by enhancing features aligned with the target class while suppressing features of the original class within a middle layer of the surrogate model.}

\begin{table*}[tb]
    \centering
    
    \caption{Attack success rates of various advanced objective functions where the adversarial examples are generated on RN50. We have highlighted the approaches that achieve the top-1 attack performance in \colorbox{gray!30}{gray}.}
    \label{tab:targeted:objective-function}
    \vspace{-.5em}
    \resizebox{\textwidth}{!}{
    \begin{tabular}{l|cccccccc|ccccc|c}
        \toprule
        \multicolumn{1}{c}{\multirow{2}{*}{Attacks}} & \multicolumn{8}{c}{Standardly trained models} & \multicolumn{5}{c}{Defenses} & \multicolumn{1}{c}{\multirow{2}{*}{Average}} \\
        \cmidrule(lr){2-9} \cmidrule(lr){10-14}
        & RN-50 & VGG-16 & MN-v2 & Inc-v3 & ViT & PiT & Visformer & Swin & AT & HGD& RS & NRP & DiffPure\\
        \midrule\midrule
        Po+Trip~\cite{li2020towards} & ~100.0 & ~~5.0 & ~~7.4 & ~~7.7 & ~~0.8 & ~~5.2 & ~13.9 & ~~5.0 & 0.1 & ~~5.9 & 0.0 & 0.0 & 0.0 & ~11.6\\
        Logit~\cite{zhao2021success} & ~~57.2 & ~22.8 & ~16.7 & ~19.4 & ~~2.4 & ~~5.5 & ~24.7 & ~10.5 & 0.1 & ~32.6 & 0.0 & 0.0 & 0.1 & ~14.8\\
        Logit Margin~\cite{weng2023logit}& ~~56.5 & ~21.9 & ~17.1 & ~20.9 & ~~1.6 & ~~6.0 & ~24.3 & ~~8.8 & 0.1 & ~30.8 & 0.0 & 0.0 & 0.0 & ~14.5\\
    \rowcolor{Gray}CFM~\cite{byun2023introducing}& ~~99.1 & ~54.2 & ~62.9 & ~46.9 & ~21.6 &~38.1 & ~63.7 & ~46.0 & 0.2 & ~54.2 & 0.0& 0.5 & 0.8 &~37.7\\
    FFT~\cite{zeng2024enhancing} & ~~97.3 & ~11.5 & ~14.3 & ~~6.3 & ~~1.9 & ~~3.8 & ~19.1 & ~~7.4 & 0.1 & ~11.1 & 0.0 & 0.0 & 0.0 & ~13.3\\
        \noalign{\vskip 0.1ex}\cdashline{1-15}\noalign{\vskip 0.1ex}
        ILPD~\cite{li2023improving} &~~14.2 &~~0.5 &~~0.7 &~~0.8 &~~0.0 &~~1.0 & ~~1.6 & ~~1.1 & 0.1 & ~~0.7 & 0.0 & 0.2 & 0.2 & ~~1.6\\
        BFA~\cite{wang2024improving} & ~~24.8 &~~4.2 &~~3.8 &~~3.5 &~~1.0 &~~3.2 & ~~5.8 &~~4.6 & 0.0 & ~~3.9 & 0.0 & 0.5 & 0.0 & ~~4.3 \\
        \bottomrule
        
    \end{tabular}
    }
    \vspace{-.5em}
\end{table*}

\begin{table*}[tb]
    \centering
    
    \caption{Attack success rates of various generation-based attacks. We have highlighted the approaches that achieve the top-1 attack performance in \colorbox{gray!30}{gray}.}
    \label{tab: target generation-based-attack}
    \vspace{-.5em}
    \resizebox{\textwidth}{!}{
    \begin{tabular}{l|cccccccc|ccccc|c}
        \toprule
        \multicolumn{1}{c}{\multirow{2}{*}{Attacks}} & \multicolumn{8}{c}{Standardly trained models} & \multicolumn{5}{c}{Defenses} & \multicolumn{1}{c}{\multirow{2}{*}{Average}} \\
        \cmidrule(lr){2-9} \cmidrule(lr){10-14}
        & RN-50 & VGG-16 & MN-v2 & Inc-v3 & ViT & PiT & Visformer & Swin & AT & HGD& RS & NRP & DiffPure\\
        \midrule\midrule
        TTP~\cite{naseer2021generating} & 71.1 & 71.8 & 52.4 & 32.8 & 8.0 & 12.6 & 47.8 & 30.5 & 0.2 & 57.8 & 0.3 & 1.8 & 1.5 &29.9 \\
        \rowcolor{Gray}M3D~\cite{zhao2023minimizing} & 88.5 & 89.2 & 79.8 & 59.0 & 32.5 & 35.8 & 76.4 & 60.5 & 0.2 & 80.2 & 0.4 & 3.5 & 1.7 & 46.8\\
        AIM~\cite{li2025aim} & 69.4 & 51.6 & 43.3 & 15.7 & 12.2 & 12.8 & 40.4 & 24.6 & 0.1 & 30.3 & 0.2 & 0.4 & 0.7& 23.2\\
        \noalign{\vskip 0.1ex}
        \bottomrule
    \end{tabular}
    }
    \vspace{-.5em}
\end{table*}
\textbf{Evaluations.} 
As shown in Tab.~\ref{tab:targeted:objective-function}, specially designed methods consistently outperform their untargeted counterparts. Among these methods, CFM clearly emerges as the most effective approach, indicating that stochastically mixing adversarial features with clean feature maps provides a substantially stronger inductive bias for steering perturbations toward the target manifold than conventional optimization-based strategies. Although FFT and Po+Trip improve upon standard logit-based objectives by operating in feature space or incorporating more sophisticated metric constraints, their transferability remains limited in the absence of the regularization induced by feature mixing. 

\begin{tcolorbox}[
    colframe=black,
    colback=white,
    left=0.1cm,
    right=0.1cm,
    top=0.1cm,
    bottom=0.1cm,
    boxrule=0.8pt,
    enhanced,
    drop fuzzy shadow,
    opacityback=0.95,
]
\textbf{Takeaways.}

\CirNum{1} Mixing adversarial features with benign features drawn from other images introduces a strong form of regularization, effectively mitigating overfitting to the surrogate model’s feature space and promoting better generalization across architectures.

\CirNum{2} There is a clear hierarchy in targeted transferability: manipulating intermediate representations~\cite{byun2023introducing,zeng2024enhancing,li2020towards} consistently outperforms manipulating the final output logits~\cite{zhao2021success,weng2023logit}. Intermediate features contain more shared semantic information across different architectures.


\end{tcolorbox}

\subsection{Generation-based Attacks}

\Review{yzy,wxs}{\textbf{Transferable Targeted Perturbations (TTP)}~\cite{naseer2021generating} builds upon the CDA framework~\cite{naseer2019cross} by additionally perturbing augmented images during training. It minimizes the discrepancy between the victim model’s outputs for adversarially perturbed benign or augmented inputs and the target image, measured using KL divergence and neighborhood similarity.
}


\Review{yzy,wxs}{\textbf{Minimizing Maximum Model Discrepancy (M3D)} \cite{zhao2023minimizing} theoretically shows that transferability depends on both the empirical attack error on surrogate models and the maximum discrepancy among them. Accordingly, M3D trains a generator with two discriminators to jointly minimize classification loss and inter-discriminator discrepancy.

}


\Review{zrx,wxs}{
\textbf{Semantic Injection Module (SIM)}~\cite{li2025aim} utilizes the semantics contained in an additional
image associated with the target concept (label), which is injected into the generator.}


\textbf{Evaluations.} As shown in ~\cref{tab: target generation-based-attack}, generation-based attacks exhibit substantial variation in transferability across different methods. M3D achieves the highest overall average success rate, suggesting that reducing discrepancies among surrogate models is an effective strategy to improve transferability in black-box settings. In contrast, TTP attains higher success rates on target models that share similar architectures with the surrogate models, which supports the effectiveness of exploiting neighborhood similarity to constrain the target distribution.

\begin{tcolorbox}[
    colframe=black,
    colback=white,
    left=0.1cm,
    right=0.1cm,
    top=0.1cm,
    bottom=0.1cm,
    boxrule=0.8pt,
    enhanced,
    drop fuzzy shadow,
    opacityback=0.95,
]
\textbf{Takeaways.}

\CirNum{1} Incorporating semantic information of the target class into adversarial example generators~\cite{li2025aim} helps establish a stable, target-oriented constraint in the adversarial space.

\CirNum{2}
Relying solely on label-level or feature-level signals is insufficient to ensure targeted transferability. Thus, jointly exploiting multi-level information has become a prevalent paradigm in generative targeted adversarial attacks.

\CirNum{3}
Targeted transferability benefits from reducing generator overfitting to a single discriminator through dual discriminators in M3D~\cite{zhao2023minimizing}. 

\CirNum{4}
Data augmentation can enhance the robustness~\cite{naseer2021generating} of adversarial examples during generator training.

\end{tcolorbox}

\subsection{Ensemble-based Attack}

\Review{fz,wxs}{\textbf{Sharpness-Aware Self-Distillation with Weight Scaling (SASD-WS)}~\cite{wu2024improving} employs sharpness-aware minimization and distillation to endow the source model with a flatter loss landscape, and uses a weight-scaled model to approximate an ensemble of pruned models.}

\Review{zrx,wxs}{\textbf{Robust Feature Coverage Attack (RFCoA)}~\cite{wang2025breaking} optimizes the robust adversarial features over multiple classifiers, which are fused with those of clean samples and decoded back into the image space.}


\textbf{Evaluations.} As shown in Tab.~\ref{tab:targeted:ensemble-attack}, SASD-WS achieves the highest attack success rate, indicating that strengthening surrogate models and aggregating information from multiple surrogates can substantially enhance transferability. In contrast, RFCoA exhibits notable performance variability across different target models, which may be attributed to its primary design focus on physical-world attacks, where robustness to real-world perturbations is prioritized over transferability across digital architectures.



\begin{tcolorbox}[
    colframe=black,
    colback=white,
    left=0.1cm,
    right=0.1cm,
    top=0.1cm,
    bottom=0.1cm,
    boxrule=0.8pt,
    enhanced,
    drop fuzzy shadow,
    opacityback=0.95,
]
\textbf{Takeaways.}

\CirNum{1} Fusing information of several surrogate models effectively boosts transferability~\cite{wu2024improving,wang2025breaking}.

\CirNum{2} Fine-tuning surrogate models further improves the effectiveness of ensemble-based attacks by strengthening the quality and alignment of the transferred adversarial signals.


\end{tcolorbox}



\begin{table*}[tb]

    \centering

    \caption{Attack success rates of various ensemble-based attacks. We have highlighted the approaches that achieve the top-1 attack performance in \colorbox{gray!30}{gray}.}

    \vspace{-.5em}

    \resizebox{\textwidth}{!}{

    \label{tab:targeted:ensemble-attack}

    \begin{tabular}{l|cccccccc|ccccc|c}

        \toprule


        \multicolumn{1}{c}{\multirow{2}{*}{Attacks}} & \multicolumn{8}{c}{Standardly trained models} & \multicolumn{5}{c}{Defenses} & \multicolumn{1}{c}{\multirow{2}{*}{Average}} \\
        \cmidrule(lr){2-9} \cmidrule(lr){10-14}

        & RN-50 & VGG-16 & MN-v2 & Inc-v3 & ViT & PiT & Visformer & Swin & AT & HGD & RS & NRP & DiffPure \\

        \midrule\midrule

        \rowcolor{Gray}SASD-WS~\cite{wu2024improving} & ~90.5 & 82.2 & 78.9 & 66.8 & 18.3 & ~27.2 &~64.5 &~41.8 & 0.2 & ~76.3 & 0.0 & ~~0.9 & 0.7 & ~42.2 \\

        RFCoA~\cite{wang2025breaking} & ~99.5 & 98.4 &~5.9 & 98.8 &~0.7 & ~~3.3 &~10.6 & ~~5.1 & 0.0 & ~~0.0 & 0.0 & ~~0.1 & 0.0 & ~24.8 \\

        \noalign{\vskip 0.1ex}\cdashline{1-15}\noalign{\vskip 0.1ex}

        Base~\cite{liu2016delving} &  ~99.7 & 40.4 & 99.7 & 98.9 &~0.0 &~~0.6 &~~0.9 &~~0.9 & 0.0 & ~~0.1 & 0.0 & ~~0.0 & 0.0 & ~26.2\\

        SVRE~\cite{xiong2022stochastic} &~~9.2 & 50.2 & 20.6 &~9.0 &~0.6 &~~2.1 &~~7.7 &~~5.3 & 0.0 & ~~1.5 & 0.0 & ~18.1 & 0.0 & ~~9.6\\

        MBA~\cite{li2023making} &~97.1 & 85.3 & 84.6 & 39.4 & 14.8 & ~27.1 &~68.5 & ~52.3 & 0.3 & ~56.8 & 0.0 & ~~0.1 & 0.2 & ~40.5\\

        AdaEA~\cite{chen2023anadaptive} & ~19.9 &~0.5 &~5.1 & 78.2 &~0.0 & ~~0.1 &~~0.0 &~~0.0 & 0.0 & ~~0.1 & 0.0 & ~~0.0 & 0.0 & ~~8.0\\

        \bottomrule

    \end{tabular}

    }

    \vspace{-.5em}

\end{table*}

\section{Further discussion}
\label{sec:dis}
During the review and evaluations, we find several critical issues that may not receive adequate consideration:
\begin{itemize}
    \item \textbf{Comparison}. The effectiveness of adversarial attacks can vary substantially across categories. As these methods are not mutually exclusive, direct cross-category comparisons may be misleading. We recommend focusing on intra-category evaluations and examining the interoperability of attacks across categories.
    
    \item \textbf{Model's weights}. Recent updates to the \texttt{torchvision} ResNet family incorporate stronger data augmentation and exhibit improved robustness. To ensure reliable evaluation, we recommend using the updated pretrained weights (\ie, setting \texttt{weights="DEFAULT"}).
    
    \item \textbf{Saving adversarial examples}. Several studies evaluate attacks using floating-point adversarial examples. In practice, however, transfer-based attacks are often applied via saved images, which must satisfy pixel-value constraints. We therefore recommend storing adversarial examples in standard image formats (\eg, PNG or JPG) prior to assessing transferability.

    \item \textbf{Computational cost.} Many recent attacks improve transferability by evaluating gradients over multiple augmented or replicated samples, achieving strong performance at the expense of increased computational cost. We recommend that future work explicitly report and analyze the computational overhead of attack methods.

    \item \textbf{Challenges and opportunities}. It is important and challenging to focus on adversarial transferability against stronger defenses, extend targeted attack settings, and address more complex and realistic scenarios.
\end{itemize}



\section{Transfer-based Attack beyond Image Classification}
\label{sec:beyond}

Beyond image classification, numerous studies have explored enhancing transferability across diverse domains and tasks, such as face recognition and object detection in Computer Vision (CV), text classification and text generation in Natural Language Processing (NLP), and Visual Question Answering (VQA) in Multi-modal learning. Fig.~\ref{fig:cross-model-transfer-attack} illustrates the categorization of existing transfer-based adversarial attacks beyond image classification and the corresponding methods.

\subsection{Transfer-based Attacks for Other tasks}

Given the inherent differences among  Computer Vision (CV), Natural Language Processing (NLP), and Multi-Modal Tasks, the strategies used to enhance adversarial transferability diverge significantly from those employed in standard image classification settings. Here we detail the specific challenges and methodologies within these three major categories.


\subsubsection{Computer Vision}

Image classification is the foundational CV task for exploring transferability. Given the complex applications of image processing, understanding the mechanism of attacks on other tasks is necessary to complete the transferability map for the image modality. 
a) For \textbf{face recognition}, early research established that physical realizability and robustness are core challenges~\cite{sharif2016accessorize}. To enhance transferability, subsequent works focus on incorporating diverse variants and comprehensive characteristics during data processing, including makeup variants~\cite{sun2024makeupattack}, face recovery~\cite{zhou2024improving}, multi-task constraints~\cite{li2023sibling}, and local perturbation~\cite{zhou2024rethinking, katzav2025adversarialeak}. Liu~\etal~\cite{liu2024adv} validated the effectiveness of applying perturbations on the identity-insensitive region and manipulating adversarial images in latent spaces to systematically enhance transferability.   
b) For \textbf{object detection}, transferability is closely tied to object boundary information at both global and local scales~\cite{xie2017adversarial, huang2023t, bao2024glow}. Recent work often improves transferability by manipulating salient and contextual features across viewpoints~\cite{ding2024transferable}, leveraging global layout awareness to diversify attack strategies~\cite{bao2024glow}, and revealing that fusion-based 3D detectors are constrained by their weakest constituent model~\cite{cheng2023fusion}. 
c) For \textbf{segmentation}, attacks focus primarily on disrupting pixel-level dependencies and feature consistency. DAG~\cite{xie2017adversarial} initially established the segmentation attacks by targeting region proposal networks. Subsequent work has explored manipulating specific feature interactions~\cite{wei2019transferable, xu2023backdoor, arnab2018robustness}.
d) For \textbf{image retrieval}, attacks typically involve the construction of specific retrieval features, \eg, hash-based features~\cite{gao2023backdoor, ding2023vith} and semantic-based features~\cite{li2023semantic}. The diverse investigations on various image processing features expand the understanding of the adversarial transferability.

\subsubsection{Natural Language Processing}
Early textual attacks primarily targeted text classification to induce misleading results. With the emergence of Large Language Models (LLMs), text generation tasks (\eg, question answering, translation, and summarization) have garnered increased attention. 
a) For \textbf{text classification}, early heuristic attacks~\cite{jin2020bert,ren2019generating} utilize word importance ranking (saliency) to greedily substitute key tokens with synonyms. While effective in white-box settings, these greedy approaches often fall into local optima, limiting their transferability to other architectures. To generate more robust transferable examples, Alzantot~\etal~\cite{alzantot2018generating} employed \textbf{genetic algorithms} to iteratively evolve adversarial populations, while Zang~\etal~\cite{zang2020word} utilized a PSO-based strategy to treat word substitution as a combinatorial optimization problem. To ensure that manipulated tokens remain semantically coherent and effective across different model encoders, several works~\cite{li2020bert,garg2020bae} leverage masked language models to generate context-dependent substitutes. 
b) For \textbf{text generation}, the focus of adversarial attacks shifts from classification errors to text generation control, specifically \textit{jailbreaking}. Similar to transfer-based attacks in classification, jailbreaking aims to circumvent alignment and generate harmful content, often leveraging optimization techniques~\cite{zou2023universal, lapid2024open} and adversarial prompt learning~\cite{mehrotra2023tree,liuautodan,chao2025jailbreaking,jin2024guard,zeng2024johnny} that can generalize across models. The transferability of these attacks depends on the commonality between model architectures and the similarity in the word embedding space. Liao~\etal~\cite{liao2024amplegcg} proposed using the samples generated by GCG~\cite{zou2023universal} to train models for universal and transferable suffixes. Diverse prompt transformations and substitutions~\cite{jiang2024artprompt, ghanim2024jailbreaking, li2024drattack, lin2024towards} target model vulnerabilities determined by the coverage and alignment of training data.

\definecolor{coral}{rgb}{1.0, 0.5, 0.31}
\definecolor{darkpink}{rgb}{0.91, 0.33, 0.5}
\definecolor{celadon}{rgb}{0.67, 0.88, 0.69}
\definecolor{tuftsblue}{rgb}{0.28, 0.57, 0.81}

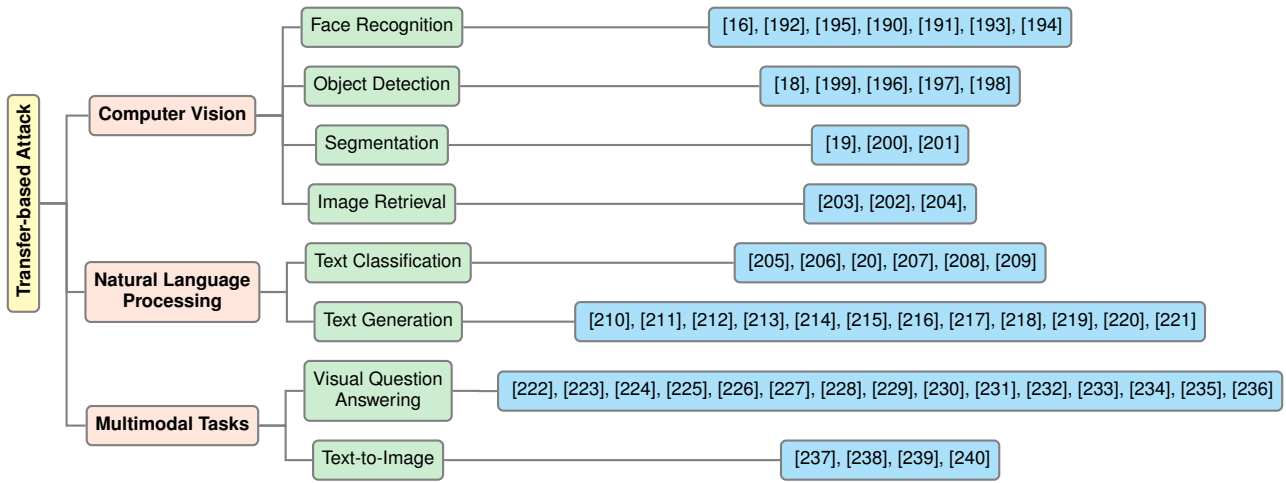
\begin{figure*}
    \centering
    \begin{forest}
        for tree={
            draw={
                gray, 
                thick, 
                font=\sffamily\scriptsize, 
                rounded corners=2, 
                fill=., 
            },
            where level=0{fill=yellow!30}{}, 
            where level=1{fill=coral!20}{}, 
            where level=2{fill=celadon!60}{}, 
            where level=3{fill=cyan!30}{}, 
            l sep+=5pt,
            grow'=east,
            edge={gray, thick},
            parent anchor=east,
            child anchor=west,
            if n children=0{tier=last}{},
            edge path={
                \noexpand\path [draw, \forestoption{edge}] (!u.parent anchor) -- +(10pt,0) |- (.child anchor)\forestoption{edge label};
            },
            if={isodd(n_children())}{
                for children={
                    if={equal(n,(n_children("!u")+1)/2)}{calign with current}{}
                }
            }{},
            align=center
        }
        [{\rotatebox{90}{\textbf{Transfer-based Attack}}}
            [{\begin{tabular}{@{}c@{}}\textbf{Computer Vision}\end{tabular}}
                [{\begin{tabular}{@{}c@{}}Face Recognition\end{tabular}}
                    [{\begin{tabular}{@{}c@{}}\cite{sharif2016accessorize}, \cite{ li2023sibling}, \cite{ liu2024adv}, \cite{sun2024makeupattack}, \cite{zhou2024improving}, \cite{zhou2024rethinking}, \cite{katzav2025adversarialeak}
                    \end{tabular}}]
                ]
                [{\begin{tabular}{@{}c@{}}Object Detection\end{tabular}}
                    [{\begin{tabular}{@{}c@{}}\cite{xie2017adversarial}, \cite{ cheng2023fusion}, \cite{huang2023t}, \cite{bao2024glow},\cite{ding2024transferable}
                    \end{tabular}}]
                ]
                [{\begin{tabular}{@{}c@{}}Segmentation\end{tabular}}[{\begin{tabular}{@{}c@{}} \cite{ wei2019transferable},\cite{xu2023backdoor},\cite{arnab2018robustness}
                        \end{tabular}}]
                ]
                [{\begin{tabular}{@{}c@{}}Image Retrieval\end{tabular}}
                    [{\begin{tabular}{@{}c@{}}\cite{ding2023vith}, \cite{gao2023backdoor}, \cite{ li2023semantic},
                        \end{tabular}}]
                ]
            ]
            [{\begin{tabular}{@{}c@{}}\textbf{Natural Language}\\ \textbf{Processing}\end{tabular}}
                [{\begin{tabular}{@{}c@{}}Text Classification\end{tabular}}
                    [{\begin{tabular}{@{}c@{}}\cite{jin2020bert}, \cite{ren2019generating}, \cite{alzantot2018generating}, \cite{zang2020word}, \cite{li2020bert}, \cite{garg2020bae}
                        \end{tabular}}]
                ]
                [{\begin{tabular}{@{}c@{}}Text Generation\end{tabular}}
                    [{\begin{tabular}{@{}c@{}}\cite{zou2023universal}, \cite{ lapid2024open}, \cite{mehrotra2023tree}, \cite{liuautodan},
                    \cite{chao2025jailbreaking}, \cite{jin2024guard}, \cite{zeng2024johnny}, \cite{liao2024amplegcg}, \cite{jiang2024artprompt}, \cite{ghanim2024jailbreaking},
                    \cite{li2024drattack}, \cite{lin2024towards}
                        \end{tabular}}]
                ]                
            ]
            [{\begin{tabular}{@{}c@{}}\textbf{Multimodal Tasks}\end{tabular}}
                [{\begin{tabular}{@{}c@{}}Visual Question \\ Answering\end{tabular}}
                    [{\begin{tabular}{@{}c@{}} \cite{lu2023set},\cite{zheng2024unified}, \cite{zhang2024universal}, \cite{zhang2025adversarial}, \cite{qraitem2024vision}, \cite{zhao2024evaluating}, \cite{ zhang2022towards}, \cite{wang2024break}, \cite{xu2024highly},\cite{yin2024vqattack}, \cite{gu2024agent}, \cite{gao2025boosting}, \cite{lyu2024trojvlm}, \cite{ wang2024transferable}, \cite{ yin2023vlattack}
                        \end{tabular}}]
                ]
                [{\begin{tabular}{@{}c@{}}Text-to-Image\end{tabular}}
                    [{\begin{tabular}{@{}c@{}}\cite{yang2024sneakyprompt}, \cite{ liu2023riatig}, \cite{ tsai2024ring}, \cite{zhuang2023pilot}
                        \end{tabular}}]
                ]
            ]
        ]
    \end{forest}
    \caption{Overview of existing transfer-based adversarial attacks beyond Image Classification. The tasks are categorized into three types: Computer Vision, Natural Language Processing and Multi-Modal Tasks. Subsequently, they are further divided into sub-tasks to depict the nuances of adversarial attacks across various tasks. The arrangement of the referenced papers follows a chronological order, based on their dates of publication. }
    \label{fig:cross-model-transfer-attack}
    \vspace{-.5em}
\end{figure*}

\subsubsection{Multimodal Tasks} 

Adversarial attacks increasingly exploit cross-modal misalignment, yet insufficient modeling of cross-modal interactions often limits transferability. 
a) For \textbf{Visual Question Answering (VQA)}, recent studies enhance transferability by enforcing semantic consistency, including employing set-level alignment-preserving augmentation and cross-modal guidance~\cite{lu2023set}, incorporating momentum mechanisms with data augmentation~\cite{zheng2024unified}, and investigating modality attack priority~\cite{zhang2024universal, zhang2025adversarial, qraitem2024vision, zhao2024evaluating}. Furthermore, some works emphasize modality alignment as a prerequisite for semantic consistency, utilizing pre-training manipulation or encoder-level objectives~\cite{zhang2022towards, wang2024break}, leveraging text-based attacks to guide image perturbations~\cite{xu2024highly}, and enhancing adversarial sample diversity~\cite{yin2024vqattack, gu2024agent, gao2025boosting, lyu2024trojvlm, wang2024transferable, yin2023vlattack}. 
b) For \textbf{text-to-image tasks}, transferability is typically achieved by targeting the shared embedding space or exploiting universal patterns.  Researchers have shown that adversarial prompts optimized on surrogate text encoders or latent spaces can effectively transfer to black-box diffusion models~\cite{yang2024sneakyprompt, zhuang2023pilot, liu2023riatig}. Also, identifying universal adversarial suffixes~\cite{tsai2024ring} has proven effective for bypassing safety filters across diverse generative architectures.

In summary, extensive research has investigated adversarial transferability across various domains and tasks, including face recognition, object detection, segmentation, and image retrieval in CV, text classification and generation in NLP, VQA and text-to-image tasks in multimodal learning. Due to fundamental differences among these tasks, strategies for enhancing transferability vary across domains. Perturbation propagation through key visual features is effective in CV. The approaches in NLP emphasize semantic-preserving transformations and embedding-space manipulations. Multimodal attacks rely critically on cross-modal interactions. While task-specific designs have advanced transferability within individual settings, they often obscure shared underlying principles. To this end, we abstract the task characteristics and adopt a scenario-based framework to systematically analyze transferability across models, datasets, and domains.

\subsection{Cross-Scenarios Analysis}

Current research seeks to enhance transferability by uncovering intrinsic commonalities across different scenarios. Identifying shared characteristics that span model architectures, data distributions, and knowledge representations is key to improving transferability. Accordingly, we present a systematic analysis from three complementary perspectives: cross-model, cross-dataset, and cross-domain transferability.


\subsubsection{Cross-model transferability}
The primary challenge in cross-model transferability stems from discrepancies in decision boundaries and feature manifolds between surrogate and victim models. Enhancing transferability requires identifying model-agnostic vulnerabilities and mitigating overfitting to the surrogate architecture~\cite{wu2024lrs,  zhao2023minimizing}.  
a) In \textbf{CV}, early iterative attacks tend to overfit the surrogate loss landscape. Momentum-based methods~\cite{dong2018boosting} and variance tuning~\cite{wang2021enhancing} mitigate this issue by stabilizing update directions. Since intermediate layers capture more common semantic features, Huang~\etal~\cite{huang2019enhancing} maximize perturbations in these shared feature maps instead of specific logits. It further evolved toward optimizing the surrogate model itself. For instance, AWT~\cite{chen2025enhancing} adaptively fine-tunes surrogate parameters during attack generation to smooth the loss landscape and capture common vulnerability patterns across model families.
b) In \textbf{NLP}, attackers exploit generalized linguistic weaknesses arising from discrete tokenization. Ensemble-based methods distill transferable adversarial patterns into abstract rules (\eg, synonym substitutions or POS constraints~\cite{yuan2021transferability, roth2024token}), enabling attacks to generalize across language models by targeting shared linguistic biases.
c) For \textbf{vision-languade models}, transferability is facilitated by the shared embedding space induced by contrastive learning~\cite{wei2022cross}. VLATTACK~\cite{yin2023vlattack} leverages an iterative cross-search strategy to jointly optimize visual and textual perturbations,  effectively transferring across unseen VLMs by disrupting aligned multimodal representations.

\subsubsection{Cross-dataset transferability} 
Cross-dataset transferability arises when surrogate and victim models are trained on different data distributions, requiring attacks to capture distribution-invariant vulnerabilities. To bridge data gaps, ICE~\cite{qin2025generalized} introduces a meta-learning framework that trains a universal surrogate to learn transferable patterns across datasets, substantially outperforming standard data augmentation. In \textbf{CV}, input transformations are commonly used to simulate diverse views and enforce feature invariance~\cite{xie2019improving, dong2019evading}. In \textbf{NLP}, especially in cross-lingual settings, artifacts introduced by machine translation provide systematic adversarial cues that can be exploited to enable transfer across models trained on different language datasets~\cite{yuan2021transferability,roth2024token}.

\subsubsection{Cross-domain transferability} 
Cross-domain transferability is the most challenging setting, as it involves both task and semantic shifts. Prior work shows that models often over-rely on texture and style cues. StyleAdv~\cite{fu2023styleadv} exploits this bias by perturbing style representations in latent space rather than pixel space. Generative frameworks with relativistic supervision~\cite{naseer2019cross} have successfully synthesized domain-invariant perturbations, validating the existence of a shared adversarial space. To further address realistic black-box constraints, CDTA~\cite{li2023cdta} disrupts intermediate feature representations via contrastive spectral training. When labeled data are unavailable in the target domains, DR-UDA~\cite{wang2020unsupervised} leverages disentangled representations to construct domain-invariant features, thereby substantially improving generalization in face presentation attack detection. Complementary to learning invariance, integrating task-specific structural priors offers another effective pathway. By targeting task-inherent characteristics, attackers can generate perturbations that are robust to domain shifts, such as identity-insensitive regions in face recognition~\cite{li2023sibling} and global layout dependencies in object detection~\cite{bao2024glow}. Extending to LLMs, the mechanism of transferability shifts from perceptual features to cognitive misalignment. Universal adversarial prompts targeting shared reasoning and instruction-following templates have proven capable of bypassing safety constraints across diverse domains, independent of the specific textual distribution~\cite{zou2023universal, chao2025jailbreaking}.

The evolution of adversarial transferability exhibits a unified trend: moving from instance-specific optimization to the systematic exploitation of shared, invariant feature representations and system-level structural vulnerabilities that transcend specific models, datasets, and domains.

\section{Conclusion}
\label{sec:con}
In this work, we have presented a comprehensive survey of transfer-based attacks, organizing over one hundred of attacks into a unified taxonomy comprising gradient-based, input transformation-based, advanced objective function, model-related, ensemble-based, and generation-based attacks. Recognizing the lack of standardized evaluation criteria, we established a rigorous benchmark to assess these approaches under identical experimental conditions, spanning diverse architectures from CNNs to ViTs, and delineated common insights and potential factors for boosting transferability. Beyond image classification, we briefly summarized transferability trends across CV, NLP, and multimodal tasks, revealing a unified shift towards exploiting system-level invariants. We hope the survey and benchmark serve as a solid foundation for rectifying evaluation protocols and inspiring more robust, transferable attack and defenses.

{\scriptsize
\bibliographystyle{IEEEtran}
\bibliography{egbib}

@inproceedings{goodfellow2015explaining,
    author = {Ian J. Goodfellow and Jonathon Shlens and Christian Szegedy},
    booktitle = {{Proceedings of the International Conference on Learning Representations}},
    title = {{Explaining and Harnessing Adversarial Examples}},
    year = {2015}
}

@inproceedings{kurakin2017adversarial,
    author = {Alexey Kurakin and Ian J. Goodfellow and Samy Bengio},
    booktitle = {{Proceedings of the International Conference on Learning Representations (Workshops)}},
    title = {{Adversarial Examples in the Physical World}},
    year = {2017}
}

@inproceedings{dong2018boosting,
    author = {Yinpeng Dong and Fangzhou Liao and Tianyu Pang and Hang Su and Jun Zhu and Xiaolin Hu and Jianguo Li},
    booktitle = {{Proceedings of the IEEE/CVF Conference on Computer Vision and Pattern Recognition}},
    pages = {9185--9193},
    title = {{Boosting Adversarial Attacks With Momentum}},
    year = {2018}
}

@inproceedings{lin2020nesterov,
    author = {Jiadong Lin and Chuanbiao Song and Kun He and Liwei Wang and John E. Hopcroft},
    booktitle = {{Proceedings of the International Conference on Learning Representations}},
    title = {{Nesterov Accelerated Gradient and Scale Invariance for Adversarial Attacks}},
    year = {2020}
}

@inproceedings{gao2020patch,
    author = {Lianli Gao and Qilong Zhang and Jingkuan Song and Xianglong Liu and Heng Tao Shen},
    booktitle = {{Proceedings of the European Conference on Computer Vision}},
    pages = {307--322},
    title = {{Patch-Wise Attack for Fooling Deep Neural Network}},
    year = {2020}
}

@inproceedings{wang2021enhancing,
    author = {Xiaosen Wang and Kun He},
    booktitle = {{Proceedings of the IEEE/CVF Conference on Computer Vision and Pattern Recognition}},
    pages = {1924--1933},
    title = {{Enhancing the Transferability of Adversarial Attacks Through Variance Tuning}},
    year = {2021}
}

@inproceedings{wang2021boosting,
    author = {Xiaosen Wang and Jiadong Lin and Han Hu and Jingdong Wang and Kun He},
    booktitle = {{Proceedings of the British Machine Vision Conference}},
    pages = {272},
    title = {{Boosting Adversarial Transferability through Enhanced Momentum}},
    year = {2021}
}

@inproceedings{zou2022making,
    author = {Junhua Zou and Yexin Duan and Boyu Li and Wu Zhang and Yu Pan and Zhisong Pan},
    booktitle = {{Proceedings of the AAAI Conference on Artificial Intelligence}},
    pages = {3662--3670},
    title = {{Making Adversarial Examples More Transferable and Indistinguishable}},
    year = {2022}
}

@article{gao2021staircase,
    author = {Lianli Gao and Qilong Zhang and Xiaosu Zhu and Jingkuan Song and Heng Tao Shen},
    journal = {{arXiv preprint arXiv:2104.09722}},
    title = {{Staircase Sign Method for Boosting Adversarial Attacks}},
    year = {2021}
}

@article{han2023samplingbased,
    author = {Xu Han and Anmin Liu and Chenxuan Yao and Yanbo Fan and Kun He},
    title = {{Sampling-based Fast Gradient Rescaling Method for Highly Transferable Adversarial Attacks}},
    journal = {{arXiv preprint arXiv:2307.02828}},
    year = {2023}
}

@inproceedings{zhang2022improvingb,
    author = {Ming Zhang and Xiaohui Kuang and Hu Li and Zhendong Wu and Yuanping Nie and Gang Zhao},
    booktitle = {{Proceedings of the International Joint Conference on Artificial Intelligence}},
    pages = {1629--1635},
    title = {{Improving Transferability of Adversarial Examples with Virtual Step and Auxiliary Gradients}},
    year = {2022}
}

@inproceedings{qin2022boosting,
    author = {Zeyu Qin and Yanbo Fan and Yi Liu and Li Shen and Yong Zhang and Jue Wang and Baoyuan Wu},
    booktitle = {{Proceedings of the Advances in Neural Information Processing Systems}},
    title = {{Boosting the Transferability of Adversarial Attacks with Reverse Adversarial Perturbation}},
    year = {2022}
}

@article{wan2023adversarial,
    author = {Chen Wan and Fangjun Huang},
    journal = {{arXiv preprint arXiv:2306.01809}},
    title = {{Adversarial Attack Based on Prediction-Correction}},
    year = {2023}
}

@inproceedings{peng2023boosting,
    author = {Peng, Anjie and Lin, Zhi and Zeng, Hui and Yu, Wenxin and Kang, Xiangui},
    booktitle = {{IEEE International Conference on Acoustics, Speech and Signal Processing (ICASSP)}},
    pages = {1--5},
    title = {{Boosting Transferability of Adversarial Example via an Enhanced Euler’s Method}},
    year = {2023}
}

@inproceedings{zhu2023boosting,
  title={Boosting adversarial transferability via gradient relevance attack},
  author={Zhu, Hegui and Ren, Yuchen and Sui, Xiaoyan and Yang, Lianping and Jiang, Wuming},
  booktitle={Proceedings of the IEEE/CVF international conference on computer vision},
  pages={4741--4750},
  year={2023}
}

@article{wu2023gnp,
    author = {Tao Wu and Tie Luo and Donald C. Wunsch},
    title = {{GNP Attack: Transferable Adversarial Examples via Gradient Norm Penalty}},
    journal = {{Proceedings of the IEEE International Conference on Image Processing }},
    year = {2023}
}

@inproceedings{ma2023transferable,
    title = {Transferable adversarial attack for both vision transformers and convolutional networks via momentum integrated gradients},
    author = {Ma, Wenshuo and Li, Yidong and Jia, Xiaofeng and Xu, Wei},
    booktitle = {Proceedings of the IEEE/CVF International Conference on Computer Vision},
    pages = {4630--4639},
    year = {2023}
}

@article{yang2023improving,
    title = {{Improving the Transferability of Adversarial Examples via Direction Tuning}},
    author = {Yang, Xiangyuan and Lin, Jie and Zhang, Hanlin and Yang, Xinyu and Zhao, Peng},
    journal = {{Information Sciences}},
    volume = {647},
    pages = {119491},
    year = {2023}
}

@inproceedings{ge2023boosting,
     title = {{Boosting Adversarial Transferability by Achieving Flat Local Maxima}},
     author = {Zhijin Ge and Hongying Liu and Xiaosen Wang and Fanhua Shang and Yuanyuan Liu},
     booktitle = {{Proceedings of the Advances in Neural Information Processing Systems}},
     year = {2023}
}

@article{qiu2025mef,
      title={Boosting Adversarial Transferability with Low-Cost Optimization via Maximin Expected Flatness}, 
      author={Chunlin Qiu and Ang Li and Yiheng Duan and Shenyi Zhang and Yuanjie Zhang and Lingchen Zhao and Qian Wang},
      year={2025},
      eprint={2405.16181},
      archivePrefix={arXiv},
      primaryClass={cs.CV},
      url={https://arxiv.org/abs/2405.16181}, 
}

@inproceedings{fang2024strong,
    title = {{Strong Transferable Adversarial Attacks via Ensembled Asymptotically Normal Distribution Learning}},
    author = {Fang, Zhengwei and Wang, Rui and Huang, Tao and Jing, Liping},
    booktitle = {{Proceedings of the IEEE/CVF Conference on Computer Vision and Pattern Recognition}},
    pages = {24841--24850},
    year = {2024}
}

@article{wang2022boosting,
    author = {Jiafeng Wang and Zhaoyu Chen and Kaixun Jiang and Dingkang Yang and Lingyi Hong and Yan Wang and Wenqiang Zhang},
    journal = {{arXiv preprint arXiv:2211.11236}},
    title = {{Boosting the Transferability of Adversarial Attacks with Global Momentum Initialization}},
    year = {2022}
}

@article{wang2024fgsra,
    title = {Improving Adversarial Transferability via Frequency-Guided Sample Relevance Attack},
    author = {Wang, Xinyi and Jin, Zhibo and Zhu, Zhiyu and Zhang, Jiayu and Chen, Huaming},
    booktitle = {Proceedings of the 33rd ACM International Conference on Information and Knowledge Management},
    pages = {2410-2419},
    year = {2024}
}

@article{li2024foolmix,
  title={Foolmix: Strengthen the transferability of adversarial examples by dual-blending and direction update strategy},
  author={Li, Zhankai and Wang, Weiping and Li, Jie and Chen, Kai and Zhang, Shigeng},
  journal={IEEE Transactions on Information Forensics and Security},
  year={2024},
  publisher={IEEE}
}

@article{gan2025boosting,
  title={Boosting the Transferability of Adversarial Examples through Gradient Aggregation},
  author={Gan, Fuquan and Wo, Yan},
  journal={IEEE Transactions on Information Forensics and Security},
  year={2025},
  publisher={IEEE}
}

@inproceedings{xie2019improving,
    author = {Cihang Xie and Zhishuai Zhang and Yuyin Zhou and Song Bai and Jianyu Wang and Zhou Ren and Alan L. Yuille},
    booktitle = {{Proceedings of the IEEE/CVF Conference on Computer Vision and Pattern Recognition}},
    pages = {2730--2739},
    title = {{Improving Transferability of Adversarial Examples With Input Diversity}},
    year = {2019}
}

@inproceedings{dong2019evading,
    author = {Yinpeng Dong and Tianyu Pang and Hang Su and Jun Zhu},
    booktitle = {{Proceedings of the IEEE/CVF Conference on Computer Vision and Pattern Recognition}},
    pages = {4312--4321},
    title = {{Evading Defenses to Transferable Adversarial Examples by Translation-Invariant Attacks}},
    year = {2019}
}

@inproceedings{zou2020improving,
    author = {Junhua Zou and Zhisong Pan and Junyang Qiu and Xin Liu and Ting Rui and Wei Li},
    booktitle = {{Proceedings of the European Conference on Computer Vision}},
    pages = {563--579},
    title = {{Improving the Transferability of Adversarial Examples with Resized-Diverse-Inputs, Diversity-Ensemble and Region Fitting}},
    year = {2020}
}

@inproceedings{wang2021admix,
    author = {Xiaosen Wang and Xuanran He and Jingdong Wang and Kun He},
    booktitle = {{Proceedings of the IEEE/CVF International Conference on Computer Vision}},
    pages = {16138--16147},
    title = {{Admix: Enhancing the Transferability of Adversarial Attacks}},
    year = {2021}
}

@inproceedings{wu2021improving,
    author = {Weibin Wu and Yuxin Su and Michael R. Lyu and Irwin King},
    booktitle = {{Proceedings of the IEEE/CVF Conference on Computer Vision and Pattern Recognition}},
    pages = {9024--9033},
    title = {{Improving the Transferability of Adversarial Samples With Adversarial Transformations}},
    year = {2021}
}

@inproceedings{long2022frequency,
    author = {Yuyang Long and Qilong Zhang and Boheng Zeng and Lianli Gao and Xianglong Liu and Jian Zhang and Jingkuan Song},
    booktitle = {{Proceedings of the European Conference on Computer Vision}},
    pages = {549--566},
    title = {{Frequency Domain Model Augmentation for Adversarial Attack}},
    year = {2022}
}

@inproceedings{yuan2022adaptive,
    title = {Adaptive image transformations for transfer-based adversarial attack},
    author = {Yuan, Zheng and Zhang, Jie and Shan, Shiguang},
    booktitle = {European Conference on Computer Vision},
    pages = {1--17},
    year = {2022},
    organization = {Springer}
}

@inproceedings{zhang2023improving,
    author = {Jianping Zhang and Jen{-}tse Huang and Wenxuan Wang and Yichen Li and Weibin Wu and Xiaosen Wang and Yuxin Su and Michael R. Lyu},
    booktitle = {{Proceedings of the IEEE/CVF Conference on Computer Vision and Pattern Recognition}},
    title = {{Improving the Transferability of Adversarial Samples by Path-Augmented Method}},
    pages = {8173-8182},
    year = {2023}
}

@article{Wei2023BoostingAT,
    author = {Xingxing Wei and Shiji Zhao},
    title = {{Boosting Adversarial Transferability with Learnable Patch-wise Masks}},
    journal = {{IEEE Transactions on Multimedia}},
    pages = {3778-3787},
    volume = {26},
    year = {2024},
    
}

@inproceedings{wang2023structure,
    author = {Xiaosen Wang and Zeliang Zhang and Jianping Zhang},
    booktitle = {{Proceedings of the IEEE/CVF International Conference on Computer Vision}},
    title = {{Structure Invariant Transformation for better Adversarial Transferability}},
    pages={4607-4619},
    year = {2023}
}

@inproceedings{ge2023improving,
     title={{Improving the Transferability of Adversarial Examples with Arbitrary Style Transfer}},
     author={Zhijin Ge and Fanhua Shang and Hongying Liu and Yuanyuan Liu and Liang Wan and Wei Feng and Xiaosen Wang},
     booktitle={{Proceedings of the ACM International Conference on Multimedia}},
     year={2023}
}

@article{wang2023boost,
    title = {{Boost Adversarial Transferability by Uniform Scale and Mix Mask Method}},
    author = {Wang, Tao and Ying, Zijian and Li, Qianmu and Lian, Zhichao},
    journal = {{arXiv preprint arXiv:2311.12051}},
    year = {2023}
}

@inproceedings{lin2024boosting,
    title = {{Boosting Adversarial Transferability across Model Genus by Deformation-Constrained Warping}},
    author = {Lin, Qinliang and Luo, Cheng and Niu, Zenghao and He, Xilin and Xie, Weicheng and Hou, Yuanbo and Shen, Linlin and Song, Siyang},
    booktitle = {{Proceedings of the AAAI Conference on Artificial Intelligence}},
    year = {2024}
}

@inproceedings{zhu2024learning,
    title = {{Learning to Transform Dynamically for Better Adversarial Transferability}},
    author = {Zhu, Rongyi and Zhang, Zeliang and Liang, Susan and Liu, Zhuo and Xu, Chenliang},
    booktitle = {{Proceedings of the IEEE/CVF Conference on Computer Vision and Pattern Recognition}},
    year = {2024}
}

@inproceedings{wang2024boosting,
     title = {{Boosting Adversarial Transferability by Block Shuffle and Rotation}},
     author = {Kunyu Wang and Xuanran He and Wenxuan Wang and Xiaosen Wang},
     booktitle = {{Proceedings of the IEEE/CVF International Conference on Computer Vision}},
     pages = {24336-24346},
     year = {2024}
}

@inproceedings{guo2025OPS,
  title={Boosting Adversarial Transferability through Augmentation in Hypothesis Space},
  author={Guo, Yu and Liu, Weiquan and Xu, Qingshan and Zheng, Shijun and Huang, Shujun and Zang, Yu and Shen, Siqi and Wen, Chenglu and Wang, Cheng},
  booktitle={Proceedings of the Computer Vision and Pattern Recognition Conference},
  pages={19175--19185},
  year={2025}
}

@article{wang2025improving,
  title={Improving the Transferability of Adversarial Examples by Feature Augmentation},
  author={Wang, Donghua and Yao, Wen and Jiang, Tingsong and Zheng, Xiaohu and Wu, Junqi},
  journal={IEEE Transactions on Neural Networks and Learning Systems},
  year={2025},
}

@inproceedings{zhou2018transferable,
    author = {Wen Zhou and Xin Hou and Yongjun Chen and Mengyun Tang and Xiangqi Huang and Xiang Gan and Yong Yang},
    booktitle = {{Proceedings of the European Conference on Computer Vision}},
    pages = {471--486},
    title = {{Transferable Adversarial Perturbations}},
    year = {2018}
}

@inproceedings{huang2019enhancing,
    author = {Qian Huang and Isay Katsman and Zeqi Gu and Horace He and Serge J. Belongie and Ser{-}Nam Lim},
    booktitle = {{Proceedings of the IEEE/CVF International Conference on Computer Vision}},
    pages = {4732--4741},
    title = {{Enhancing Adversarial Example Transferability With an Intermediate Level Attack}},
    year = {2019}
}

@inproceedings{wu2020boosting,
    author = {Weibin Wu and Yuxin Su and Xixian Chen and Shenglin Zhao and Irwin King and Michael R. Lyu and Yu{-}Wing Tai},
    booktitle = {{Proceedings of the IEEE/CVF Conference on Computer Vision and Pattern Recognition}},
    pages = {1158--1167},
    title = {{Boosting the Transferability of Adversarial Samples via Attention}},
    year = {2020}
}

@inproceedings{li2020yet,
    author = {Qizhang Li and Yiwen Guo and Hao Chen},
    booktitle = {{Proceedings of the European Conference on Computer Vision}},
    pages = {241--257},
    title = {{Yet Another Intermediate-Level Attack}},
    year = {2020}
}

@inproceedings{wang2021feature,
    author = {Zhibo Wang and Hengchang Guo and Zhifei Zhang and Wenxin Liu and Zhan Qin and Kui Ren},
    booktitle = {{Proceedings of the IEEE/CVF International Conference on Computer Vision}},
    pages = {7619--7628},
    title = {{Feature Importance-aware Transferable Adversarial Attacks}},
    year = {2021}
}

@inproceedings{wang2021unified,
    author = {Xin Wang and Jie Ren and Shuyun Lin and Xiangming Zhu and Yisen Wang and Quanshi Zhang},
    booktitle = {{Proceedings of the International Conference on Learning Representations}},
    title = {{A Unified Approach to Interpreting and Boosting Adversarial Transferability}},
    year = {2021}
}

@article{wang2021exploring,
    author = {Ruikui Wang and Yuanfang Guo and Ruijie Yang and Yunhong Wang},
    journal = {{arXiv preprint arXiv:2108.07033}},
    title = {{Exploring Transferable and Robust Adversarial Perturbation Generation from the Perspective of Network Hierarchy}},
    year = {2021}
}

@inproceedings{huang2022integtransferable,
    author = {Yi Huang and Adams Wai{-}Kin Kong},
    booktitle = {{Proceedings of the International Conference on Learning Representations}},
    title = {{Transferable Adversarial Attack Based on Integrated Gradients}},
    year = {2022}
}

@article{he2022enhancing,
    author = {Xianglong He and Yuezun Li and Haipeng Qu and Junyu Dong},
    journal = {{arXiv preprint arXiv:2204.10606}},
    title = {{Enhancing the Transferability via Feature-Momentum Adversarial Attack}},
    year = {2022}
}

@inproceedings{zhang2022improvinga,
    author = {Jianping Zhang and Weibin Wu and Jen{-}tse Huang and Yizhan Huang and Wenxuan Wang and Yuxin Su and Michael R. Lyu},
    booktitle = {{Proceedings of the IEEE/CVF Conference on Computer Vision and Pattern Recognition}},
    pages = {14973--14982},
    title = {{Improving Adversarial Transferability via Neuron Attribution-based Attacks}},
    year = {2022}
}

@inproceedings{zhang2022enhancing,
    author = {Yaoyuan Zhang and Yu{-}an Tan and Tian Chen and Xinrui Liu and Quanxin Zhang and Yuanzhang Li},
    booktitle = {{Proceedings of the International Joint Conference on Artificial Intelligence}},
    pages = {1672--1678},
    title = {{Enhancing the Transferability of Adversarial Examples with Random Patch}},
    year = {2022}
}

@article{yang2023fuzziness,
    title = {{Fuzziness-tuned: Improving the Transferability of Adversarial Examples}},
    author = {Yang, Xiangyuan and Lin, Jie and Zhang, Hanlin and Yang, Xinyu and Zhao, Peng},
    journal = {{arXiv preprint arXiv:2303.10078}},
    year = {2023}
}

@inproceedings{jin2023danaa,
    title={{DANAA: Towards transferable attacks with double adversarial neuron attribution}},
    author={Jin, Zhibo and Zhu, Zhiyu and Wang, Xinyi and Zhang, Jiayu and Shen, Jun and Chen, Huaming},
    booktitle={{International Conference on Advanced Data Mining and Applications}},
    pages={456--470},
    year={2023},
    organization={Springer}
}

@inproceedings{li2023improving,
    author = {Qizhang Li and Yiwen Guo and Wangmeng Zuo and Hao Chen},
    title = {{Improving Adversarial Transferability via Intermediate-level Perturbation Decay}},
    booktitle = {{Proceedings of the Advances in Neural Information Processing Systems}},
    year = {2023}
}

@article{wang2024improving,
    author = {Maoyuan Wang and Jinwei Wang and Bin Ma and Xiangyang Luo},
    title = {{Improving the transferability of adversarial examples through black-box feature attacks}},
    journal = {{Neurocomputing}},
    year = {2024}
}

@article{zheng2025enhancing,
author = {Zheng, Desheng and Ke, Wuping and Li, Xiaoyu and Duan, Yaoxin and Yin, Guangqiang and Min, Fan},
title = {{Enhancing the Transferability of Adversarial Attacks via Multi-Feature Attention}},
year = {2025},
journal = {Trans. Info. For. Sec.},
pages = {1462–1474}
}

@inproceedings{naseer2019cross,
    author = {Muzammal Naseer and Salman H. Khan and Muhammad Haris Khan and Fahad Shahbaz Khan and Fatih Porikli},
    booktitle = {{Proceedings of the Advances in Neural Information Processing Systems}},
    pages = {12885--12895},
    title = {{Cross-Domain Transferability of Adversarial Perturbations}},
    year = {2019}
}

@inproceedings{kanth2021learning,
    title={{Learning Transferable Adversarial Perturbations}},
    author={kanth Nakka, Krishna and Salzmann, Mathieu},
    booktitle={{Advances in Neural Information Processing Systems}},
    year={2021}
}

@inproceedings{kim2022diverse,
    author = {Kim, Woo Jae and Hong, Seunghoon and Yoon, Sung-Eui},
    title = {{Diverse Generative Perturbations on Attention Space for Transferable Adversarial Attacks}},
    booktitle = {{Proceedings of the IEEE International Conference on Image Processing}},
    year = {2022}
}

@inproceedings{zhu2024geadvgan,
    author = {Zhiyu Zhu and Huaming Chen and Xinyi Wang and Jiayu Zhang and Zhibo Jin and Kim-Kwang Raymond Choo and Jun Shen and Dong Yuan},
    title = {{GE-AdvGAN: Improving the transferability of adversarial samples by gradient editing-based adversarial generative model}},
    booktitle = {{Proceedings of the SIAM International Conference on Data Mining}},
    year = {2024}
}

@article{chen2024diffattack,
  title={Diffusion models for imperceptible and transferable adversarial attack},
  author={Chen, Jianqi and Chen, Hao and Chen, Keyan and Zhang, Yilan and Zou, Zhengxia and Shi, Zhenwei},
  journal={IEEE Transactions on Pattern Analysis and Machine Intelligence},
  year={2024},
  publisher={IEEE}
}

@article{wang2024boostingb,
  title={Boosting the transferability of adversarial attacks with frequency-aware perturbation},
  author={Wang, Yajie and Wu, Yi and Wu, Shangbo and Liu, Ximeng and Zhou, Wanlei and Zhu, Liehuang and Zhang, Chuan},
  journal={IEEE Transactions on Information Forensics and Security},
  volume={19},
  pages={6293--6304},
  year={2024},
  publisher={IEEE}
}

@article{wu2025dsva,
  title={Boosting Generative Adversarial Transferability with Self-supervised Vision Transformer Features},
  author={Wu, Shangbo and Tan, Yu-an and Ma, Ruinan and Ma, Wencong and Zhu, Dehua and Li, Yuanzhang},
  journal={arXiv preprint arXiv:2506.21046},
  year={2025}
}

@inproceedings{caron2021dino,
  title={Emerging properties in self-supervised vision transformers},
  author={Caron, Mathilde and Touvron, Hugo and Misra, Ishan and J{\'e}gou, Herv{\'e} and Mairal, Julien and Bojanowski, Piotr and Joulin, Armand},
  booktitle={Proceedings of the IEEE/CVF international conference on computer vision},
  pages={9650--9660},
  year={2021}
}

@inproceedings{he2022mae,
  title={Masked autoencoders are scalable vision learners},
  author={He, Kaiming and Chen, Xinlei and Xie, Saining and Li, Yanghao and Doll{\'a}r, Piotr and Girshick, Ross},
  booktitle={Proceedings of the IEEE/CVF conference on computer vision and pattern recognition},
  pages={16000--16009},
  year={2022}
}

@inproceedings{wu2020skip,
    author = {Dongxian Wu and Yisen Wang and Shu{-}Tao Xia and James Bailey and Xingjun Ma},
    booktitle = {{Proceedings of the International Conference on Learning Representations}},
    title = {{Skip Connections Matter: On the Transferability of Adversarial Examples Generated with ResNets}},
    year = {2020}
}

@inproceedings{guo2020backpropagating,
    author = {Yiwen Guo and Qizhang Li and Hao Chen},
    booktitle = {{Proceedings of the Advances in Neural Information Processing Systems}},
    title = {{Backpropagating Linearly Improves Transferability of Adversarial Examples}},
    year = {2020}
}

@inproceedings{wei2022towards,
    author = {Zhipeng Wei and Jingjing Chen and Micah Goldblum and Zuxuan Wu and Tom Goldstein and Yu{-}Gang Jiang},
    booktitle = {{Proceedings of the AAAI Conference on Artificial Intelligence}},
    pages = {2668--2676},
    title = {{Towards Transferable Adversarial Attacks on Vision Transformers}},
    year = {2022}
}

@inproceedings{fang2022learning,
    author = {Shuman Fang and Jie Li and Xianming Lin and Rongrong Ji},
    booktitle = {{Proceedings of the AAAI Conference on Artificial Intelligence}},
    pages = {571--579},
    title = {{Learning to Learn Transferable Attack}},
    year = {2022}
}

@inproceedings{zhu2021rethinking,
    author = {Yao Zhu and Jiacheng Sun and Zhenguo Li},
    booktitle = {{Proceedings of the International Conference on Learning Representations}},
    title = {{Rethinking Adversarial Transferability from a Data Distribution Perspective}},
    year = {2022}
}

@article{zhou2022improving,
    author = {Huipeng Zhou and Yu{-}an Tan and Yajie Wang and Haoran Lyu and Shangbo Wu and Yuanzhang Li},
    journal = {{arXiv preprint arXiv:2204.12680}},
    title = {{Improving the Transferability of Adversarial Examples with Restructure Embedded Patches}},
    year = {2022}
}

@inproceedings{naseer2022on,
    author = {Muzammal Naseer and Kanchana Ranasinghe and Salman Khan and Fahad Shahbaz Khan and Fatih Porikli},
    booktitle = {{Proceedings of the International Conference on Learning Representations}},
    title = {{On Improving Adversarial Transferability of Vision Transformers}},
    year = {2022}
}

@inproceedings{wang2022generating,
    author = {Wang, Yuxuan and Wang, Jiakai and Yin, Zixin and Gong, Ruihao and Wang, Jingyi and Liu, Aishan and Liu, Xianglong},
    title = {{Generating Transferable Adversarial Examples against Vision Transformers}},
    booktitle = {{Proceedings of the ACM International Conference on Multimedia}},
    year = {2022}
}

@article{zhu2022towards,
    author = {Yao Zhu and Yuefeng Chen and Xiaodan Li and Kejiang Chen and Yuan He and Xiang Tian and Bolun Zheng and Yaowu Chen and Qingming Huang},
    journal = {{IEEE Transactions on Image Processing}},
    title = {{Towards Understanding and Boosting Adversarial Transferability from a Distribution Perspective}},
    year = {2022}
}

@inproceedings{qin2023training,
    title={{Training meta-surrogate model for transferable adversarial attack}},
    author={Qin, Yunxiao and Xiong, Yuanhao and Yi, Jinfeng and Hsieh, Cho-Jui},
    booktitle={{Proceedings of the AAAI Conference on Artificial Intelligence}},
    pages={9516--9524},
    year={2023}
}

@inproceedings{yang2023generating,
    author = {Yang, Dingcheng and Yu, Wenjian and Xiao, Zihao and Luo, Jiaqi},
    title = {{Generating Adversarial Examples with Better Transferability via Masking Unimportant Parameters of Surrogate Model}},
    booktitle = {{Proceedings of International Joint Conference on Neural Networks}},
    year = {2023}
}

@inproceedings{zhang2023transferable,
    author = {Jianping Zhang and Yizhan Huang and Weibin Wu and Michael R. Lyu},
    booktitle = {{Proceedings of the IEEE/CVF Conference on Computer Vision and Pattern Recognition}},
    title = {{Transferable Adversarial Attacks on Vision Transformers with Token Gradient Regularization}},
    pages = {16415-16424},
    year = {2023}
}

@inproceedings{yang2023boosting,
    author = {Dingcheng Yang and Zihao Xiao and Wenjian Yu},
    title = {{Boosting the Adversarial Transferability of Surrogate Models with Dark Knowledge}},
    booktitle = {{Proceedings of IEEE International Conference on Tools with Artificial Intelligence}},
    year = {2023}
}

@inproceedings{wang2023diversifying,
     title={{Diversifying the High-level Features for better Adversarial Transferability}},
     author={Zhiyuan Wang and Zeliang Zhang and Siyuan Liang and Xiaosen Wang},
     booktitle={{Proceedings of the British Machine Vision Conference}},
     year={2023},
}

@inproceedings{wang2023rethinking,
    author = {Xiaosen Wang and Kangheng Tong and Kun He},
    title = {{Rethinking the Backward Propagation for Adversarial Transferability}},
    booktitle = {{Proceedings of the Advances in Neural Information Processing Systems}},
    year = {2023}
}

@inproceedings{wang2024ags,
    author = {Wang, Ruikui and Guo, Yuanfang and Wang, Yunhong},
    title = {{AGS: Affordable and Generalizable Substitute Training for Transferable Adversarial Attack}},
    booktitle = {{Proceedings of the AAAI Conference on Artificial Intelligence}},
    year = {2024}
}

@inproceedings{zhang2024improving,
    author = {Zhang, Jianping and Huang, Yizhan and Xu, Zhuoer and Wu, Weibin and Lyu, Michael R.},
    title = {{Improving the Adversarial Transferability of Vision Transformers with Virtual Dense Connection}},
    booktitle = {{Proceedings of the AAAI Conference on Artificial Intelligence}},
    year = {2024}
}

@inproceedings{wu2024lrs,
    author = {Wu, Tao and Luo, Tie and Wunsch II, Donald C.},
    title = {{LRS: Enhancing Adversarial Transferability through Lipschitz Regularized Surrogate}},
    booktitle = {{Proceedings of the AAAI Conference on Artificial Intelligence}},
    year = {2024}
}

@article{weng2024exploring,
    author = {Juanjuan Weng and Zhiming Luo and Shaozi Li},
    title = {{Exploring Frequencies via Feature Mixing and Meta-Learning for Improving Adversarial Transferability}},
    journal = {{arXiv preprint arXiv:2405.03193}},
    year = {2024}
}

@article{ma2024improving,
    author = {Avery Ma and Amir-massoud Farahmand and Yangchen Pan and Philip Torr and Jindong Gu},
    title = {{Improving Adversarial Transferability via Model Alignment}},
    journal = {{arXiv preprint arXiv:2311.18495}},
    year = {2024}
}

@article{ming2024att,
  title={Boosting the transferability of adversarial attack on vision transformer with adaptive token tuning},
  author={Ming, Di and Ren, Peng and Wang, Yunlong and Feng, Xin},
  journal={Advances in Neural Information Processing Systems},
  volume={37},
  pages={20887--20918},
  year={2024}
}

@inproceedings{ren2025fpr,
  title={Improving adversarial transferability on vision transformers via forward propagation refinement},
  author={Ren, Yuchen and Zhao, Zhengyu and Lin, Chenhao and Yang, Bo and Zhou, Lu and Liu, Zhe and Shen, Chao},
  booktitle={Proceedings of the Computer Vision and Pattern Recognition Conference},
  pages={25071--25080},
  year={2025}
}

@article{yang2024quantization,
    author = {Yang, Yulong and Lin, Chenhao and Li, Qian and Zhao, Zhengyu and Fan, Haoran and Zhou, Dawei and Wang, Nannan and Liu, Tongliang and Shen, Chao},
    title = {Quantization Aware Attack: Enhancing Transferable Adversarial Attacks by Model Quantization},
    journal = {Trans. Info. For. Sec.},
    volume = {19},
    pages = {3265–3278},
    year = {2024}
}

@inproceedings{chen2025enhancing,
  title={Enhancing Adversarial Transferability with Adversarial Weight Tuning},
  author={Chen, Jiahao and Feng, Zhou and Zeng, Rui and Pu, Yuwen and Zhou, Chunyi and Jiang, Yi and Gan, Yuyou and Li, Jinbao and Ji, Shouling},
  booktitle={Proceedings of the AAAI Conference on Artificial Intelligence},
  year={2025}
}

@article{liu2025ll2s,
  title={Harnessing the Computation Redundancy in ViTs to Boost Adversarial Transferability},
  author={Liu, Jiani and Wang, Zhiyuan and Zhang, Zeliang and Huang, Chao and Liang, Susan and Tang, Yunlong and Xu, Chenliang},
  journal={arXiv preprint arXiv:2504.10804},
  year={2025}
}

@inproceedings{liu2016delving,
    author = {Yanpei Liu and Xinyun Chen and Chang Liu and Dawn Song},
    booktitle = {{Proceedings of the International Conference on Learning Representations}},
    title = {{Delving into Transferable Adversarial Examples and Black-box Attacks}},
    year = {2017}
}

@inproceedings{li2020learning,
    author = {Yingwei Li and Song Bai and Yuyin Zhou and Cihang Xie and Zhishuai Zhang and Alan L. Yuille},
    booktitle = {{Proceedings of the AAAI Conference on Artificial Intelligence}},
    pages = {11458--11465},
    title = {{Learning Transferable Adversarial Examples via Ghost Networks}},
    year = {2020}
}

@inproceedings{xiong2022stochastic,
    author = {Yifeng Xiong and Jiadong Lin and Min Zhang and John E. Hopcroft and Kun He},
    booktitle = {{Proceedings of the IEEE/CVF Conference on Computer Vision and Pattern Recognition}},
    pages = {14963--14972},
    title = {{Stochastic Variance Reduced Ensemble Adversarial Attack for Boosting the Adversarial Transferability}},
    year = {2022}
}

@inproceedings{gubri2022lgv,
    author = {Martin Gubri and Maxime Cordy and Mike Papadakis and Yves Le Traon and Koushik Sen},
    booktitle = {{Proceedings of the European Conference on Computer Vision}},
    pages = {603--618},
    title = {{{ LGV:} Boosting Adversarial Example Transferability from Large Geometric Vicinity}},
    year = {2022}
}

@inproceedings{li2023making,
    author = {Qizhang Li and Yiwen Guo and Wangmeng Zuo and Hao Chen},
    title = {{Making Substitute Models More Bayesian Can Enhance Transferability of Adversarial Examples}},
    booktitle = {{Proceedings of the International Conference on Learning Representations}},
    year = {2023}
}

@inproceedings{chen2023anadaptive,
    author = {Chen, Bin and Yin, Jiali and Chen, Shukai and Chen, Bohao and Liu, Ximeng},
    title = {{An Adaptive Model Ensemble Adversarial Attack for Boosting Adversarial Transferability}},
    booktitle = {{Proceedings of IEEE/CVF International Conference on Computer Vision}},
    year = {2023}
}

@inproceedings{chen2024rethinking,
    author = {Huanran Chen and Yichi Zhang and Yinpeng Dong and Xiao Yang and Hang Su and Jun Zhu},
    title = {{Rethinking Model Ensemble in Transfer-based Adversarial Attacks}},
    booktitle = {{Proceedings of the International Conference on Learning Representations}},
    year = {2024}
}

@inproceedings{tang2024ensemble,
    title = {{Ensemble Diversity Facilitates Adversarial Transferability}},
    author = {Bowen, Tang and Zheng, Wang and Yi, Bin and Qi, Dou and Yang, Yang and Heng Tao, Shen},
    booktitle = {{Proceedings of the IEEE International Conference on Computer Vision and Pattern Recognition}},
    year = {2024}
}

@inproceedings{byun2022improving,
    author = {Junyoung Byun and Seungju Cho and Myung{-}Joon Kwon and Heeseon Kim and Changick Kim},
    booktitle = {{Proceedings of the IEEE/CVF Conference on Computer Vision and Pattern Recognition}},
    pages = {15223--15232},
    title = {{Improving the Transferability of Targeted Adversarial Examples through Object-Based Diverse Input}},
    year = {2022}
}

@inproceedings{wei2023enhancing,
    author = {Wei, Zhipeng and Chen, Jingjing and Wu, Zuxuan and Jiang, Yu-Gang},
    booktitle = {{Proceedings of the IEEE/CVF Conference on Computer Vision and Pattern Recognition}},
    title = {{Enhancing the Self-Universality for Transferable Targeted Attacks}},
    pages = {12281-12290},
    year = {2023}
}

@article{liu2024boosting,
    author = {Junlin Liu and Xinchen Lyu},
    title = {{Boosting the Transferability of Adversarial Examples via Local Mixup and Adaptive Step Size}},
    journal = {{arXiv preprint arXiv:2401.13205}},
    year = {2024}
}

@inproceedings{inkawhich2019feature,
    author = {Nathan Inkawhich and Wei Wen and Hai (Helen) Li and Yiran Chen},
    booktitle = {{Proceedings of the IEEE/CVF Conference on Computer Vision and Pattern Recognition}},
    pages = {7066--7074},
    title = {{Feature Space Perturbations Yield More Transferable Adversarial Examples}},
    year = {2019}
}

@inproceedings{inkawhich2020transferable,
    author = {Nathan Inkawhich and Kevin J. Liang and Lawrence Carin and Yiran Chen},
    booktitle = {{Proceedings of the International Conference on Learning Representations}},
    title = {{Transferable Perturbations of Deep Feature Distributions}},
    year = {2020}
}

@inproceedings{li2020towards,
    author = {Maosen Li and Cheng Deng and Tengjiao Li and Junchi Yan and Xinbo Gao and Heng Huang},
    booktitle = {{Proceedings of the IEEE/CVF Conference on Computer Vision and Pattern Recognition}},
    pages = {638--646},
    title = {{Towards Transferable Targeted Attack}},
    year = {2020}
}

@inproceedings{inkawhich2020perturbing,
    title = {{Perturbing across the feature hierarchy to improve standard and strict blackbox attack transferability}},
    author = {Inkawhich, Nathan and Liang, Kevin and Wang, Binghui and Inkawhich, Matthew and Carin, Lawrence and Chen, Yiran},
    booktitle = {{Proceedings of the Advances in Neural Information Processing Systems}},
    volume = {33},
    pages = {20791--20801},
    year = {2020}
}

@inproceedings{gao2021feature,
    author = {Lianli Gao and Yaya Cheng and Qilong Zhang and Xing Xu and Jingkuan Song},
    booktitle = {{Proceedings of the International Joint Conference on Artificial Intelligence}},
    pages = {671--677},
    title = {{Feature Space Targeted Attacks by Statistic Alignment}},
    year = {2021}
}

@inproceedings{zhao2021success,
    author = {Zhengyu Zhao and Zhuoran Liu and Martha A. Larson},
    booktitle = {{Proceedings of the Advances in Neural Information Processing Systems}},
    pages = {6115--6128},
    title = {{On Success and Simplicity:{ A} Second Look at Transferable Targeted Attacks}},
    year = {2021}
}

@article{weng2023logit,
    author = {Weng, Juanjuan and Luo, Zhiming and Li, Shaozi and Sebe, Nicu and Zhong, Zhun},
    title = {{Logit Margin Matters: Improving Transferable Targeted Adversarial Attack by Logit Calibration}},
    journal = {{IEEE Transactions on Information Forensics and Security}},
    volume={18},
    pages = {3561-3574},
    year = {2023}
}

@inproceedings{byun2023introducing,
    author = {Byun, Junyoung and Kwon, Myung-Joon and Cho, Seungju and Kim, Yoonji and Kim, Changick},
    booktitle = {{Proceedings of the IEEE/CVF Conference on Computer Vision and Pattern Recognition}},
    pages = {24648--24657},
    title = {{Introducing Competition To Boost the Transferability of Targeted Adversarial Examples Through Clean Feature Mixup}},
    year = {2023}
}

@inproceedings{zeng2024enhancing,
    author = {Zeng, Hui and Chen, Biwei and Peng, Anjie},
    title = {{Enhancing Targeted Transferability VIA Feature Space Fine-Tuning}},
    booktitle = {{Proceedings of IEEE International Conference on Acoustics, Speech and Signal Processing}},
    year = {2024}
}

@inproceedings{naseer2021generating,
    author = {Muzammal Naseer and Salman H. Khan and Munawar Hayat and Fahad Shahbaz Khan and Fatih Porikli},
    booktitle = {{Proceedings of the IEEE/CVF International Conference on Computer Vision}},
    pages = {7688--7697},
    title = {{On Generating Transferable Targeted Perturbations}},
    year = {2021}
}

@inproceedings{zhao2023minimizing,
    title={{Minimizing Maximum Model Discrepancy for Transferable Black-box Targeted Attacks}},
    author={Zhao, Anqi and Chu, Tong and Liu, Yahao and Li, Wen and Li, Jingjing and Duan, Lixin},
    booktitle={{Proceedings of the IEEE/CVF Conference on Computer Vision and Pattern Recognition}},
    pages={8153--8162},
    year={2023}
}

@inproceedings{wang2023towardsb,
    author = {Wang, Zhibo and Yang, Hongshan and Feng, Yunhe and Sun, Peng and Guo, Hengchang and Zhang, Zhifei and Ren, Kui},
    title = {{Towards Transferable Targeted Adversarial Examples}},
    booktitle = {{Proceedings of the IEEE/CVF Conference on Computer Vision and Pattern Recognition}},
    pages = {20534-20543},
    year = {2023}
}

@inproceedings{wang2023lfaa,
    author = {Kunyu Wang and Juluan Shi and Wenxuan Wang},
    title = {{LFAA: Crafting Transferable Targeted Adversarial Examples with Low-Frequency Perturbations}},
    booktitle = {{Proceedings of European Conference on Artificial Intelligence}},
    year = {2023}
}

@inproceedings{wang2024enhancing,
    author = {Hung{-}Jui Wang and Yu{-}Yu Wu and Shang{-}Tse Chen},
    title = {{Enhancing Targeted Attack Transferability via Diversified Weight Pruning}},
    booktitle = {{Proceedings of the IEEE/CVF International Conference on Computer Vision}},
    year = {2024}
}

@inproceedings{wu2024improving,
    title={Improving Transferable Targeted Adversarial Attacks with Model Self-Enhancement},
    author={Wu, Han and Ou, Guanyan and Wu, Weibin and Zheng, Zibin},
    booktitle={Proceedings of the IEEE/CVF Conference on Computer Vision and Pattern Recognition},
    pages={24615--24624},
    year={2024}
}

@inproceedings{sundararajan2017axiomatic,
  title={Axiomatic attribution for deep networks},
  author={Sundararajan, Mukund and Taly, Ankur and Yan, Qiqi},
  booktitle={International conference on machine learning},
  pages={3319--3328},
  year={2017},
  organization={PMLR}
}

@article{xu2023comprehensive,
  title={A comprehensive survey of image augmentation techniques for deep learning},
  author={Xu, Mingle and Yoon, Sook and Fuentes, Alvaro and Park, Dong Sun},
  journal={Pattern Recognition},
  volume={137},
  pages={109347},
  year={2023},
  publisher={Elsevier}
}

@inproceedings{selvaraju2017grad,
	author = {Ramprasaath R. Selvaraju and Michael Cogswell and Abhishek Das and Ramakrishna Vedantam and Devi Parikh and Dhruv Batra},
	booktitle = {{Proceedings of the IEEE/CVF International Conference on Computer Vision}},
	pages = {618--626},
	title = {{Grad-CAM: Visual Explanations from Deep Networks via Gradient-Based Localization}},
	year = {2017}
}

@inproceedings{kingma2015adam,
	author = {Diederik P. Kingma and Jimmy Ba},
	booktitle = {{Proceedings of the International Conference on Learning Representations}},
	title = {{Adam:{ A} Method for Stochastic Optimization}},
	year = {2015}
}

@inproceedings{yun2019cutmix,
	author = {Sangdoo Yun and Dongyoon Han and Sanghyuk Chun and Seong Joon Oh and Youngjoon Yoo and Junsuk Choe},
	booktitle = {{Proceedings of the IEEE/CVF International Conference on Computer Vision}},
	pages = {6022--6031},
	title = {{CutMix: Regularization Strategy to Train Strong Classifiers With Localizable Features}},
	year = {2019}
}

@inproceedings{alex2012ImageNet,
	author = {Alex Krizhevsky and Ilya Sutskever and Geoffrey E. Hinton},
	booktitle = {{Proceedings of the Advances in Neural Information Processing Systems}},
	pages = {1106--1114},
	title = {{ImageNet Classification with Deep Convolutional Neural Networks}},
	year = {2012}
}

@inproceedings{simonyan2015very,
	author = {Karen Simonyan and Andrew Zisserman},
	booktitle = {{Proceedings of the International Conference on Learning Representations}},
	title = {{Very Deep Convolutional Networks for Large-Scale Image Recognition}},
	year = {2015}
}

@inproceedings{he2016deep,
	author = {Kaiming He and Xiangyu Zhang and Shaoqing Ren and Jian Sun},
	booktitle = {{Proceedings of the IEEE/CVF Conference on Computer Vision and Pattern Recognition}},
	pages = {770--778},
	title = {{Deep Residual Learning for Image Recognition}},
	year = {2016}
}

@inproceedings{szegedy2016rethinking,
  title={Rethinking the inception architecture for computer vision},
  author={Szegedy, Christian and Vanhoucke, Vincent and Ioffe, Sergey and Shlens, Jon and Wojna, Zbigniew},
  booktitle={Proceedings of the IEEE conference on computer vision and pattern recognition},
  pages={2818--2826},
  year={2016}
}

@inproceedings{huang2017densely,
	author = {Gao Huang and Zhuang Liu and Laurens van der Maaten and Kilian Q. Weinberger},
	booktitle = {{Proceedings of the IEEE/CVF Conference on Computer Vision and Pattern Recognition}},
	pages = {2261--2269},
	title = {{Densely Connected Convolutional Networks}},
	year = {2017}
}

@inproceedings{sandler2018mobilenetv2,
  title={Mobilenetv2: Inverted residuals and linear bottlenecks},
  author={Sandler, Mark and Howard, Andrew and Zhu, Menglong and Zhmoginov, Andrey and Chen, Liang-Chieh},
  booktitle={Proceedings of the IEEE conference on computer vision and pattern recognition},
  pages={4510--4520},
  year={2018}
}

@inproceedings{dosovitskiy2020image,
	author = {Alexey Dosovitskiy and Lucas Beyer and Alexander Kolesnikov and Dirk Weissenborn and Xiaohua Zhai and Thomas Unterthiner and Mostafa Dehghani and Matthias Minderer and Georg Heigold and Sylvain Gelly and Jakob Uszkoreit and Neil Houlsby},
	booktitle = {{Proceedings of the International Conference on Learning Representations}},
	title = {{An Image is Worth 16x16 Words: Transformers for Image Recognition at Scale}},
	year = {2021}
}

@inproceedings{liu2021swin,
	author = {Ze Liu and Yutong Lin and Yue Cao and Han Hu and Yixuan Wei and Zheng Zhang and Stephen Lin and Baining Guo},
	booktitle = {{Proceedings of the IEEE/CVF International Conference on Computer Vision}},
	pages = {9992--10002},
	title = {{Swin Transformer: Hierarchical Vision Transformer Using Shifted Windows}},
	year = {2021}
}

@inproceedings{szegedy2014intriguing,
	author = {Christian Szegedy and Wojciech Zaremba and Ilya Sutskever and Joan Bruna and Dumitru Erhan and Ian J. Goodfellow and Rob Fergus},
	booktitle = {{Proceedings of the International Conference on Learning Representations}},
	title = {{Intriguing Properties of Neural Networks}},
	year = {2014}
}

@inproceedings{eykholt2018robust,
	author = {Kevin Eykholt and Ivan Evtimov and Earlence Fernandes and Bo Li and Amir Rahmati and Chaowei Xiao and Atul Prakash and Tadayoshi Kohno and Dawn Song},
	booktitle = {{Proceedings of the IEEE/CVF Conference on Computer Vision and Pattern Recognition}},
	pages = {1625--1634},
	title = {{Robust Physical-World Attacks on Deep Learning Visual Classification}},
	year = {2018}
}

@inproceedings{xie2017adversarial,
	author = {Cihang Xie and Jianyu Wang and Zhishuai Zhang and Yuyin Zhou and Lingxi Xie and Alan L. Yuille},
	booktitle = {{Proceedings of the IEEE/CVF International Conference on Computer Vision}},
	pages = {1378--1387},
	title = {{Adversarial Examples for Semantic Segmentation and Object Detection}},
	year = {2017}
}

@inproceedings{alzantot2018generating,
	author = {Moustafa Alzantot and Yash Sharma and Ahmed Elgohary and Bo{-}Jhang Ho and Mani B. Srivastava and Kai{-}Wei Chang},
	booktitle = {{Proceedings of the Conference on Empirical Methods in Natural Language Processing}},
	pages = {2890--2896},
	title = {{Generating Natural Language Adversarial Examples}},
	year = {2018}
}

@inproceedings{wang2021adversarial,
	author = {Xiaosen Wang and Yichen Yang and Yihe Deng and Kun He},
	booktitle = {{Proceedings of the AAAI Conference on Artificial Intelligence}},
	pages = {13997--14005},
	title = {{Adversarial Training with Fast Gradient Projection Method against Synonym Substitution Based Text Attacks}},
	year = {2021}
}

@inproceedings{sharif2016accessorize,
	author = {Mahmood Sharif and Sruti Bhagavatula and Lujo Bauer and Michael K. Reiter},
	booktitle = {{Proceedings of the ACM Conference on Computer and Communications Security}},
	pages = {1528--1540},
	title = {{Accessorize to a Crime: Real and Stealthy Attacks on State-of-the-Art Face Recognition}},
	year = {2016}
}

@inproceedings{moosavi2016deepfool,
	author = {Seyed{-}Mohsen Moosavi{-}Dezfooli and Alhussein Fawzi and Pascal Frossard},
	booktitle = {{Proceedings of the IEEE/CVF Conference on Computer Vision and Pattern Recognition}},
	pages = {2574--2582},
	title = {{DeepFool:{ A} Simple and Accurate Method to Fool Deep Neural Networks}},
	year = {2016}
}

@inproceedings{madry2017towards,
	author = {Aleksander Madry and Aleksandar Makelov and Ludwig Schmidt and Dimitris Tsipras and Adrian Vladu},
	booktitle = {{Proceedings of the International Conference on Learning Representations}},
	title = {{Towards Deep Learning Models Resistant to Adversarial Attacks}},
	year = {2018}
}

@inproceedings{croce2020reliable,
	author = {Francesco Croce and Matthias Hein},
	booktitle = {{Proceedings of the International Conference on Machine Learning}},
	pages = {2206--2216},
	title = {{Reliable Evaluation of Adversarial Robustness with an Ensemble of Diverse Parameter-free Attacks}},
	year = {2020}
}

@inproceedings{wang2022triangle,
	author = {Xiaosen Wang and Zeliang Zhang and Kangheng Tong and Dihong Gong and Kun He and Zhifeng Li and Wei Liu},
	booktitle = {{Proceedings of the European Conference on Computer Vision}},
	pages = {156--174},
	title = {{Triangle Attack:{ A} Query-Efficient Decision-Based Adversarial Attack}},
	year = {2022}
}

@inproceedings{brendel2017decision,
	author = {Wieland Brendel and Jonas Rauber and Matthias Bethge},
	booktitle = {{Proceedings of the International Conference on Learning Representations}},
	title = {{Decision-Based Adversarial Attacks: Reliable Attacks Against Black-Box Machine Learning Models}},
	year = {2018}
}

@inproceedings{cheng2018query,
	author = {Minhao Cheng and Thong Le and Pin{-}Yu Chen and Huan Zhang and Jinfeng Yi and Cho{-}Jui Hsieh},
	booktitle = {{Proceedings of the International Conference on Learning Representations}},
	title = {{Query-Efficient Hard-label Black-box Attack: An Optimization-based Approach}},
	year = {2019}
}

@inproceedings{li2020qeba,
	author = {Huichen Li and Xiaojun Xu and Xiaolu Zhang and Shuang Yang and Bo Li},
	booktitle = {{Proceedings of the IEEE/CVF Conference on Computer Vision and Pattern Recognition}},
	pages = {1218--1227},
	title = {{{ QEBA:} Query-Efficient Boundary-Based Blackbox Attack}},
	year = {2020}
}

@inproceedings{ilyas2018black,
	author = {Andrew Ilyas and Logan Engstrom and Anish Athalye and Jessy Lin},
	booktitle = {{Proceedings of the International Conference on Machine Learning}},
	pages = {2142--2151},
	title = {{Black-box Adversarial Attacks with Limited Queries and Information}},
	year = {2018}
}

@inproceedings{cheng2019improving,
	author = {Shuyu Cheng and Yinpeng Dong and Tianyu Pang and Hang Su and Jun Zhu},
	booktitle = {{Proceedings of the Advances in Neural Information Processing Systems}},
	pages = {10932--10942},
	title = {{Improving Black-box Adversarial Attacks with a Transfer-based Prior}},
	year = {2019}
}

@inproceedings{du2019query,
	author = {Jiawei Du and Hu Zhang and Joey Tianyi Zhou and Yi Yang and Jiashi Feng},
	booktitle = {{Proceedings of the International Conference on Learning Representations}},
	title = {{Query-efficient Meta Attack to Deep Neural Networks}},
	year = {2020}
}

@inproceedings{al2019sign,
	author = {Abdullah Al{-}Dujaili and Una{-}May O'Reilly},
	booktitle = {{Proceedings of the International Conference on Learning Representations}},
	title = {{Sign Bits Are All You Need for Black-Box Attacks}},
	year = {2020}
}

@inproceedings{springer2021little,
	author = {Jacob M. Springer and Melanie Mitchell and Garrett T. Kenyon},
	booktitle = {{Proceedings of the Advances in Neural Information Processing Systems}},
	pages = {9759--9773},
	title = {{A Little Robustness Goes a Long Way: Leveraging Robust Features for Targeted Transfer Attacks}},
	year = {2021}
}

@inproceedings{heo2021rethinking,
  title={Rethinking spatial dimensions of vision transformers},
  author={Heo, Byeongho and Yun, Sangdoo and Han, Dongyoon and Chun, Sanghyuk and Choe, Junsuk and Oh, Seong Joon},
  booktitle={Proceedings of the IEEE/CVF international conference on computer vision},
  pages={11936--11945},
  year={2021}
}

@inproceedings{chen2021visformer,
  title={Visformer: The vision-friendly transformer},
  author={Chen, Zhengsu and Xie, Lingxi and Niu, Jianwei and Liu, Xuefeng and Wei, Longhui and Tian, Qi},
  booktitle={Proceedings of the IEEE/CVF international conference on computer vision},
  pages={589--598},
  year={2021}
}

@inproceedings{liao2018defense,
  title={{Defense against Adversarial Attacks using High-level Representation Guided Denoiser}},
  author={Liao, Fangzhou and Liang, Ming and Dong, Yinpeng and Pang, Tianyu and Hu, Xiaolin and Zhu, Jun},
  booktitle={Proceedings of the IEEE/CVF Conference on Computer Vision and Pattern Recognition},
  pages={1778--1787},
  year={2018}
}

@inproceedings{naseer2020a,
  title={{A Self-supervised Approach for Adversarial Robustness}},
  author={Naseer, Muzammal and Khan, Salman and Hayat, Munawar and Khan, Fahad Shahbaz and Porikli, Fatih},
  booktitle={Proceedings of the IEEE/CVF Conference on Computer Vision and Pattern Recognition},
  pages={262--271},
  year={2020}
}

@inproceedings{Cohen2019RS,
 author = {Cohen, Jeremy M and Rosenfeld, Elan and Kolter, J Zico},
 booktitle = {International Conference on Machine Learning},
 title = {{Certified adversarial robustness via randomized smoothing}},
 year = {2019}
}

@article{nie2022diffusion,
  title={Diffusion models for adversarial purification},
  author={Nie, Weili and Guo, Brandon and Huang, Yujia and Xiao, Chaowei and Vahdat, Arash and Anandkumar, Anima},
  journal={arXiv preprint arXiv:2205.07460},
  year={2022}
}

@article{michel1999interaction,
	author = {Michel Grabisch and Marc Roubens},
	journal = {{Int. J. Game Theory}},
	number = {4},
	pages = {547--565},
	title = {{An Axiomatic Approach to the Concept of Interaction among Players in Cooperative Games}},
	volume = {28},
	year = {1999}
}

@inproceedings{yu2022texthacker,
     title={{TextHacker: Learning based Hybrid Local Search Algorithm for Text Hard-label Adversarial Attack}},
     author={Zhen Yu and Xiaosen Wang and Wanxiang Che and Kun He},
     booktitle={Proceedings of the Conference on Empirical Methods in Natural Language Processing (Findings)},
     pages={622–637},
     year={2022}
}

@article{wong2020fast,
  title={Fast is better than free: Revisiting adversarial training},
  author={Wong, Eric and Rice, Leslie and Kolter, J Zico},
  journal={arXiv preprint arXiv:2001.03994},
  year={2020}
}

@article{li2024towards,
  title={Towards evaluating transfer-based attacks systematically, practically, and fairly},
  author={Li, Qizhang and Guo, Yiwen and Zuo, Wangmeng and Chen, Hao},
  journal={Advances in Neural Information Processing Systems},
  volume={36},
  year={2024}
}

@article{gu2023survey,
  title={A survey on transferability of adversarial examples across deep neural networks},
  author={Gu, Jindong and Jia, Xiaojun and de Jorge, Pau and Yu, Wenqain and Liu, Xinwei and Ma, Avery and Xun, Yuan and Hu, Anjun and Khakzar, Ashkan and Li, Zhijiang and others},
  journal={arXiv preprint arXiv:2310.17626},
  year={2023}
}

@article{zhao2023revisiting,
  title={Revisiting transferable adversarial image examples: Attack categorization, evaluation guidelines, and new insights},
  author={Zhao, Zhengyu and Zhang, Hanwei and Li, Renjue and Sicre, Ronan and Amsaleg, Laurent and Backes, Michael and Li, Qi and Shen, Chao},
  journal={arXiv preprint arXiv:2310.11850},
  year={2023}
}

@article{zhang2024bag,
     title={{Bag of Tricks to Boost Adversarial Transferability}},
     author={Zeliang Zhang and Rongyi Zhu and Wei Yao and Xiaosen Wang and Chenliang Xu},
     journal={Proceedings of the European Conference on Computer Vision},
     year={2024}
}

@article{jin2024short,
  title={Short: Benchmarking transferable adversarial attacks},
  author={Jin, Zhibo and Zhang, Jiayu and Zhu, Zhiyu and Chen, Huaming},
  journal={arXiv preprint arXiv:2402.00418},
  year={2024}
}

@article{wang2023beyond,
  title={Beyond Boundaries: A Comprehensive Survey of Transferable Attacks on AI Systems},
  author={Wang, Guangjing and Zhou, Ce and Wang, Yuanda and Chen, Bocheng and Guo, Hanqing and Yan, Qiben},
  journal={arXiv preprint arXiv:2311.11796},
  year={2023}
}

@inproceedings{yang2024mma,
     title={{MMA-Diffusion: MultiModal Attack on Diffusion Models}},
     author={Yijun Yang and Ruiyuan Gao and Xiaosen Wang and Nan Xu and Qiang Xu},
     booktitle={Proceedings of the IEEE/CVF International Conference on Computer Vision},
     pages={7737-7746},
     year={2024}
}

@article{vaswani2017attention,
  title={Attention is all you need},
  author={Vaswani, Ashish and Shazeer, Noam and Parmar, Niki and Uszkoreit, Jakob and Jones, Llion and Gomez, Aidan N and Kaiser, {\L}ukasz and Polosukhin, Illia},
  journal={Advances in neural information processing systems},
  volume={30},
  year={2017}
}

@article{devlin2018bert,
  title={Bert: Pre-training of deep bidirectional transformers for language understanding},
  author={Devlin, Jacob and Chang, Ming-Wei and Lee, Kenton and Toutanova, Kristina},
  journal={arXiv preprint arXiv:1810.04805},
  year={2018}
}

@article{brown2020language,
  title={Language models are few-shot learners},
  author={Brown, Tom and Mann, Benjamin and Ryder, Nick and Subbiah, Melanie and Kaplan, Jared D and Dhariwal, Prafulla and Neelakantan, Arvind and Shyam, Pranav and Sastry, Girish and Askell, Amanda and others},
  journal={Advances in neural information processing systems},
  volume={33},
  pages={1877--1901},
  year={2020}
}

@article{hochreiter1997long,
  title={Long short-term memory},
  author={Hochreiter, Sepp and Schmidhuber, J{\"u}rgen},
  journal={Neural computation},
  volume={9},
  number={8},
  pages={1735--1780},
  year={1997},
  publisher={MIT press}
}

@article{dong2021query,
  title={Query-efficient black-box adversarial attacks guided by a transfer-based prior},
  author={Dong, Yinpeng and Cheng, Shuyu and Pang, Tianyu and Su, Hang and Zhu, Jun},
  journal={IEEE Transactions on Pattern Analysis and Machine Intelligence},
  volume={44},
  number={12},
  pages={9536--9548},
  year={2021},
  publisher={IEEE}
}

@inproceedings{reza2023cgba,
  title={Cgba: curvature-aware geometric black-box attack},
  author={Reza, Md Farhamdur and Rahmati, Ali and Wu, Tianfu and Dai, Huaiyu},
  booktitle={Proceedings of the IEEE/CVF International Conference on Computer Vision},
  pages={124--133},
  year={2023}
}

@inproceedings{wang2018cosface,
  title={Cosface: Large margin cosine loss for deep face recognition},
  author={Wang, Hao and Wang, Yitong and Zhou, Zheng and Ji, Xing and Gong, Dihong and Zhou, Jingchao and Li, Zhifeng and Liu, Wei},
 booktitle = {Conference on Computer Vision and Pattern Recognition},
  year={2018}
}

@inproceedings{wen2016discriminative,
  title={A discriminative feature learning approach for deep face recognition},
  author={Wen, Yandong and Zhang, Kaipeng and Li, Zhifeng and Qiao, Yu},
  booktitle={European Conference on Computer Vision},
  year={2016},
}

@inproceedings{tang2004video,
  title={Video based face recognition using multiple classifiers},
  author={Tang, Xiaoou and Li, Zhifeng},
  booktitle={IEEE International Conference on Automatic Face and Gesture Recognition},
  pages={345--349},
  year={2004},
  organization={IEEE}
}

@inproceedings{xu2017end,
  title={End-to-end learning of driving models from large-scale video datasets},
  author={Xu, Huazhe and Gao, Yang and Yu, Fisher and Darrell, Trevor},
  booktitle={Conference on Computer Vision and Pattern Recognition},
  pages={2174--2182},
  year={2017}
}

@inproceedings{zhou2018voxelnet,
  title={Voxelnet: End-to-end learning for point cloud based 3d object detection},
  author={Zhou, Yin and Tuzel, Oncel},
  booktitle={Proceedings of the IEEE conference on computer vision and pattern recognition},
  pages={4490--4499},
  year={2018}
}

@inproceedings{vora2020pointpainting,
  title={Pointpainting: Sequential fusion for 3d object detection},
  author={Vora, Sourabh and Lang, Alex H and Helou, Bassam and Beijbom, Oscar},
  booktitle={Proceedings of the IEEE/CVF conference on computer vision and pattern recognition},
  pages={4604--4612},
  year={2020}
}

@inproceedings{wei2019transferable,
	author = {Xingxing Wei and Siyuan Liang and Ning Chen and Xiaochun Cao},
	booktitle = {{Proceedings of the International Joint Conference on Artificial Intelligence}},
	pages = {954--960},
	title = {{Transferable Adversarial Attacks for Image and Video Object Detection}},
	year = {2019}
}

@inproceedings{lu2023set,
  title={Set-level guidance attack: Boosting adversarial transferability of vision-language pre-training models},
  author={Lu, Dong and Wang, Zhiqiang and Wang, Teng and Guan, Weili and Gao, Hongchang and Zheng, Feng},
  booktitle={Proceedings of the IEEE/CVF International Conference on Computer Vision},
  pages={102--111},
  year={2023}
}

@article{zou2023universal,
  title={Universal and transferable adversarial attacks on aligned language models},
  author={Zou, Andy and Wang, Zifan and Kolter, J Zico and Fredrikson, Matt},
  journal={arXiv preprint arXiv:2307.15043},
  year={2023}
}

@inproceedings{xu2024highly,
  title={Highly Transferable Diffusion-based Unrestricted Adversarial Attack on Pre-trained Vision-Language Models},
  author={Xu, Wenzhuo and Chen, Kai and Gao, Ziyi and Wei, Zhipeng and Chen, Jingjing and Jiang, Yu-Gang},
  booktitle={ACM Multimedia},
  year={2024}
}

@inproceedings{ding2024transferable,
  title={Transferable Adversarial Attacks for Object Detection Using Object-Aware Significant Feature Distortion},
  author={Ding, Xinlong and Chen, Jiansheng and Yu, Hongwei and Shang, Yu and Qin, Yining and Ma, Huimin},
  booktitle={Proceedings of the AAAI Conference on Artificial Intelligence},
  volume={38},
  number={2},
  pages={1546--1554},
  year={2024}
}

@inproceedings{liu2024adv,
  title={Adv-diffusion: imperceptible adversarial face identity attack via latent diffusion model},
  author={Liu, Decheng and Wang, Xijun and Peng, Chunlei and Wang, Nannan and Hu, Ruimin and Gao, Xinbo},
  booktitle={Proceedings of the AAAI Conference on Artificial Intelligence},
  volume={38},
  number={4},
  pages={3585--3593},
  year={2024}
}

@inproceedings{huang2023t,
  title={T-sea: Transfer-based self-ensemble attack on object detection},
  author={Huang, Hao and Chen, Ziyan and Chen, Huanran and Wang, Yongtao and Zhang, Kevin},
  booktitle={Proceedings of the IEEE/CVF conference on computer vision and pattern recognition},
  pages={20514--20523},
  year={2023}
}

@inproceedings{lapid2024open,
  title={Open Sesame! Universal Black-Box Jailbreaking of Large Language Models},
  author={Lapid, Raz and Langberg, Ron and Sipper, Moshe},
  booktitle={ICLR 2024 Workshop on Secure and Trustworthy Large Language Models}
}

@article{mehrotra2023tree,
  title={Tree of attacks: Jailbreaking black-box llms automatically},
  author={Mehrotra, Anay and Zampetakis, Manolis and Kassianik, Paul and Nelson, Blaine and Anderson, Hyrum and Singer, Yaron and Karbasi, Amin},
  journal={arXiv preprint arXiv:2312.02119},
  year={2023}
}

@inproceedings{liuautodan,
  title={AutoDAN: Generating Stealthy Jailbreak Prompts on Aligned Large Language Models},
  author={Liu, Xiaogeng and Xu, Nan and Chen, Muhao and Xiao, Chaowei},
  booktitle={The Twelfth International Conference on Learning Representations}
}

@article{jin2024guard,
  title={GUARD: Role-playing to generate natural-language jailbreakings to test guideline adherence of large language models},
  author={Jin, Haibo and Chen, Ruoxi and Zhou, Andy and Zhang, Yang and Wang, Haohan},
  journal={arXiv preprint arXiv:2402.03299},
  year={2024}
}

@article{zeng2024johnny,
  title={How johnny can persuade llms to jailbreak them: Rethinking persuasion to challenge ai safety by humanizing llms},
  author={Zeng, Yi and Lin, Hongpeng and Zhang, Jingwen and Yang, Diyi and Jia, Ruoxi and Shi, Weiyan},
  journal={arXiv preprint arXiv:2401.06373},
  year={2024}
}

@inproceedings{zheng2024unified,
  title={A Unified Understanding of Adversarial Vulnerability Regarding Unimodal Models and Vision-Language Pre-training Models},
  author={Zheng, Haonan and Deng, Xinyang and Jiang, Wen and Li, Wenrui},
  booktitle={Proceedings of the 32nd ACM International Conference on Multimedia},
  pages={18--27},
  year={2024}
}

@inproceedings{zhang2024universal,
  title={Universal adversarial perturbations for vision-language pre-trained models},
  author={Zhang, Peng-Fei and Huang, Zi and Bai, Guangdong},
  booktitle={Proceedings of the 47th International ACM SIGIR Conference on Research and Development in Information Retrieval},
  pages={862--871},
  year={2024}
}

@inproceedings{zhang2025adversarial,
  title={Adversarial prompt tuning for vision-language models},
  author={Zhang, Jiaming and Ma, Xingjun and Wang, Xin and Qiu, Lingyu and Wang, Jiaqi and Jiang, Yu-Gang and Sang, Jitao},
  booktitle={European Conference on Computer Vision},
  pages={56--72},
  year={2025},
  organization={Springer}
}

@article{qraitem2024vision,
  title={Vision-llms can fool themselves with self-generated typographic attacks},
  author={Qraitem, Maan and Tasnim, Nazia and Teterwak, Piotr and Saenko, Kate and Plummer, Bryan A},
  journal={arXiv preprint arXiv:2402.00626},
  year={2024}
}

@article{zhao2024evaluating,
  title={On evaluating adversarial robustness of large vision-language models},
  author={Zhao, Yunqing and Pang, Tianyu and Du, Chao and Yang, Xiao and Li, Chongxuan and Cheung, Ngai-Man Man and Lin, Min},
  journal={Advances in Neural Information Processing Systems},
  volume={36},
  year={2024}
}

@inproceedings{sun2024makeupattack,
  title={MakeupAttack: Feature Space Black-Box Backdoor Attack on Face Recognition via Makeup Transfer},
  author={Sun, Ming and Jing, Lihua and Zhu, Zixuan and Wang, Rui},
  booktitle={ECAI},
  year={2024}
}

@inproceedings{katzav2025adversarialeak,
  title={Adversarialeak: External information leakage attack using adversarial samples on face recognition systems},
  author={Katzav, Roye and Giloni, Amit and Grolman, Edita and Saito, Hiroo and Shibata, Tomoyuki and Omino, Tsukasa and Komatsu, Misaki and Hanatani, Yoshikazu and Elovici, Yuval and Shabtai, Asaf},
  booktitle={European Conference on Computer Vision},
  pages={288--303},
  year={2025},
  organization={Springer}
}

@inproceedings{zhou2024improving,
  title={Improving Visual Quality and Transferability of Adversarial Attacks on Face Recognition Simultaneously with Adversarial Restoration},
  author={Zhou, Fengfan and Ling, Hefei and Shi, Yuxuan and Chen, Jiazhong and Li, Ping},
  booktitle={ICASSP 2024-2024 IEEE International Conference on Acoustics, Speech and Signal Processing (ICASSP)},
  pages={4540--4544},
  year={2024},
  organization={IEEE}
}

@inproceedings{zhou2024rethinking,
  title={Rethinking impersonation and dodging attacks on face recognition systems},
  author={Zhou, Fengfan and Zhou, Qianyu and Yin, Bangjie and Zheng, Hui and Lu, Xuequan and Ma, Lizhuang and Ling, Hefei},
  booktitle={Proceedings of the 32nd ACM International Conference on Multimedia},
  pages={2487--2496},
  year={2024}
}

@inproceedings{li2023sibling,
  title={Sibling-attack: Rethinking transferable adversarial attacks against face recognition},
  author={Li, Zexin and Yin, Bangjie and Yao, Taiping and Guo, Junfeng and Ding, Shouhong and Chen, Simin and Liu, Cong},
  booktitle={Proceedings of the IEEE/CVF Conference on Computer Vision and Pattern Recognition},
  pages={24626--24637},
  year={2023}
}

@inproceedings{bao2024glow,
  title={GLOW: Global Layout Aware Attacks on Object Detection},
  author={Bao, Jun and Liu, Buyu and Ren, Kui and Yu, Jun},
  booktitle={Proceedings of the IEEE/CVF Conference on Computer Vision and Pattern Recognition},
  pages={12057--12066},
  year={2024}
}

@inproceedings{xu2023backdoor,
  title={Backdoor Attack on 3D Grey Image Segmentation},
  author={Xu, Honghui and Cai, Zhipeng and Xiong, Zuobin and Li, Wei},
  booktitle={2023 IEEE International Conference on Data Mining (ICDM)},
  pages={708--717},
  year={2023},
  organization={IEEE}
}

@article{gao2023backdoor,
  title={Backdoor Attack on Hash-based Image Retrieval via Clean-label Data Poisoning},
  author={Gao, Kuofeng and Bai, Jiawang and Chen, Bin and Wu, Dongxian and Xia, Shu-Tao},
  year={2023}
}

@article{ding2023vith,
  title={ViTH-RFG: Vision Transformer Hashing with Residual Fuzzy Generation for Targeted Attack in Medical Image Retrieval},
  author={Ding, Weiping and Liu, Chuansheng and Huang, Jiashuang and Cheng, Chun and Ju, Hengrong},
  journal={IEEE Transactions on Fuzzy Systems},
  year={2023},
  publisher={IEEE}
}

@article{li2023semantic,
  title={Semantic-Aware Attack and Defense on Deep Hashing Networks for Remote-Sensing Image Retrieval},
  author={Li, Yansheng and Hao, Mengze and Liu, Rongjie and Zhang, Zhichao and Zhu, Hu and Zhang, Yongjun},
  journal={IEEE Transactions on Geoscience and Remote Sensing},
  volume={61},
  pages={1--14},
  year={2023},
  publisher={IEEE}
}

@inproceedings{jin2020bert,
  title={Is bert really robust? a strong baseline for natural language attack on text classification and entailment},
  author={Jin, Di and Jin, Zhijing and Zhou, Joey Tianyi and Szolovits, Peter},
  booktitle={Proceedings of the AAAI conference on artificial intelligence},
  volume={34},
  number={05},
  pages={8018--8025},
  year={2020}
}

@inproceedings{zang2020word,
  title={Word-level textual adversarial attacking as combinatorial optimization},
  author={Zang, Yuan and Qi, Fanchao and Yang, Chenghao and Liu, Zhiyuan and Zhang, Meng and Liu, Qun and Sun, Maosong},
  booktitle={Proceedings of the 58th annual meeting of the association for computational linguistics},
  pages={6066--6080},
  year={2020}
}

@inproceedings{jiang2024artprompt,
  title={ArtPrompt: ASCII Art-based Jailbreak Attacks against Aligned LLMs},
  author={Jiang, Fengqing and Xu, Zhangchen and Niu, Luyao and Xiang, Zhen and Ramasubramanian, Bhaskar and Li, Bo and Poovendran, Radha},
  booktitle={ICLR 2024 Workshop on Secure and Trustworthy Large Language Models}
}

@inproceedings{ghanim2024jailbreaking,
  title={Jailbreaking LLMs with Arabic Transliteration and Arabizi},
  author={Ghanim, Mansour and Almohaimeed, Saleh and Zheng, Mengxin and Solihin, Yan and Lou, Qian},
  booktitle={Proceedings of the 2024 Conference on Empirical Methods in Natural Language Processing},
  pages={18584--18600},
  year={2024}
}

@article{li2024drattack,
  title={Drattack: Prompt decomposition and reconstruction makes powerful llm jailbreakers},
  author={Li, Xirui and Wang, Ruochen and Cheng, Minhao and Zhou, Tianyi and Hsieh, Cho-Jui},
  journal={arXiv preprint arXiv:2402.16914},
  year={2024}
}

@article{lin2024towards,
  title={Towards Understanding Jailbreak Attacks in LLMs: A Representation Space Analysis},
  author={Lin, Yuping and He, Pengfei and Xu, Han and Xing, Yue and Yamada, Makoto and Liu, Hui and Tang, Jiliang},
  journal={arXiv preprint arXiv:2406.10794},
  year={2024}
}

@article{liao2024amplegcg,
  title={Amplegcg: Learning a universal and transferable generative model of adversarial suffixes for jailbreaking both open and closed llms},
  author={Liao, Zeyi and Sun, Huan},
  journal={arXiv preprint arXiv:2404.07921},
  year={2024}
}

@inproceedings{yin2024vqattack,
  title={VQAttack: Transferable Adversarial Attacks on Visual Question Answering via Pre-trained Models},
  author={Yin, Ziyi and Ye, Muchao and Zhang, Tianrong and Wang, Jiaqi and Liu, Han and Chen, Jinghui and Wang, Ting and Ma, Fenglong},
  booktitle={Proceedings of the AAAI Conference on Artificial Intelligence},
  volume={38},
  number={7},
  pages={6755--6763},
  year={2024}
}

@inproceedings{zhang2022towards,
  title={Towards adversarial attack on vision-language pre-training models},
  author={Zhang, Jiaming and Yi, Qi and Sang, Jitao},
  booktitle={Proceedings of the 30th ACM International Conference on Multimedia},
  pages={5005--5013},
  year={2022}
}

@inproceedings{gu2024agent,
  title={Agent Smith: A Single Image Can Jailbreak One Million Multimodal LLM Agents Exponentially Fast},
  author={Gu, Xiangming and Zheng, Xiaosen and Pang, Tianyu and Du, Chao and Liu, Qian and Wang, Ye and Jiang, Jing and Lin, Min},
  booktitle={ICLR 2024 Workshop on Large Language Model (LLM) Agents},
year={2024}
}

@inproceedings{gao2025boosting,
  title={Boosting transferability in vision-language attacks via diversification along the intersection region of adversarial trajectory},
  author={Gao, Sensen and Jia, Xiaojun and Ren, Xuhong and Tsang, Ivor and Guo, Qing},
  booktitle={European Conference on Computer Vision},
  pages={442--460},
  year={2025},
  organization={Springer}
}

@article{lyu2024trojvlm,
  title={Trojvlm: Backdoor attack against vision language models},
  author={Lyu, Weimin and Pang, Lu and Ma, Tengfei and Ling, Haibin and Chen, Chao},
  journal={arXiv preprint arXiv:2409.19232},
  year={2024}
}

@inproceedings{wang2024break,
  title={Break the visual perception: Adversarial attacks targeting encoded visual tokens of large vision-language models},
  author={Wang, Yubo and Liu, Chaohu and Qu, Yanqiu and Cao, Haoyu and Jiang, Deqiang and Xu, Linli},
  booktitle={Proceedings of the 32nd ACM International Conference on Multimedia},
  pages={1072--1081},
  year={2024}
}

@inproceedings{wang2024transferable,
  title={Transferable multimodal attack on vision-language pre-training models},
  author={Wang, Haodi and Dong, Kai and Zhu, Zhilei and Qin, Haotong and Liu, Aishan and Fang, Xiaolin and Wang, Jiakai and Liu, Xianglong},
  booktitle={2024 IEEE Symposium on Security and Privacy (SP)},
  pages={102--102},
  year={2024},
  organization={IEEE Computer Society}
}

@inproceedings{liupixel2feature,
  title={Pixel2Feature Attack (P2FA): Rethinking the Perturbed Space to Enhance Adversarial Transferability},
  author={Liu, Renpu and Wu, Hao and Zhang, Jiawei and Cheng, Xin and Luo, Xiangyang and Ma, Bin and Wang, Jinwei},
  booktitle={Forty-second International Conference on Machine Learning},
  year={2025}
}

@inproceedings{ren2025improving,
  title={Improving integrated gradient-based transferable adversarial examples by refining the integration path},
  author={Ren, Yuchen and Zhao, Zhengyu and Lin, Chenhao and Yang, Bo and Zhou, Lu and Liu, Zhe and Shen, Chao},
  booktitle={Proceedings of the AAAI Conference on Artificial Intelligence},
  volume={39},
  number={7},
  pages={6731--6739},
  year={2025}
}

@inproceedings{li2025aim,
  title={AIM: Additional Image Guided Generation of Transferable Adversarial Attacks},
  author={Li, Teng and Ma, Xingjun and Jiang, Yu-Gang},
  booktitle={Proceedings of the AAAI Conference on Artificial Intelligence},
  volume={39},
  number={5},
  pages={4941--4949},
  year={2025}
}

@inproceedings{wang2025breaking,
  title={Breaking barriers in physical-world adversarial examples: Improving robustness and transferability via robust feature},
  author={Wang, Yichen and Chou, Yuxuan and Zhou, Ziqi and Zhang, Hangtao and Wan, Wei and Hu, Shengshan and Li, Minghui},
  booktitle={Proceedings of the AAAI Conference on Artificial Intelligence},
  volume={39},
  number={8},
  pages={8069--8077},
  year={2025}
}

@article{li2025enhancing,
  title={Enhancing Adversarial Transferability with Alignment Network},
  author={Li, Zhiwei and Li, Qi and Ren, Min and Ru, Yiwei and Sun, Zhenan},
  journal={IEEE Transactions on Information Forensics and Security},
  year={2025},
  publisher={IEEE}
}

@inproceedings{zeng2025everywhere,
  title={Everywhere Attack: Attacking Locally and Globally to Boost Targeted Transferability},
  author={Zeng, Hui and Cui, Sanshuai and Chen, Biwei and Peng, Anjie},
  booktitle={Proceedings of the AAAI Conference on Artificial Intelligence},
  volume={39},
  number={9},
  pages={9789--9796},
  year={2025}
}

@article{qian2024enhancing,
  author={Qian, Yaguan and Chen, Kecheng and Wang, Bin and Gu, Zhaoquan and Ji, Shouling and Wang, Wei and Zhang, Yanchun},
  journal={IEEE Transactions on Information Forensics and Security}, 
  title={Enhancing Transferability of Adversarial Examples Through Mixed-Frequency Inputs}, 
  year={2024},
  volume={19},
  pages={7633-7645}
}

@inproceedings{long2024convergence,
  title={On the convergence of an adaptive momentum method for adversarial attacks},
  author={Long, Sheng and Tao, Wei and Lei, Jun and Zhang, Jun and others},
  booktitle={Proceedings of the AAAI Conference on Artificial Intelligence},
  volume={38},
  number={13},
  pages={14132--14140},
  year={2024}
}

@inproceedings{gao2024attacking,
  title={Attacking transformers with feature diversity adversarial perturbation},
  author={Gao, Chenxing and Zhou, Hang and Yu, Junqing and Ye, YuTeng and Cai, Jiale and Wang, Junle and Yang, Wei},
  booktitle={Proceedings of the AAAI Conference on Artificial Intelligence},
  volume={38},
  number={3},
  pages={1788--1796},
  year={2024}
}

@inproceedings{yuan2021transferability,
  title={On the transferability of adversarial attacks against neural text classifier},
  author={Yuan, Liping and Zheng, Xiaoqing and Zhou, Yi and Hsieh, Cho-Jui and Chang, Kai-Wei},
  booktitle={Proceedings of the 2021 Conference on Empirical Methods in Natural Language Processing},
  pages={1612--1625},
  year={2021}
}

@article{roth2024token,
  title={Token-modification adversarial attacks for natural language processing: A survey},
  author={Roth, Tom and Gao, Yansong and Abuadbba, Alsharif and Nepal, Surya and Liu, Wei},
  journal={AI Communications},
  volume={37},
  number={4},
  pages={655--676},
  year={2024},
  publisher={SAGE Publications Sage UK: London, England}
}

@inproceedings{wei2022cross,
  title={Cross-modal transferable adversarial attacks from images to videos},
  author={Wei, Zhipeng and Chen, Jingjing and Wu, Zuxuan and Jiang, Yu-Gang},
  booktitle={Proceedings of the IEEE/CVF conference on computer vision and pattern recognition},
  pages={15064--15073},
  year={2022}
}

@article{yin2023vlattack,
  title={Vlattack: Multimodal adversarial attacks on vision-language tasks via pre-trained models},
  author={Yin, Ziyi and Ye, Muchao and Zhang, Tianrong and Du, Tianyu and Zhu, Jinguo and Liu, Han and Chen, Jinghui and Wang, Ting and Ma, Fenglong},
  journal={Advances in Neural Information Processing Systems},
  volume={36},
  pages={52936--52956},
  year={2023}
}

@article{qin2025generalized,
  title={Generalized Transferable Attack across Datasets},
  author={Qin, Yunxiao and Xiong, Yuanhao and Yi, Jinfeng and Cao, Lihong and Hsieh, Cho-Jui},
  journal={IEEE Transactions on Circuits and Systems for Video Technology},
  year={2025},
  publisher={IEEE}
}

@inproceedings{fu2023styleadv,
  title={Styleadv: Meta style adversarial training for cross-domain few-shot learning},
  author={Fu, Yuqian and Xie, Yu and Fu, Yanwei and Jiang, Yu-Gang},
  booktitle={Proceedings of the IEEE/CVF conference on computer vision and pattern recognition},
  pages={24575--24584},
  year={2023}
}

@inproceedings{li2023cdta,
  title={CDTA: a cross-domain transfer-based attack with contrastive learning},
  author={Li, Zihan and Wu, Weibin and Su, Yuxin and Zheng, Zibin and Lyu, Michael R},
  booktitle={Proceedings of the AAAI Conference on Artificial Intelligence},
  volume={37},
  number={2},
  pages={1530--1538},
  year={2023}
}

@article{wang2020unsupervised,
  title={Unsupervised adversarial domain adaptation for cross-domain face presentation attack detection},
  author={Wang, Guoqing and Han, Hu and Shan, Shiguang and Chen, Xilin},
  journal={IEEE Transactions on Information Forensics and Security},
  volume={16},
  pages={56--69},
  year={2020},
  publisher={IEEE}
}

@inproceedings{chao2025jailbreaking,
  title={Jailbreaking black box large language models in twenty queries},
  author={Chao, Patrick and Robey, Alexander and Dobriban, Edgar and Hassani, Hamed and Pappas, George J and Wong, Eric},
  booktitle={2025 IEEE Conference on Secure and Trustworthy Machine Learning (SaTML)},
  pages={23--42},
  year={2025},
  organization={IEEE}
}

@String(CVPR= {IEEE Conf. Comput. Vis. Pattern Recog.})

@String(ICCV= {Int. Conf. Comput. Vis.})

@String(NIPS= {Adv. Neural Inform. Process. Syst.})

@String(ICASSP=	{ICASSP})

@String(ICLR = {Int. Conf. Learn. Represent.})

@String(AAAI = {AAAI})

@String(EMNLP = {Conf. Empir. Meth. Nat. Lang. Process.})

@String(ECAI = {Eur. Conf. Artif. Intell.})

@String(ICDM = {IEEE Int. Conf. Data Min.})

@String(ACL = {Ann. Meet. Assoc. Comput. Linguist.})

@String(SATML = {IEEE Conf. Secure Trustworthy Mach. Learn.})

@String(CVPR  = {CVPR})

@String(ICCV  = {ICCV})

@String(NIPS  = {NeurIPS})

@String(ICLR  = {ICLR})

@String(EMNLP={EMNLP})

@String(ECAI={ECAI})

@String(ICDM={ICDM})

@String(ACL = {ACL})

@String(SIGIR={ACM SIGIR})

@String(SP={S\&P})

@String(SaTML={SaTML})

@String(AsiaCCS={AsiaCCS})

@inproceedings{fan2025maskblock,
    author = {Mingyuan Fan and Cen Chen and Ximeng Liu and Wenzhong Guo},
    booktitle = AsiaCCS,
    title = {{Transferable Adversarial Examples with Bayes Approach}},
    year = {2025}
}

@inproceedings{wang2025attention,
     title={{Attention! Your Vision Language Model Could Be Maliciously Manipulated}},
     author={Xiaosen Wang and Shaokang Wang and Zhijin Ge and Yuyang Luo and Shudong Zhang},
     booktitle=NIPS,
     year={2025}
}

@inproceedings{cao2025vit,
  title={ViT-EnsembleAttack: Augmenting Ensemble Models for Stronger Adversarial Transferability in Vision Transformers},
  author={Cao, Hanwen and Lu, Haobo and Wang, Xiaosen and He, Kun},
  booktitle=ICCV,
  year={2025}
}

@misc{luo2025disrupting,
  title={Disrupting Semantic and Abstract Features for Better Adversarial Transferability},
  author={Luo, Yuyang and Wang, Xiaosen and Ge, Zhijin and He, Yingzhe},
  note={arXiv:2507.16052},
  year={2025}
}

@misc{liu2025boosting,
  title={Boosting the Local Invariance for Better Adversarial Transferability},
  author={Liu, Bohan and Wang, Xiaosen},
  note={arXiv:2503.06140},
  year={2025}
}

@inproceedings{ren2019generating,
  title={{Generating Natural Language Adversarial Examples Through Probability Weighted Word Saliency}},
  author={Ren, Shuhuai and Deng, Yihe and He, Kun and Che, Wanxiang},
  booktitle=ACL,
  year={2019}
}

@inproceedings{li2020bert,
  title={{BERT-ATTACK: Adversarial Attack Against Bert Using Bert}},
  author={Li, Linyang and Ma, Ruotian and Guo, Qipeng and Xue, Xiangyang and Qiu, Xipeng},
  booktitle=EMNLP,
  year={2020}
}

@inproceedings{garg2020bae,
  title={{BAE: BERT-Based Adversarial Examples for Text Classification}},
  author={Garg, Siddhant and Ramakrishnan, Goutham},
  booktitle=EMNLP,
  year={2020}
}

@inproceedings{yang2024sneakyprompt,
  title={{SneakyPrompt: Jailbreaking Text-to-Image Generative Models}},
  author={Yang, Yuchen and Hui, Bo and Yuan, Haolin and Gong, Neil and Cao, Yinzhi},
  booktitle=SP,
  year={2024},
}

@inproceedings{zhuang2023pilot,
  title={{A Pilot Study of Query-Free Adversarial Attack Against Stable Diffusion}},
  author={Zhuang, Haomin and Zhang, Yihua and Liu, Sijia},
  booktitle=CVPR,
  year={2023}
}

@inproceedings{tsai2024ring,
  title={{Ring-A-Bell! How Reliable Are Concept Removal Methods for Diffusion Models?}},
  author={Tsai, Yu-Lin and Hsu, Chia-Yi and Xie, Chulin and Lin, Chih-Hsun and Chen, Jia-You and Li, Bo and Chen, Pin-Yu and Yu, Chia-Mu and Huang, Chun-Ying},
  booktitle=ICLR,
  year={2024}
}

@inproceedings{liu2023riatig,
  title={{RIATIG: Reliable and Imperceptible Adversarial Text-to-Image Generation with Natural Prompts}},
  author={Liu, Han and Wu, Yuhao and Zhai, Shixuan and Yuan, Bo and Zhang, Ning},
  booktitle=CVPR,
  year={2023}
}

@inproceedings{cheng2023fusion,
  title={{Fusion Is Not Enough: Single Modal Attacks on Fusion Models for 3D Object Detection}},
  author={Cheng, Zhiyuan and Choi, Hongjun and Liang, James and Feng, Shiwei and Tao, Guanhong and Liu, Dongfang and Zuzak, Michael and Zhang, Xiangyu},
  booktitle=ICLR,
  year={2023}
}

@inproceedings{arnab2018robustness,
  title={{On the Robustness of Semantic Segmentation Models to Adversarial Attacks}},
  author={Arnab, Anurag and Miksik, Ondrej and Torr, Philip HS},
  booktitle=CVPR,
  year={2018}
}
}

\vfill

\end{document}